\documentclass[conference]{IEEEtran}

\usepackage{times}

\usepackage[numbers]{natbib}
\usepackage{multicol}
\usepackage[bookmarks=true]{hyperref}
\usepackage{amsmath}
\usepackage{bm}
\usepackage{amssymb}
\usepackage{graphicx}
\usepackage{multirow}

\ifCLASSOPTIONcompsoc
    \usepackage[caption=false, font=normalsize, labelfont=sf, textfont=sf]{subfig}
\else
\usepackage[caption=false, font=footnotesize]{subfig}

\usepackage[dvipsnames]{xcolor}
\usepackage{cancel}
\usepackage{stmaryrd}

\usepackage{calrsfs}
\DeclareMathAlphabet{\pazocal}{OMS}{zplm}{m}{n}

\newcommand{\calA}{\pazocal{A}}

\newcommand{\calH}{\pazocal{H}}

\newcommand{\calI}{\pazocal{I}}

\newcommand{\calJ}{\pazocal{J}}

\newcommand{\calL}{\pazocal{L}}

\newcommand{\calN}{\pazocal{N}}

\newcommand{\calP}{\pazocal{P}}

\newcommand{\calU}{\pazocal{U}}

\newcommand{\calV}{\pazocal{V}}

\newcommand{\calX}{\pazocal{X}}

\usepackage{eufrak}

\newcommand{\Eb}{\mathbb{E}}
\newcommand{\Pb}{\mathbb{P}}

\newcommand{\Rb}{\mathbb{R}}
\newcommand{\Sb}{\mathbb{S}}
\newcommand{\Zb}{\mathbb{Z}}

\newcommand{\vzero}{{\bf 0}}

\newcommand{\diag}{\mathrm{diag}}

\newcommand{\tr}{\mathrm{tr}}

\newcommand{\Cov}{\mathrm{Cov}}

\newcommand{\vGamma}{{\mathbf{\Gamma}}}

\newcommand{\vG}{{\bf G}}

\newcommand{\vI}{{\bf I}}

\newcommand{\vA}{{\bf A}}
\newcommand{\vR}{{\bf R}}

\newcommand{\vK}{{\bf K}}

\newcommand{\vF}{{\bf F}}
\newcommand{\vM}{{\bf M}}

\newcommand{\vH}{{\bf H}}

\newcommand{\vQ}{{\bf Q}}

\newcommand{\vS}{{\bf S}}
\newcommand{\vX}{{\bf X}}

\newcommand{\vP}{{\bf P}}

\newcommand{\vPhi}{{\bf \Phi}}

\newcommand{\argmin}{\operatornamewithlimits{argmin}}
\newcommand{\T}{^\mathrm{T}}

\newcommand{\ba}{{\bm a}}

\newcommand{\bu}{{\bm u}}

\newcommand{\bw}{{\bm w}}
\newcommand{\bx}{{\bm x}}
\newcommand{\by}{{\bm y}}
\newcommand{\bz}{{\bm z}}

\newcommand{\bQ}{{\bm Q}}

\newcommand{\bdelta}{{\bm \delta}}

\newcommand{\bzeta}{{\bm \zeta}}
\newcommand{\blambda}{{\bm \lambda}}
\newcommand{\bmu}{{\bm \mu}}
\newcommand{\bnu}{{\bm \nu}}

\newcommand{\bSigma}{{\bm \Sigma}}

\DeclareFontFamily{OT1}{pzc}{}
\DeclareFontShape{OT1}{pzc}{m}{it}{<-> s * [1.10] pzcmi7t}{}
\DeclareMathAlphabet{\mathpzc}{OT1}{pzc}{m}{it}

\usepackage{amsthm}
\usepackage{soul}

\newtheorem{proposition}{Proposition}

\newtheorem{corollary}{Corollary}

\newtheorem{problem}{Problem}

\begin{document}

% paper title
\title{Scaling Robust Optimization for Multi-Agent Robotic Systems: A Distributed Perspective}
\author{\IEEEauthorblockN{Arshiya Taj Abdul\IEEEauthorrefmark{1},
Augustinos D. Saravanos\IEEEauthorrefmark{1},
Evangelos A. Theodorou
\IEEEauthorblockA{Georgia Institute of Technology,
GA, USA 
\\
\{aabdul6, asaravanos, evangelos.theodorou \}@gatech.edu
\\
\IEEEauthorrefmark{1}These authors contributed equally
}}}

\maketitle
\begin{abstract}

This paper presents a novel distributed robust optimization scheme for steering distributions of multi-agent systems under stochastic and deterministic uncertainty. Robust optimization is a subfield of optimization which aims to discover an optimal solution that remains robustly feasible for all possible realizations of the problem parameters within a given uncertainty set. Such approaches would naturally constitute an ideal candidate for multi-robot control, where in addition to stochastic noise, there might be exogenous deterministic disturbances. Nevertheless, as these methods are usually associated with significantly high computational demands, their application to multi-agent robotics has remained limited. The scope of this work is to propose a scalable robust optimization framework that effectively addresses both types of uncertainties, while retaining computational efficiency and scalability. In this direction, we provide tractable approximations for robust constraints that are relevant in multi-robot settings. Subsequently, we demonstrate how computations can be distributed through an Alternating Direction Method of Multipliers (ADMM) approach towards achieving scalability and communication efficiency. All improvements are also theoretically justified by establishing and comparing the resulting computational complexities. Simulation results highlight the performance of the proposed algorithm in effectively handling both stochastic and deterministic uncertainty in multi-robot systems. The scalability of the method is also emphasized by showcasing tasks with up to hundreds of agents. The results of this work indicate the promise of blending robust optimization, distribution steering and distributed optimization towards achieving scalable, safe and robust multi-robot control.  

\end{abstract}

\IEEEpeerreviewmaketitle

\section{Introduction}

Multi-agent optimization and control problems are emerging increasingly often in a vast variety of robotics applications such as multi-vehicle coordination \cite{cooperative2020, xiao2021bridging}, swarm robotics \cite{schranz2020swarm}, cooperative manipulation \cite{feng2020overview} and others. As the size and complexity of such systems has been rapidly escalating, a substantial requirement for developing scalable and distributed algorithms for tackling these problems has been developing as well \cite{saravanos2023ddp_tro, shah2022large, zhu_ferrari2021adaptive}. In the meantime, accompanying such algorithms with safety and robustness guarantees in environments with substantial levels of uncertainty is also a highly desirable attribute for securing the safe operation of all agents.

Uncertainty has been primarily modeled as stochastic noise in a significant portion of the multi-agent control literature. Some indicative approaches involve multi-agent LQG control \cite{alvergue2016consensus, nourian2012mean, shamma2008cooperative}, distributed covariance steering \cite{Saravanos-RSS-21, saravanos2024distributed_iros}, multi-agent path integral control \cite{gomez2016real, wan2021cooperative} and learning-based stochastic optimal control methods \cite{lin2021multi, Pereira-RSS-22}, among others. Nevertheless, such approaches would typically fall short in scenarios where the underlying disturbances might be exogenous interventions, environmental factors, modelling inaccuracies, etc., that cannot be properly represented as stochastic signals \cite{RobustNemirovski}. Such problems create the necessity of finding robust optimal policies that would guarantee feasibility for all possible values of such disturbances within some bounds. 

\textit{Robust optimization} is a subfield of optimization that addresses the problem of discovering an optimal solution that remains \textit{robust feasible} for all possible realizations of the problem parameters within a given uncertainty set \cite{RobustNemirovski, ben2006extending}. Although robust optimization has found significant successful applications in many areas such as power systems \cite{bertsimas2013adaptive}, supply chain networks \cite{pishvaee2011robust}, communication systems \cite{cai2020joint}, etc., there have only been limited attempts for addressing problems relevant to robotics \cite{kotsalis2020convex, lorenzen2019robust, quintero2021robust, taylor2021towards}. The field of robust control is a also closely related area with long history in addressing disturbances from predefined sets \cite{dorato1987historical, zhou1998essentials}. However, as this area is mainly focused on synthesis techniques for stabilization and performance maintenance, merging such methods with distributed optimization and steering state distributions for multi-agent systems is not as straightforward. On the contrary, as robust optimization allows for more flexibility in designing such approaches, the current work is focused on leveraging the latter for safe multi-agent control.

Another promising area that has recently emerged in the context of safe stochastic control is the one of steering the state distributions of systems under predefined target distributions - mostly known as \textit{covariance steering} \mbox{
\cite{bakolas2018finite, p:chen2015covariance1, 
liu2022optimal}}\hskip0pt. Such methods have found several successful applications in path planning \cite{okamoto2019optimal}, trajectory optimization \cite{balci2022constrained, yin2022trajectory}, robotic manipulation \cite{lee2022hierarchical}, flight control \cite{benedikter2022convex, rapakoulias2023discrete}, multi-robot control \cite{Saravanos-RSS-21, saravanos2024distributed_iros} and others. More recently, a convex optimization framework for steering state distributions and effectively addressing both stochastic noise and bounded deterministic disturbances was presented \cite{kotsalis2020convex}. In a similar direction, a data-driven robust optimization-based method was recently proposed in \mbox{
\cite{pilipovsky2023data} }\hskip0pt
, including estimating the noise realization from collected data. Nevertheless, as such approaches result in high-dimensional semidefinite programming (SDP) problems, the computational demands for addressing multi-agent control tasks have remained prohibitive.

The scope of this work is to provide a scalable robust optimization framework for steering distributions of multi-agent systems to specific targets under stochastic and deterministic uncertainty. Towards this direction, we present a distributed algorithm for safe and robust multi-robot control whose effectiveness and scalability are verified through simulation experiments.
The specific contributions of this paper can be listed as follows.
\begin{itemize}
\item We combine elements from the areas of robust optimization, distribution steering and distributed computation architectures for deriving a scalable and robust approach for multi-agent control. To our best knowledge, this is the first work to fuse these areas into a single methodology.
\item We circumvent computational expensiveness by providing tractable approximations of robust optimization constraints that are relevant for multi-robot control problems.
\item We present a distributed algorithm based on an Alternating Direction Method of Multipliers (ADMM) approach for multi-agent robust distribution steering. 
\item All computational improvements are theoretically justified by comparing the resulting computational complexities of different potential formulations.
\item Simulation results that demonstrate the effectiveness of the proposed algorithm in handling stochastic and deterministic uncertainty are provided in a variety of multi-robot tasks. The scalability of the method is also highlighted by showcasing scenarios with up to hundreds of agents and by comparing against a centralized approach.
\end{itemize}

The remaining of this paper is organized as follows. In Section \ref{sec: problem statement}, we state the multi-agent problem to be considered in this work. The proposed distributed robust optimization framework is presented in Section \ref{Distributed Optimization under deterministic representations of uncertainty}. An analysis of performance and scalability is demonstrated through simulation experiments in Section \ref{sec: simulation results}. Finally, the conclusion of this paper, as well as future research directions, are provided in Section \ref{sec:conclusion}.

\section{Problem Statement}
\label{sec: problem statement}

In this section, we present the multi-agent robust optimization problem to be addressed in this work. Initially, we familiarize the reader with the notation used throughout this paper. Subsequently, we present the basic multi-agent problem setup and the dynamics of the agents. The two main characterizations of underlying uncertainty, i.e., deterministic and stochastic, are then formulated. Then, we specify the control policies of the agents and the robust constraints that are considered in our framework. Finally, the exact formulation of the multi-agent problem is provided.

\subsection{Notation}
The integer set $[a,b] \cap \Zb$ is denoted with $\llbracket a, b \rrbracket$.
The space of matrices $\vX \in \Rb^{n \times n}$ that are symmetric positive (semi)-definite, i.e., $\vX \succ 0$ ($\vX \succeq 0$) is denoted with $\Sb_n^{++}$ ($\Sb_n^{+}$). The inner product of two vectors $\bx, \by \in \Rb^n$ is denoted with $\langle \bx, \by \rangle = \bx\T \by$.
% = \sum_{i=1}^n x_i y_i$. 
The $\ell_2$-norm of a vector 
$\bx = 
\begin{bmatrix}
x_1 & \dots & x_n
\end{bmatrix} 
\in \Rb^n, 
$ is defined as $\| \bx \|_2 = \sqrt{\langle \bx, \bx \rangle}$.
% = \sqrt{\sum_{i=1}^n x_i^2}$. 
The weighted norm $\| \bx \|_{\vQ} = \| \bQ^{1/2} \bx \|_2 = \sqrt{\bx\T \bQ \bx}$ is also defined for any $\vQ \in \Sb_n^{++}$. In addition, the Frobenius norm of a matrix $\vX \in \Rb^{m \times n}$ 
% = [x_{ij}] {\substack{{i \in \llbracket 1, m \rrbracket} \\j \in \llbracket 1,n \rrbracket}}$ 
%_{i \in \llbracket 1, m \rrbracket, j \in \llbracket 1,n \rrbracket} $ 
is given by $\| \vX \|_F = \sqrt{\tr(\vX\T \vX)}$. 
% = \sqrt{\sum_{i = 1}^m \sum_{j=1}^n X_{x_ij}^2}$.
%
With $[\bx_1 ; \dots; \bx_n]$, we denote the vertical concatenation of a series of vectors $\bx_1, \dots, \bx_n$. Furthermore, the probability of an event $\calA$ is denoted with $\Pb(\calA)$. Given a random variable (r.v.) $\bx \in \Rb^n$, its expectation and covariance are denoted with $\Eb[x] \in \Rb^n$ and $\Cov(\bx) \in \Sb_n^{+}$, respectively. 
The cardinality of a set $\calX$ is denoted as $n(\calX)$. 
Finally, given a set $\calX$, the indicator function $\calI_{\calX}$ is defined as $\calI_{\calX}(x) = 0$, if $x \in \calX$, and $\calI_{\calX}(x) = + \infty$, otherwise.

\subsection{Problem Setup}
Let us consider a multi-agent system of $N$ agents defined by the set $\calV = \{ 1, \dots, N \}$. All agents $i \in \calV$ may be subject to diverse dynamics, inter-agent interactions, and exogenous deterministic and stochastic uncertainty. Each agent $i \in \calV$ has a set of neighbors defined by $\calN_i \subseteq \calV$. Furthermore, for each agent $i \in \calV$, we define the set of agents that consider $i$ as a neighbor through $\calP_i = \{ j \in \calV | i \in \calN_j \}$. 
% To characterize the system of agents, we define the following two sets for each agent $i$ - 'neighbor' set $\calN_i$ and 'neighbor to' set $\calP_i$. The set $\calN_i$ of an agent $i$ consists of all the agents which are neighbor agents to 'i', while the set $\calP_i$ consists of all the agent to which the agent $i$ is a neighbor. 
% In the following sections, we define the dynamics, uncertainty set, and the control policy for each agent. 
%
\subsection{Dynamics}
We consider the following discrete-time linear time-varying dynamics for each agent $i \in \calV$, 
\begin{equation}    
x_{k+1}^i = A_k^i x_k^i + B_k^i u_k^i + C_k^i d_k^i + D_k^i w_k^i, ~ k \in \llbracket 0, T-1 \rrbracket, 
\label{dynamics}
\end{equation}
where $x_k^i \in \Rb^{n_{x_i}}$ is the state, $u_k^i \in \Rb^{n_{u_i}}$ is the control input of agent $i$ at time $k$, and $T$ is the time horizon. The terms $d_k^i \in \Rb^{n_{d_i}}$ represent \textit{exogenous deterministic} disturbances, while the terms $w_i^k \in \Rb^{n_{w_i}}$ refer to \textit{stochastic disturbances}, with their exact mathematical representations presented in Section \ref{sec: char of unc}. Each initial state $x_0^i, ~ i \in \calV$, is of the following form
\begin{equation}    
x_0^i = \bar{x}_0^i + \bar{d}_0^i + s_0^i,
\end{equation}
where $\bar{x}_0^i$ is the known part, while $\bar{d}_0^i$ and $s_0^i$ refer to the unknown exogenous deterministic and stochastic parts, respectively. Finally, $A_k^i, B_k^i, C_k^i, D_k^i$ are the dynamics matrices of appropriate dimensions. 

For convenience, let us also define the sequences 
$\bx^i = [ x_0^i; \dots; x_T^i ] \in \Rb^{(T+1) n_{x_i}}$, 
$\bu^i = [u_0^i; \dots; u_{T-1}^i] $ $\in \Rb^{T n_{u_i}}$, 
$\bw^i = [s_0^i; w_0^i; \dots; w_{T-1}^i] \in \Rb^{n_{x_i} + T n_{w_i}}$
and
$\boldsymbol{\zeta}^i = [\bar{d}_0^i; d_0^i; \dots; d_{T-1}^i] \in \Rb^{n_{x_i} + T n_{d_i}}$.
The dynamics \eqref{dynamics} can then be rewritten in a more compact form as 
\begin{equation}
\bx^i = \vG_0^i \bar{x}_0^i + \vG_u^i \bu^i + \vG_{\bw}^i \boldsymbol{\bw}^i + \vG_{\zeta}^i \boldsymbol{\zeta}^i,
\label{compact dynamics}
\end{equation} 
where the matrices $\vG_0^i, \vG_u^i, \vG_{\bw}^i, \vG_{\zeta}^i$ are provided in the Appendix-A.
% \begin{align}
% &
% \vG_0^i = 
% \begin{bmatrix}
% \vI; & 
% \vPhi^i(1,0) ; & 
% \vPhi^i(2,0) ; &
% \dots ; &
% \vPhi^i(T,0)
% \end{bmatrix} \nonumber
% %\in \Rb^{(T+1) n_x \times n_x }, 
% \\
% &
% \vG_u^i =  \vG_f(\{ \vB^i_t \}_{t=0}^{T-1}), \quad
% \vG_{\epsilon}^i = [\vG_0^i, \vG_f^i(\{ \vC^i_t \}_{t=0}^{T-1})] 
% \nonumber \\
% &
% \vG_{\zeta}^i = [\vG_0^i, \vG_f(\{ D_k^i \}_{t=0}^{T-1})]
% \nonumber \\
% &
% \vG_f^i(\{ \vY_t \}_{t=0}^{T-1}) =
% \begin{bmatrix}
% \vzero & \vzero & \dots & \vzero \\ 
% \vY_0 & \vzero & \dots & \vzero \\ 
% \vPhi^i(2,1) \vY_0 & \vY_1 & \dots & \vzero \\ 
% \vdots & \vdots & \vdots & \vdots\\ 
% \vPhi^i(T,1) \vY_0 & \vPhi^i(T,2) \vY_1 & \dots & \vY_{T-1}
% \end{bmatrix} \nonumber \\
% &
% \vPhi^i(t_1,t_2) = \vA_{t_1 -1} \vA_{t_1-2} \dots \vA_{t_2} \;
% \text{for} \; t_1 > t_2, \; \vPhi^i(t_1,t_1) = \vI \nonumber
% \end{align}
%
% We also define $\vP_t \in \Rb^{n_x \times (T+1) n_x}$ such that $x_k^i = \vP_t \bx^i$.
% 
\subsection{Characterizations of Uncertainty}
\label{sec: char of unc}
As previously mentioned, we consider the existence of both exogenous deterministic and stochastic uncertainties. These two types of uncertainty are formally described as follows.
\paragraph{Exogenous deterministic uncertainty} 
 The unknown deterministic disturbances $\boldsymbol{\zeta}^i$ typically stem from external factors that cannot be accurately represented through stochastic signals. This creates a requirement to formulate optimization problems and achieve optimal policies that are feasible in a \textit{robust} sense for all possible values of such disturbances in a bounded \textit{uncertainty set}. 
 % Examples of such uncertainty sets include ellipsoids, polytopes, and others \cite{RobustNemirovski}.
 In this work, we assume the sequence $\boldsymbol{\zeta}^i$ to be lying inside an ellipsoidal uncertainty set defined as follows
\begin{align}
\calU_i[\tau^i] = 
\big\{
\boldsymbol{\zeta}^i \in \Rb^{n_{\zeta_i}} &| 
\;	\exists (\bz_i \in \Rb^{\bar{n}_i}, \tau^i \in \Rb): 
\; 
\\ 
& \boldsymbol{\zeta}^i = \boldsymbol{\Gamma}_i \bz_i, 
\nonumber
\; \langle \vS_i \bz_i, \bz_i \rangle \leq \tau^i
\big\}, 
\label{ellipsoid uncertainty set}
\end{align}
where $n_{\zeta_i} = n_{x_i} + T n_{d_i}$, $\boldsymbol{\Gamma}_i \in \Rb^{n_{\zeta_i} \times \bar{n}_i}$, $\vS_i \in \Sb_{++}^{\bar{n}_i}$ and $\tau^i >0$ is the uncertainty level.
Ellipsoidal sets are used to model uncertainty in various robust control applications \cite{Petersen2000}. Further, ellipsoid sets have simple geometric structure, and more complex sets can be approximated using ellipsoids.
\paragraph{Stochastic uncertainty} The unknown stochastic component $\bw^i$ is assumed to be a  Gaussian random vector with mean $\bmu_{{\bw}^i} = \Eb[\bw^i] = 0$, and covariance $\bSigma_{\boldsymbol{\bw}^i} = \Cov[\bw^i]$.

\subsection{Control Policies}
Towards addressing both types of uncertainty, we consider affine \textit{purified state} feedback policies \cite{RobustNemirovski, kotsalis2020convex} for all agents. Let us first introduce the disturbance-free states $\hat{x}_k^i$, whose dynamics are given by
\begin{align}
\hat{x}_{k+1}^i & = A_k^i \hat{x}_t^i + B_k^i u_k^i, \quad k \in \llbracket 0,T-1 \rrbracket,
\\
\hat{x}_0^i & = \bar{x}_0^i.
\end{align}
Subsequently, we can define the purified states $\delta_k^i = x_k^i - \hat{x}_k^i$, whose dynamics are given by
\begin{equation}
\begin{aligned}
\delta_{k+1}^i & = A_k^i \delta_k^i + C_k^i d_k^i + D_k^i w_k^i \quad k \in \llbracket 0,T-1 \rrbracket, \\
\delta_0^i & = \bar{d}_0^i + s_0^i,
\end{aligned}
\end{equation}
or more compactly by 
\begin{equation}
\bdelta^i = \vG_{\bw}^i \boldsymbol{\bw}^i + \vG_{\zeta}^i \boldsymbol{\zeta}^i, 
\end{equation}
where $\bdelta^i = \begin{bmatrix}
\bdelta_0^i; & \dots; & \bdelta_T^i
\end{bmatrix}$. 
We consider the following affine purified state feedback control policy for each agent $i$,
\begin{equation}
u^i_k = \bar{u}^i_k + \sum_{\ell =  k - \gamma_h +1}^k K^i_{k,\ell} \delta^i_\ell, \; 
\forall \; k \in \llbracket 0,T-1 \rrbracket,
\end{equation}
where $\bar{u}^i_k \in \Rb^{n_{u_i}}$ are the feed-forward control inputs, and $K^i_{k,\ell} \in \Rb^{n_{u_i} \times n_{x_i}}$ are feedback gains on the purified states. Further, $\gamma_h$ defines the length of the history interval considered for the feedback. The above policies can be written in compact form (when $\gamma_h = T$) as
\begin{equation}
\bu^i = \bar{\bu}^i + \vK^i \bdelta^i,
\label{control expression}
\end{equation}
where 
\begin{align}
\bar{\bu}^i = \begin{bmatrix}
\bar{u}_0^i; & \dots; & \bar{u}_{T-1}^i
\end{bmatrix}, 
\nonumber
\end{align}
\begin{align}
\vK^i = 
\begin{bmatrix}
K_{0,0}^i & \vzero & \dots & \vzero & \vzero \\ 
K_{1,0}^i & K_{1,1}^i & \dots & \vzero & \vzero \\  
\vdots & \vdots & \vdots & \vdots & \vdots\\ 
K_{T-1,0}^i & K_{T-1,1}^i & \dots & K_{T-1, T-1}^i & \vzero
\end{bmatrix}.
\nonumber
\end{align}
\subsection{Robust Constraints}
Next, we introduce the types of constraints that are considered in this work. We highlight that the following constraints must be \textit{satisfied robustly} for \textit{all possible values} of $\bzeta^i$ lying inside the uncertainty sets $\calU_i, \ i \in \calV$.  

\paragraph{Robust linear mean constraints} The first class of constraints we consider are robust linear constraints on the expectation of the state sequence $\bx^i$ of agent $i \in \calV$,
\begin{equation}
\vA_i \Eb[\bx^i] \leq \bm{b}_i, \quad \forall \bzeta^i \in \calU_i,
\label{Robust Linear Constraints}
\end{equation}
where $\vA_i \in \Rb^{m \times  (T+1)}n_{x_i}$, $\bm{b}_i \in \Rb^{m}$.
In a robotics setting, such constraints could typically represent bounds on the mean position, velocity, etc.

\paragraph{Robust nonconvex norm-of-mean constraints} Furthermore, we consider nonconvex inequality constraints of the form 

\begin{equation}
\| \vA_i \Eb[x_k^i] - \bm{b}_i \|_2 \geq c_i, \quad \forall \bzeta^i \in \calU_i,
\label{Robust nonconvex norm-of-mean constraints}
\end{equation}
where $\vA_i \in \Rb^{m \times n_{x_i}}$, $\bm{b}_i \in \Rb^m$, $c_i \in \Rb^+$. Such constraints are typically useful for capturing avoiding obstacles, forbidden regions, etc. In addition, we extend to the inter-agent version of these constraints which is given by
\begin{equation}
\| \vA_i \Eb[\bx^i_k] - \vA_j \Eb[\bx^j_k] \|_2 \geq c_{ij}, \quad \forall \bzeta^i \in \calU_i, ~ \bzeta^j \in \calU_j, ~ j \in \calN_i
\label{robust inter-agent collision avoidance}
\end{equation}
where $\vA_i \in \Rb^{m \times n_{x_i}}$, $\vA_j \in \Rb^{m \times n_{x_j}}$, $c_{ij} \in \Rb^+$. This class of constraints typically represents collision avoidance constraints between different robots.

\paragraph{Robust linear chance constraints} Subsequently, we consider robust chance constraints on the states of the agents,
\begin{equation}
\Pb (\ \ba_i \T \bx^i  > b_i ) \leq p, 
\quad \forall \bzeta ^i \in \calU_i,
\label{Robust Chance Constraints}
\end{equation} 
 where $\ba_i \in \Rb^{(T+1)n_{x_i}}$, 
 $b_i \in \Rb$, $p$ is the probability threshold. It should be noted that these constraints are also affected by the stochastic noise since those are probabilistic constraints on the entire state and not only on the state mean.

\paragraph{Robust covariance constraints}
%
% %
Additionally, we consider the following constraints on the covariance matrices of the state sequences,
\begin{equation}
    \Cov(\bx^i) \preceq \bSigma^i, \quad \forall \bzeta^i \in \calU_i,
    \label{Covariance Constraints}
\end{equation}
where $\bSigma^i$ is a prespecified allowed  upper bound on the state covariance.
\subsection{Problem Formulation}
Finally, we present the mathematical problem formulation for the multi-agent robust optimization problem under both stochastic and bounded deterministic uncertainties. 
\begin{problem}[Multi-Agent Robust Optimization Problem]
\label{initial problem}
For all agents $i \in \calV$, find the robust optimal policies $\bar{\bu}^i,  \vK^i$, such that 
\begin{align}
&  {~~~~~~~~ } \{ \bar{\bu}^i, \vK^i \}_{i \in \calV} = \argmin
\sum_{i \in \calV} 
J_i(\bar{\bu}^i, \vK^i)
%\quad (= J (\bu^1, \bu^2, \dots \bu^N) ) 
\nonumber
\\[0.1cm]
\mathrm{s.t.} \; 
&~
x_{k+1}^i = A_k^i x_k^i + B_k^i u_k^i + C_k^i d_k^i + D_k^i w_k^i, ~ k \in \llbracket 0,T-1 \rrbracket, 
\nonumber
\\
&~
\bar{x}_0^i \text{ : given}, 
  \label{robust_optimization_problem_1}
 \\ 
&
\eqref{Robust Linear Constraints}  - \eqref{Covariance Constraints}  \nonumber 
\end{align}  
where each cost $J_i(\bar{\bu}^i, \vK^i) = \bar{\bu}^i{}\T \vR_{\bar{u}} \bar{\bu}^i 
+ || \vR_K \vK^i||_F^2$ penalizes the control effort of agent $i \in \calV$.
\end{problem}
\section{Distributed Robust Optimization under Deterministic and Stochastic Uncertainty}
\label{Distributed Optimization under deterministic representations of uncertainty}
In this section, we propose a distributed robust optimization framework for solving the multi-agent problem presented in Section \ref{sec: problem statement}. 
Initially, we provide computationally tractable approximations for the robust constraints in Section \ref{sec: distr section tract subsec}. Subsequently, Section \ref{sec: distr section problem transformation subsec} presents the new transformed problem to be solved based on the constraint approximations. Finally, in Section \ref{sec: distr section distr alg subsec}, we present a distributed algorithm for solving the transformed problem in a scalable manner.

\subsection{Reformulation of Semi-infinite Constraints}
\label{sec: distr section tract subsec}
We start by analyzing the effect of the disturbances on the system dynamics. In particular, since the robust constraints \eqref{Robust Linear Constraints} \ref{robust inter-agent collision avoidance}) \eqref{robust inter-agent collision avoidance} depend on $\Eb[\bx^i]$, let us first examine the expression for $\Eb[\bx^i]$ given by
\begin{align}
    & \Eb[\bx^i] = \vG_0^i \bar{\bx}_0^i + \vG_u^i \bar{\bu}^i + (\vG_u^i \vK^i + \vI) \vG_{\zeta}^i \boldsymbol{\zeta}^i {,
    } \label{mean_constraint_1}
\end{align}
since $\Eb[\bw^i] = 0$ (see Appendix-B). It is important to emphasize that the mean state relies only on the deterministic - and not the stochastic -  disturbances. For convenience, let us also define 
\begin{equation}
\bmu_{x,\bar{u}}^i = \vG_0^i \bar{\bx}_0^i + \vG_u^i \bar{\bu}^i, 
\quad
\vM_i =  (\vG_u^i \vK^i + \vI) \vG_{\zeta}^i,
\end{equation}
%DIF > 
as well as the matrix 
$\vP_k^i$ such that $x_k^i = \vP_k^i \bx^i$, $\tilde{\vM}_k^i = \vP^i_k \vM_i$, and $\bmu_{x_k,\bar{u}}^i = \vP^i_k \bmu_{x,\bar{u}}^i$.
Subsequently, we proceed with providing tractable approximations for the robust constraints presented in Section \ref{sec: problem statement}.
\subsubsection{Robust linear mean constraints} 
Let us consider a single constraint from the set of linear constraints \eqref{Robust Linear Constraints}, given by
\begin{align}
    \ba_i \T \Eb[\bx^i] \leq b_i, \quad \forall \bzeta^i \in \calU_i.
\end{align}
Note that the above constraint can be reformulated as
\begin{align}
    \max_{\bzeta^i \in \calU_i} \ba_i \T \Eb[\bx^i] \leq b_i {. 
    } \label{single mean constraint max}
\end{align}
The following proposition provides the exact optimal value of the maximization problem in the LHS of \eqref{single mean constraint max}.
%DIF > 
\begin{proposition} \label{ Max bound on mean} 
The maximum value of $\ba_i \T \Eb[\bx^i]$ when the uncertainty vector $\bzeta^i$ lies in the set $\calU_i$ is given by
\begin{equation}
    \max_{\bzeta^i \in \calU_i} \ba_i \T \Eb[\bx^i] =
    % (\tau^i)^{1/2} || \vP^i \T \vG_{\zeta}^i \T (\vG_u^i \vK^i + \vI) \T \ba||_{\vS_i^{-1}}
    \ba_i \T \bmu_{x,\bar{u}}^i + (\tau^i)^{1/2} || \boldsymbol{\Gamma}_i \T \vM_i \T \ba_i||_{\vS_i^{-1}}.
    \label{max bound on mean problem}
\end{equation}
\end{proposition}
\begin{proof}
    Provided in Appendix-D.
\end{proof}
% Using the proposition (\ref{ Max bound on mean}), the bounds on $\langle \; \ba, \bmu_{\bx}^i \; \rangle$ are given as follows -
% \begin{align}
%     & \forall \; \boldsymbol{\zeta}^i \in \calU_i[\tau^i] 
%     \label{mean_bounds_1}\\
%     &~~
%     \langle \; \ba, \bmu_{x,\bar{u}}^i \; \rangle - \bmu_{x, a}^i
%     \leq \langle \; \ba, \bmu_{\bx}^i \; \rangle \leq
%     \langle \; \ba, \bmu_{x,\bar{u}}^i \; \rangle + \bmu_{x, a}^i  \nonumber
% \end{align}
% %
% where $\bmu_{x, a}^i = (\tau^i)^{1/2} || \vP^i \T \vG_{\zeta}^i \T (\vG_u^i \vK^i + \vI) \T \ba||_{\vS_i^{-1}}$. \\
%
Using Proposition \ref{ Max bound on mean}, an equivalent tractable version of the constraint \eqref{single mean constraint max} can be given as follows
\begin{equation}
    \ba_i \T \bmu_{x,\bar{u}}^i + (\tau^i)^{1/2} || \boldsymbol{\Gamma}_i \T \vM_i \T \ba_i||_{\vS_i^{-1}}
    \leq b_i.
    \label{mean_constraint_form1}
\end{equation}
Similar to the maximum value case, the minimum value can be provided as
\begin{align}
    & \min_{\bzeta^i \in \calU_i} \ba_i \T \Eb[\bx^i] = \ba_i \T \bmu_{x,\bar{u}}^i - (\tau^i)^{1/2} || \boldsymbol{\Gamma}_i \T \vM_i \T \ba_i||_{\vS_i^{-1}}.
\end{align}
Based on the above observations, we introduce the following additional form of robust linear mean constraints
\begin{equation}
\begin{aligned}
     \bar{\ba}_i \T \bmu_{x,\bar{u}}^i = \bar{b}_i,
    \quad (\tau^i)^{1/2} || \boldsymbol{\Gamma}_i \T \vM_i \T \bar{\ba}_i||_{\vS_i^{-1}} \leq \epsilon_i,
    \label{New mean formulation}
\end{aligned}   
\end{equation}
with $\bar{\ba}_i \in \Rb^{ (T+1)} n_{x_i}, \bar{b}_i \in \Rb$ and $\epsilon_i \in \Rb^+$, which would imply the following constraints
\begin{align}
 \bar{\ba}_i \T \Eb[\bx^i] \leq \bar{b}_i + \epsilon_i, 
 \quad
 \bar{\ba}_i \T \Eb[\bx^i] \geq \bar{b}_i - \epsilon_i, \quad \forall \bzeta^i \in \calU_i.   
\end{align}
%DIF > 
The constraint form \eqref{New mean formulation} allows us to decouple the mean constraint in terms of $\bar{\bu}^i$ and $\vK^i$, such that the expectation of the disturbance-free state $\bmu_{x,\bar{u}}^i$ can be controlled using the feed-forward control $\bar{\bu}^i$, and the expectation of the state mean component $\vM_i \bzeta^i$ dependent on the uncertainty can be controlled using the feedback control gain $\vK^i$.
\subsubsection{Robust nonconvex norm-of-mean constraints}
Subsequently, we consider the robust nonlinear constraint \eqref{Robust nonconvex norm-of-mean constraints}.
We present two approaches for deriving the equivalent/approximate tractable constraints to the above constraint.
\paragraph{Initial Approach}\label{initial approach}
The robust constraint in \eqref{Robust nonconvex norm-of-mean constraints} can be rewritten as 
\begin{align}
     \min_{\bzeta^i \in \calU_i} \| \vA_i \Eb[x_k^i] - \bm{b}_i \|_2 \geq c_i.
\end{align}
Although the problem in the LHS is convex, the equivalent tractable constraints to the above, obtained by applying S-lemma \mbox{
\cite{kotsalis2020convex}}\hskip0pt
, might not be convex due to the non-convexity of the overall constraint (details can be referenced from Appendix-E). Thus, we first linearize the constraint \eqref{Robust nonconvex norm-of-mean constraints} with respect to control parameters 
$\Bar{\bu}^i, \vK^i$
around the nominal values $\Bar{\bu}^{i,l}, \vK^{i,l}$. 
Through this approach, we obtain the following constraint
\begin{equation}    
\begin{aligned}
 \boldsymbol{\zeta}^i {}\T \mathcal{Q} ( \vK^i, \vK^{i,l})
\boldsymbol{\zeta}^i 
+ 2 & \bar{\mathcal{Q}} ( \vK^i, \vK^{i,l}, \Bar{\bu}^i, \Bar{\bu}^{i,l})  \boldsymbol{\zeta}^i \\
& + q(\Bar{\bu}^i, \Bar{\bu}^{i,l})
\geq c_i^2, \quad \forall \; \boldsymbol{\zeta}^i \in \calU_i,
\end{aligned}
\end{equation}
%DIF > 
where $\mathcal{Q} ( \vK^i, \vK^{i,l}), \bar{\mathcal{Q}} ( \vK^i, \vK^{i,l}, \Bar{\bu}^i, \Bar{\bu}^{i,l})$ and $q(\Bar{\bu}^i, \Bar{\bu}^{i,l})$ are provided in Appendix-E.
The equivalent tractable constraints to the above constraint can be given as \cite{kotsalis2020convex} -
\begin{equation}
    \begin{aligned}
    & \beta \geq 0, \\
    &
    \begin{bmatrix}
        \boldsymbol{\Gamma}_i \T \mathcal{Q} ( \vK^i, \vK^{i,l}) \boldsymbol{\Gamma}_i 
        + \beta \vS_i 
        & \boldsymbol{\Gamma}_i \T \bar{\mathcal{Q}} ( \vK^i, \vK^{i,l}, \Bar{\bu}^i, \Bar{\bu}^{i,l}) \T  \\
        \bar{\mathcal{Q}} ( \vK^i, \vK^{i,l}, \Bar{\bu}^i, \Bar{\bu}^{i,l}) \boldsymbol{\Gamma}_i
        & q(\Bar{\bu}^i, \Bar{\bu}^{i,l}) - c_i^2 - \beta \tau^i
    \end{bmatrix}
     \succeq 0.
\end{aligned} 
\label{obstacle_semidef_old}
\end{equation}
The derived equivalent tractable constraints can be used in cases where the number of robust non-convex constraints of the defined form is relatively small. However, in applications that involve a significant number of such constraints (e.g., collision avoidance constraints), this formulation might prove to be significantly expensive computationally. For instance, consider an obstacle avoidance scenario, where each agent $i$ needs to enforce the constraints of the above form for each time step. Then, each agent $i$ in an environment with $n_{obs}$ obstacles would have $n_{obs}T$ number of constraints of the form \eqref{obstacle_semidef_old}. Each constraint of the form \eqref{obstacle_semidef_old} involves a matrix of size $\bar{n}_i + 1$ (when $\boldsymbol{\Gamma}_i = I$,  $\bar{n}_i = n_{x_i} + T n_{d_i} $ ). Thus, the computational cost of the above formulation increases exponentially with the number of obstacles, time horizon, and the dimension of the disturbances  {(see Section \ref{Computational Complexity Analysis} for a rigorous complexity analysis)}. In order to overcome this issue, we propose a second approach for deriving equivalent/approximate tractable versions of the constraint \eqref{Robust nonconvex norm-of-mean constraints}.
\paragraph{Reducing computational complexity}
\label{new reformulation}
Let us first write the constraint \eqref{Robust nonconvex norm-of-mean constraints} in terms of $\bzeta^i$ as follows -
\begin{align}
    \| \vA_i \bmu_{x_k,\bar{u}}^i +  \vA_i \tilde{\vM}_k^i \bzeta^i - \bm{b}_i \|_2 
    \geq c_i, \quad \forall \bzeta^i \in \calU_i.
\end{align}
Based on the triangle inequality, we consider the following tighter approximation of the above constraint  
\begin{align}
    \| \vA_i \bmu_{x_k,\bar{u}}^i - \bm{b}_i \|_2 
    - \| \vA_i \tilde{\vM}_k^i \bzeta^i \|_2
    \geq c_i, \quad \forall \bzeta^i \in \calU_i
\end{align}
which can be rewritten as 
\begin{align}
    \| \vA_i \bmu_{x_k,\bar{u}}^i - \bm{b}_i \|_2 \geq
    c_i + \max_{\bzeta^i \in \calU_i} \| \vA_i \tilde{\vM}_k^i \bzeta^i \|_2.
\end{align}
%DIF > 
Let us now introduce a slack variable $\tilde{c}_i$ such that we split the above constraint into the following system of two constraints %
\begin{align}
    & \| \vA_i \bmu_{x_k,\bar{u}}^i - \bm{b}_i \|_2 
    \geq  \tilde{c}_i,
    \label{nonconvex obstacle final ubar}\\
    & 
    \max_{\bzeta^i \in \calU_i} \| \vA_i \tilde{\vM}_k^i \bzeta^i \|_2 
    \leq \tilde{c}_i 
    - c_i
   \label{nonconvex obstacle K1}
\end{align}
Note, however, that the constraint \eqref{nonconvex obstacle final ubar}  is non-convex. We address this by linearizing the constraint around the nominal trajectory $\Bar{\bu}^{i,l}$.
In addition, the constraint \eqref{nonconvex obstacle K1} is still a semi-infinite constraint; thus, we need to derive its equivalent tractable constraints. Since the LHS in the constraint \eqref{nonconvex obstacle K1} involves a non-convex optimization problem, deriving the equivalent tractable constraints is NP-hard. Hence, we derive tighter approximations of the constraint \eqref{nonconvex obstacle K1} in the following proposition. 
\begin{proposition}
    The tighter approximate constraints to the semi-infinite constraint \eqref{nonconvex obstacle K1} are given by  the following second-order cone constraints (SOCP constraints)
\begin{align}
        &
         {\| \bmu_{d,k}^i \|_2 \leq \tilde{c}_i - c_i
        } \label{semi-definite obstacle new} \\
        &
        (\bmu_{d,k}^i)_{\bar{m}}\geq  
        (\tau^i)^{1/2} || \boldsymbol{\Gamma}_i \T \vM_i \T h_{k,\bar{m}}^i||_{\vS_i^{-1}}
        \label{mu_d expression}
\end{align}
where $h_{k,\bar{m}}^i {}\T $ is $\bar{m}^{th}$ row of $\vA_i \vP_k^i$.
\end{proposition}
\begin{proof}
    Provided in Appendix-F.
\end{proof}
In this formulation, only the constraints \eqref{mu_d expression} include $\vS_i$. This formulation is useful in the case where there are multiple non-convex constraints of the form \eqref{Robust nonconvex norm-of-mean constraints} at $x_k^i$, which have the same $\vA_i$ matrix. In such a case, we only need to enforce the constraint \eqref{mu_d expression} once per each unique $\vA_i$ matrix in the constraints. 
For instance, let us again consider the collision avoidance example. Since the collision constraints (both obstacle and inter-agent) are applied to the agent's position, the matrix $\vA_i$ remains the same in all such constraints, irrespective of the number of obstacles. Therefore, it is only the amount of constraints of the form \eqref{semi-definite obstacle new} that would increase with the number of obstacles. Further, the geometrical interpretation of the tighter approximations proposed for a 2D case can be found in Appendix-H.
 
Subsequently, we proceed with considering the inter-agent robust non-convex constraints \eqref{robust inter-agent collision avoidance},
 %
% \begin{equation}
% \| \vA_i \Eb[\bx^i_k] - \vA_j \Eb[\bx^j_k] \|_2 \geq c_{ij}, \quad \forall \bzeta^i \in \calU_i, ~ \bzeta^j \in \calU_j, ~ j \in \calN_i
% \nonumber
% \end{equation}
%
which can be rewritten as 
\begin{align}
    \| \tilde{\vA}_{ij} \Eb[ \Tilde{\bx}^{ij}_k] \|_2 \geq c_{ij}, \quad \forall \bzeta^i \in \calU_i, ~ \bzeta^j \in \calU_j, ~ j \in \calN_i {,
    } \label{nonconvex interagent concatenated constraint}
\end{align}
where $\tilde{\vA}_{ij} = \begin{bmatrix}
    \vA_i,& -\vA_j
\end{bmatrix}$, $\Tilde{\bx}^{ij}_k = \begin{bmatrix}
    \bx^i_k; & \bx^j_k
\end{bmatrix}$. The uncertainty involved for the variable $\Tilde{\bx}^{ij}$ would be $\Tilde{\bzeta}^{ij} = \begin{bmatrix}
    \bzeta^i; & \bzeta^j
\end{bmatrix}$. Further, it should be noted that if $\bzeta^i, \bzeta^j$ belong to the defined ellipsoidal sets $\calU_i$, $\calU_j$ respectively, then the uncertainty vector $\Tilde{\bzeta}^{ij}$ can also be modeled by an ellitope (which is an intersection of ellipsoids). Thus, by using the initial approach, the constraint \eqref{nonconvex interagent concatenated constraint} would result in semi-definite constraints that are of the size $\bar{n}_i + \bar{n}_j + 1$ \mbox{
\cite{kotsalis2020convex}}\hskip0pt. This might be particularly disadvantageous in a large-scale multi-agent setup, which might involve a significant number of neighbor agents. Thus, we derive equivalent/approximate tractable constraints using the second approach. 

For that, let us first rewrite a single constraint corresponding to a neighbor agent $j$ of the agent $i$ ($j \in \calN_i$) from the set of constraints \eqref{robust inter-agent collision avoidance} as follows.
\begin{equation}
\begin{aligned}
    \forall \bzeta^i & \in \calU_i, ~ \bzeta^j \in \calU_j,  \\
    &
    \| \vA_i \bmu_{x_k,\bar{u}}^i 
    - \vA_j \bmu_{x_k,\bar{u}}^j
    + \vA_i \tilde{\vM}_k^i \bzeta^i
    - &  \vA_j \tilde{\vM}_k^j \bzeta^j \|_2 
    \geq c_{ij}. \nonumber
\end{aligned}
\end{equation}
We then consider the following tighter approximation to the above 
constraint-
\begin{displaymath}
\begin{aligned}
    \forall \bzeta^i & \in \calU_i, ~ \bzeta^j \in \calU_j,   \\
    &
    \begin{aligned}
    \| \vA_i \bmu_{x_k,\bar{u}}^i 
    - \vA_j \bmu_{x_k,\bar{u}}^j \|_2
    - \| \vA_i \tilde{\vM}_k^i \bzeta^i
    -  \vA_j \tilde{\vM}_k^j \bzeta^j \|_2  
    \geq c_{ij} 
    \end{aligned}
\end{aligned}
\end{displaymath}
and similarly, with the previous case, we introduce a slack variable $\tilde{c}_{ij}$, through which we can rewrite the above constraint as follows
\begin{align}
    & 
    \| \vA_i \bmu_{x_k,\bar{u}}^i 
    - \vA_j \bmu_{x_k,\bar{u}}^j \|_2
    \geq \tilde{c}_{ij} 
    \label{inter-agent collision avoidance ubar}\\
    &
    \begin{aligned}
         \max_{\bzeta^i \in \calU_i, ~ \bzeta^j \in \calU_j}
         \| \vA_i \tilde{\vM}_k^i \bzeta^i
    -  \vA_j \tilde{\vM}_k^j \bzeta^j \|_2  
        \leq \tilde{c}_{ij} - c_{ij}
    \end{aligned}    
    \label{interagent collision avoidance K1}
\end{align}
The non-convex constraint \eqref{inter-agent collision avoidance ubar} is addressed by linearizing it around 
the nominal value $\Bar{\bu}^{i,l}$. 
Similar to the previous case, the constraint \eqref{interagent collision avoidance K1} is semi-infinite, and deriving its equivalent tractable constraints is NP-hard. Thus, we derive the tighter approximations of the constraint \eqref{interagent collision avoidance K1}.
\begin{proposition}
    The tighter approximate constraints to the semi-infinite constraint \eqref{interagent collision avoidance K1} are given by the following  second-order cone constraints (SOCP)
    \begin{align}
        & 
        {\| \bmu{_{d,k}^i +  }\bmu_{d,k}^j \|_2
        \leq 
        \tilde{c}_{ij} - c_{ij},
        } \label{semi-definite interagent new} \\
        &
        (\bmu_{d,k}^i)_{\bar{m}} \geq  
        (\tau^i)^{1/2} || \boldsymbol{\Gamma}_i \T \vM_i \T h_{k,\bar{m}}^i||_{\vS_i^{-1}} {,
        } \label{interagent mu_d expression} \\
        &
        (\bmu_{d,k}^j)_{\bar{m}}  \geq  
        (\tau^j)^{1/2} || \boldsymbol{\Gamma}_j \T \vM_j \T h_{k,\bar{m}}^j||_{\vS_j^{-1}} {,
        }\label{interagent mu_d expression 2}
     \end{align}
where  {$h_{k,\bar{m}}^i {}\T, h_{k,\bar{m}}^j {}\T $ are the $\bar{m}^{th}$ } rows of the matrices $\vA_i \vP_k^i, \vA_j \vP_k^j$ respectively.
\end{proposition}
\begin{proof}
    Provided in Appendix-G.
\end{proof}
\noindent
It should be noted that the constraints \eqref{interagent mu_d expression}, \eqref{interagent mu_d expression 2} will be the same as \eqref{mu_d expression} if the $\vA_i$ matrix is the same (e.g., collision avoidance scenarios), therefore alleviating the need to introduce them as additional constraints in our framework. 
\subsubsection{Robust linear chance constraints}
% Let us refer to the following robust linear chance constraints (\ref{Robust Chance Constraints}).
% %
% \begin{align}
%     \Pb (\ \ba_i \T \bx^i  > b_i ) \leq p
%     \nonumber
% \end{align}
%
The equivalent tractable constraints to the constraint \eqref{Robust Chance Constraints} are given as \cite{kotsalis2020convex} 
\begin{align}
    & 
    \ba_i \T \bmu_{x,\bar{u}}^i
    + 
    (\tau^i)^{1/2} || \boldsymbol{\Gamma}_i \T \vM_i \T \ba_i ||_{\vS_i^{-1}} \leq b_i - \alpha 
    \label{chance constraints 1} \\
    &
    \eta || \ba \T (\vG_u^i \vK^i + \vI) \vG_{\bw}^i
        \boldsymbol{\varphi}^i ||_2 \leq \alpha 
    \label{chance constraints 2}
\end{align}
where $ \eta : \Pb_{y \in \calN(0,1)}(y \geq q) = p $, and   $\boldsymbol{\varphi}^i \boldsymbol{\varphi}^i {}\T = {\bSigma_{\bw^i}}$ .
 \vspace{0.1cm}
 \subsubsection{Robust covariance constraints}
%
% Let us refer to the following covariance constraint (\ref{Covariance Constraints})
% %
% \begin{align}
%     \Cov(\bx^i) \preceq \bSigma^i \nonumber
% \end{align}
%
The covariance of the state of an agent $i$ is given as follows 
 {(derivation can be referenced from Appendix-B)
}\begin{equation}
    \Cov[\bx^i] = (\vG_u^i \vK^i + \vI) \vG_{\bw}^i
    {\bSigma_{\bw^i}}
    \vG_{\bw}^i {}\T (\vG_u^i \vK^i + \vI) \T.
\end{equation} 
It should be observed that the covariance does not depend on the deterministic uncertainty but only on the covariance of the stochastic noise. Now, let us rewrite the constraint  {\eqref{Covariance Constraints} } in terms of $\vK^i$ as %
\begin{align}
    (\vG_u^i \vK^i + \vI) \vG_{\bw}^i
     {\bSigma_{\bw^i}
    } \vG_{\bw}^i  {} \T (\vG_u^i \vK^i + \vI)  {} \T  \preceq \bSigma^i.
    \label{covariance_constraint_1}
\end{align}
The above constraint is non-convex and can be reformulated into the following equivalent convex constraint by using the Schur complement,
    \begin{equation}
    \begin{bmatrix}
        \bSigma^i &  (\vG_u^i \vK^i + \vI) \vG_{\bw}^i
        \boldsymbol{\varphi}^i \\
        ((\vG_u^i \vK^i + \vI) \vG_{\bw}^i
        \boldsymbol{\varphi}^i) \T & \vI_{(n_x + N n_w)}
    \end{bmatrix} \succeq 0 \label{covariance_constraint_convex}
\end{equation}
where $\boldsymbol{\varphi}^i \boldsymbol{\varphi}^i {}\T = {\bSigma_{\bw^i}}$  . 
\subsection{Problem Transformation}
 \label{sec: distr section problem transformation subsec}
Now that we have derived the computationally tractable constraints that could be applied to a multi-robot setting, let us define the multi-agent optimization problem using the reformulated constraints. 
\begin{problem}[Multi-Agent Robust Optimization Problem II]\label{Transformed Multi-agent robust optimization problem}
For all agents $i \in \calV$, find the robust optimal policies $\bar{\bu}^i,  \vK^i$, such that 
\begin{align}
& \{ \bar{\bu}^i, \vK^i \}_{i \in \calV} = \argmin
\sum_{i \in \calV} 
J_i(\bar{\bu}^i, \vK^i)
%\quad (= J (\bu^1, \bu^2, \dots \bu^N) ) 
\nonumber
\\[0.1cm]
\mathrm{s.t.} \; 
&~
x_{k+1}^i = A_k^i x_k^i + B_k^i u_k^i + C_k^i d_k^i + D_k^i w_k^i, ~ k \in \llbracket 0,T-1 \rrbracket, 
\nonumber
\\
&~
\bar{x}_0^i \text{ : given}, 
\label{robust_optimization_problem_2}
\\ 
&
\eqref{mean_constraint_form1}, \eqref{New mean formulation}, \eqref{nonconvex obstacle final ubar}, \eqref{semi-definite obstacle new}, \eqref{mu_d expression},  \eqref{inter-agent collision avoidance ubar}, \eqref{semi-definite interagent new},
\eqref{interagent mu_d expression}, \eqref{chance constraints 1}, \eqref{chance constraints 2}, \eqref{covariance_constraint_convex}.  \nonumber 
\end{align}  
%
% where the costs $J_i(\bar{\bu}^i, \vK^i) = \bar{\bu}^i{}\T \vR_{\bar{u}} \bar{\bu}^i 
% + || \vR_K \vK^i||_F^2$ penalize the control effort of each agent $i \in \calV$.
\end{problem}
We are interested in solving the above problem in a multi-robot setting. Hence, we consider the robust non-convex constraints to represent obstacle and inter-agent collision avoidance constraints and are applied at each time step $k$. As mentioned in the previous subsection, all the collision avoidance constraints involve the same $\vA_i$ matrix in the constraint. Thus, in the subsequent sections, we consider $\vA_i$ in the constraints \eqref{nonconvex obstacle final ubar}, \eqref{semi-definite obstacle new}, \eqref{mu_d expression}, \eqref{inter-agent collision avoidance ubar}, \eqref{semi-definite interagent new}, \eqref{interagent mu_d expression} to be $\vH_{pos}$. Further, in this case, the constraints \eqref{mu_d expression} and \eqref{interagent mu_d expression} are the same. %
\subsection{Distributed Algorithm}
\label{sec: distr section distr alg subsec}
Now that we have formulated the robust multi-agent trajectory optimization problem \ref{Transformed Multi-agent robust optimization problem}, we provide a distributed architecture for solving it.
The introduction of the variables $\{ \bmu_{d,k}^i \}_{i \in \calV}$ helps to characterize the state mean of the agents, which is not a point but rather a set of points. This allows us to develop a distributed framework where the agents need not share the control parameters, nor have any information of the other agents. 
%{\color{SkyBlue} cite DistributedCS}
\subsubsection{Problem Setup}
% The constraints (\ref{Robust Non-Linear Constraints_inter_agent_mu_ubar}) and  (\ref{Robust Non-Linear Constraints_inter_agent_mu_d}) are the coupling constraints between the agents. 
In this subsection, we provide an overview of steps taken to reformulate the problem \ref{Transformed Multi-agent robust optimization problem} into a structure that would allow us to solve it in a distributed manner. We start by considering the inter-agent constraints \eqref{inter-agent collision avoidance ubar} and \eqref{semi-definite interagent new}.
To enforce these inter-agent constraints, each agent $i \in \calV $ needs $\{ \vH_{pos} \bmu_{x_k,\bar{u}}^j, \bmu_{d,k}^j \}_{k=0}^T$ of each of its neighbor agents $j \in \calN_i$. 
For convenience, let us assume $\hat{\bmu}_{x_k,\bar{u}}^i = \vH_{pos} \bmu_{x_k,\bar{u}}^i$ and define the sequences $ \hat{\bmu}^{i}_{\bar{u}} = \begin{bmatrix}
    \hat{\bmu}_{x_0,\bar{u}}^i; & 
    \hat{\bmu}_{x_1,\bar{u}}^i; &
    \dots ;
    \hat{\bmu}_{x_T,\bar{u}}^i
\end{bmatrix}$, 
$ \bmu^{i}_d = \begin{bmatrix}
\bmu_{d,0}^i; &
\bmu_{d,1}^i; &
\dots &
\bmu_{d,T}^i
\end{bmatrix}$. 
We can rewrite  problem \ref{Transformed Multi-agent robust optimization problem} in a simplified way as 
\begin{subequations}
\begin{align}
    \{ \bar{\bu}^i, \vK^i, \hat{\bmu}^{i}_{\bar{u}}, \bmu_{d}^i \}_{i \in \calV} = \argmin
    % \min_{ \{ \bar{\bu}^i, \vK^i, \hat{\bmu}_{i}^{\bar{u}}, \bmu^{d}_i\}_{i \in \calV} } 
    & \;
    \sum_{i \in \calV} 
    \hat{J}_i(\bar{\bu}^i,  \vK^i, \hat{\bmu}^{i}_{\bar{u}}, \bmu_{d}^i) \nonumber \\
    \mathrm{s.t.} \quad  
    \forall  \; i \in \calV, \;
    & \forall \; j \in \calN_i,  \nonumber \\
    & \calH_{i,j}(\hat{\bmu}^{i}_{\bar{u}}, \hat{\bmu}^{j}_{\bar{u}}) \geq 0 
    \label{inter-agent-constraints_ubar_simplified}\\
    & \Tilde{\calH}_{i,j}(\bmu_{d}^i, \bmu_{d}^j ) \geq 0
    \label{inter-agent-constraints k simplified}
\end{align}
\label{Problem1_Simplified}
\end{subequations}
\begin{equation}
\begin{aligned}
    \text{where}
        & 
        \hat{J}_i(\bar{\bu}^i,  \vK^i, \hat{\bmu}^{i}_{\bar{u}}, \bmu_{d}^i) =  J_i(\bar{\bu}^i, \vK^i)
        % \nonumber \\
        % & \qquad \qquad \qquad \qquad \qquad \qquad \qquad 
        + \calI_{f_i} + \calI_{\Bar{u}^i} 
        + \calI_{\vK^i} 
        \nonumber \\
        &
        \calI_{f_i}: \text{Indicator function for \eqref{dynamics}}
        \\
        &
        \calI_{\Bar{u}^i}: \text{Indicator function for \eqref{mean_constraint_form1}, \eqref{New mean formulation}, 
        \eqref{nonconvex obstacle final ubar},
        \eqref{semi-definite obstacle new}}, \\
        & \qquad \qquad
        \eqref{mu_d expression},
        \eqref{chance constraints 1},
        \eqref{chance constraints 2},
        \eqref{covariance_constraint_convex} 
        \\
        % &
        % \calI_{\vK^i}: \text{Indicator function for ((\ref{semi-definite obstacle new}), \eqref{mu_d expression}}
        % \\
        &
        \calH_{i,j}( \hat{\bmu}^{i}_{\bar{u}}, \hat{\bmu}^{j}_{\bar{u}}) \geq 0 :
        \text{Constraint set representing \eqref{inter-agent collision avoidance ubar}} \\
        &
        \Tilde{\calH}_{i,j}( \bmu_{d}^i, \bmu_{d}^j) \geq 0 : \text{Constraint set representing \eqref{semi-definite interagent new}}.
    \end{aligned} 
\end{equation}
% 
% \begin{equation}
%     \begin{aligned}
%         & \text{where} \; 
%         \hat{J}_i(\bar{\bu}^i,  \vK^i, \bmu_{x_k,\bar{u}}^i, \bmu^{d}_i) =  \bar{\bu}^i \T \vR_{\bar{u}} \bar{\bu}^i + || \vR_k \vK^i||_F^2 \nonumber \\
%         & \qquad \qquad \qquad \qquad \qquad \qquad \qquad 
%         + \calI_{f_i} + \calI_{rc}\\
%         &~~
%         \calI_{f_i}: \text{Indicator function for (\ref{compact dynamics})}
%         \\
%         &~~
%         \calI_{rc}: \text{Indicator function for (\ref{robust constraints single agent modified})}
%         \\
%         &~~
%         \calH_{i,j}(\bmu_{x,\bar{u}}^i, \bmu_{x,\bar{u}}^j) \geq 0 :
%         \text{Constraint set representing (\ref{inter_agent_Constraint_ubar_1})}\\
%         &~~
%         \Tilde{\calH}_{i,j}(\bmu_{x,d}^i, \bmu_{x, d}^j) \succeq 0 : \text{Constraint set representing (\ref{semi-definite_constraint_inter_agent})}
%     \end{aligned}
% \end{equation}
%
To solve the problem in a distributed manner, we first introduce the copy variables $\Bar{\bmu}^i_{j, \bar{u}}, \Bar{\bmu}^i_{j,d}$ at each agent $i$ corresponding to 
the variables $\hat{\bmu}_{\bar{u}}^j, \Bar{\bmu}_d^{j}$ of each of its neighbor agents $j \in \calN_i$.
Each agent then uses these copy variables to enforce the inter-agent constraints. However, these copy variables should capture the actual value of their corresponding variables. Thus, we need to enforce a consensus between 
the actual values of the variables $\hat{\bmu}^{i}_{\bar{u}}, \bmu^{i}_{d} $ for each agent $i$ and their corresponding copy variables $\Bar{\bmu}^{\hat{j}}_{i, \bar{u}}$, $\Bar{\bmu}^{\hat{j}}_{i, d}$ respectively at each agent $\hat{j}$ to which agent $i$ is a neighbor of ($\hat{j} \in \calP_i$). Hence, the above problem can be rewritten in terms of the copy variables as follows -
\begin{subequations}
\begin{align}
    & \{ \bar{\bu}^i, \vK^i, \hat{\bmu}^{i}_{\bar{u}}, \bmu_{d}^i \}_{i \in \calV} = \argmin 
    \;
    \sum_{i \in \calV} 
    \hat{J}_i(\bar{\bu}^i,  \vK^i, 
    \hat{\bmu}^{i}_{\bar{u}}, \bmu_{d}^i) \nonumber \\
    & \mathrm{s.t.} \quad  
    \forall  \; i \in \calV,   \nonumber \\
    & \qquad \qquad 
    \calH_{i,j}(\hat{\bmu}^{i}_{\bar{u}}, \Bar{\bmu}^i_{j, \bar{u}}) \geq 0, \;
    \forall \; j \in \calN_i, 
    \label{Hij equation with copy variables}\\
    & \qquad \qquad 
    \Tilde{\calH}_{i,j}(\bmu^{i}_d, \Bar{\bmu}^i_{j,d}) \geq 0, \;
    \forall \; j \in \calN_i,
    \label{tildeHij equation with copy variables}
    \\
    & \qquad \qquad 
     \hat{\bmu}^{i}_{\bar{u}} = \Bar{\bmu}^{\hat{j}}_{i, \bar{u}}, \;
     \forall \; \hat{j} \in \calP_i 
     \label{consensus1} \\
     & \qquad \qquad 
     \bmu^{i}_{d} = \Bar{\bmu}^{\hat{j}}_{i, d}, \;
     \forall \; \hat{j} \in \calP_i 
     \label{consensus2} 
\end{align}
\label{Problem 1 with copy variables}
\end{subequations}
Next, in order to use consensus ADMM (CADMM) \cite{boyd2011distributed}, we introduce the global variables $ \bnu^i_{\bar{u}} , \bnu^i_{d} $ such that the constraints \eqref{consensus1}, \eqref{consensus2} can be rewritten as follows. %
\begin{align} 
    \hat{\bmu}^{i}_{\bar{u}} = \Bar{\bmu}^{\hat{j}}_{i, \bar{u}} 
    & \iff 
    \;  \hat{\bmu}^{i}_{\bar{u}} = \bnu^i_{\bar{u}} ,
    \; \bnu^i_{\bar{u}}  =  \Bar{\bmu}^{\hat{j}}_{i, \bar{u}} \\
     \bmu^{i}_d =  \Bar{\bmu}^{\hat{j}}_{i,d} 
     & \iff 
    \; \bmu^{i}_d = \bnu^i_{d}, 
    \; \Bar{\bmu}^{\hat{j}}_{i,d} = \bnu^i_{d} 
\end{align}
Then, we define the local variables of each agent $i$ as $ \Tilde{\bmu}_{\bar{u}}^i  = \begin{bmatrix}
     \hat{\bmu}^{i}_{\bar{u}}; &
    \{  \Bar{\bmu}^i_{j, \bar{u}} = \}_{j \in \calN_i}
    \end{bmatrix}$, 
$\Tilde{\bmu}_{d}^i = \begin{bmatrix}
         \bmu^{i}_d; &
        \{\Bar{\bmu}^i_{j,d} \}_{j \in \calN_i}
    \end{bmatrix}$.
For convenience, we also define the global variable sequences 
$ \Tilde{\bnu}^i_{\bar{u} }  =  \begin{bmatrix}
     \bnu^{i}_{\bar{u}} ; &
    \{  \bnu^{j}_{\bar{u}} \}_{j \in \calN_i}
\end{bmatrix}$, 
$\Tilde{\bnu}^i_{d}  = \begin{bmatrix}
     \bnu^{i}_{d}; &
\{  \bnu^{j}_{d}  \}_{j \in \calN_i}
\end{bmatrix}$.
Using the above-defined variables, problem \eqref{Problem 1 with copy variables} can be rewritten as
 \begin{equation}
{\begin{aligned}
    \{ \bar{\bu}^i, \vK^i, \Tilde{\bmu}_{\bar{u}}^i ,  \Tilde{\bmu}_{d}^i \}_{i \in \calV} & = \argmin 
    \; 
    \sum_{i \in \calV}
    \hat{\calJ}_i(\bar{\bu}^i, \vK^i, \Tilde{\bmu}_{\bar{u}}^i,  \Tilde{\bmu}_{d}^i )
    %\quad (= J (\bu^1, \bu^2, \dots \bu^N) )  
    \\ 
    &
    \mathrm{s.t.} \; \forall  \; i \in \calV, 
    % \nonumber \\
    % &
    \quad \Tilde{\bmu}_{\bar{u}}^i  =  \Tilde{\bnu}^i_{\bar{u} } , \quad 
    \Tilde{\bmu}_{d}^i  =  \Tilde{\bnu}^i_{d}  
\end{aligned}
\label{Problem 1 distributed formulation}
}\end{equation} 
where $\hat{\calJ}_i (\bar{\bu}^i, \vK^i,  \Tilde{\bmu}_{\bar{u}}^i, \Tilde{\bmu}_{d}^i ) = \hat{J}_i(\bar{\bu}^i,  \vK^i, \hat{\bmu}^{i}_{\bar{u}}, \bmu_{d}^i) + \calI_{\calH_{i,j}} + \calI_{\Tilde{\calH}_{i,j}}$; and  $\calI_{\calH_{i,j}}$, $\calI_{\Tilde{\calH}_{i,j}}$ are indicator functions for the constraints \eqref{Hij equation with copy variables}, \eqref{tildeHij equation with copy variables} respectively.
\subsubsection{Distributed Approach}
Now that we have reformulated the problem to the form of \eqref{Problem 1 distributed formulation}, we can use CADMM to solve it in a distributed manner. The variables $\{ \Tilde{\bmu}_{\bar{u}}^i , \Tilde{\bmu}_{d}^i \}_{i \in \calV}$ constitute the first block, and $\{ \Tilde{\bnu}_{\bar{u}}^i, \Tilde{\bnu}_{d}^i \}_{i=1}^N$ constitute the second block of the CADMM. 
In each iteration of CADMM, the Augmented Lagrangian (AL) of the problem is minimized with respect to each block in a sequential manner followed by a dual update. The AL of the problem \eqref{Problem 1 distributed formulation} is given by
\begin{equation}
  \begin{aligned}
     \calL_{\rho_{\bar{u}}, \rho_d} = 
     \sum_{i \in \calV} 
     &~ 
    \big( \hat{\calJ}_i(\bar{\bu}^i, \vK^i, \Tilde{\bmu}_{\bar{u}}^i, \Tilde{\bmu}_{d}^i )
     \\
    &~~
    +  \blambda^i_{\bar{u}}  {}\T (\Tilde{\bmu}_{\bar{u}}^i  - \Tilde{\bnu}_{\bar{u}}^i )
    + \blambda^i_{d} {}\T ( \Tilde{\bmu}_{d}^i  - \Tilde{\bnu}_{d}^i) \\
    &~~
    + \frac{\rho_{\bar{u}}}{2} ||  \Tilde{\bmu}_{\bar{u}}^i  -  \Tilde{\bnu}_{\bar{u}}^i ||_2^2 
    + \frac{\rho_d}{2} ||  \Tilde{\bmu}_{d}^i  -  \Tilde{\bnu}_{d}^i ||_2^2 \big)
    \end{aligned} 
\end{equation}
where $ \blambda^i_{\bar{u}} $, $ \blambda^i_{d} $ are dual variables, and $\rho_{\bar{u}}$, $\rho_d$ are penalty parameters. 
Now, the update steps in an $l^{th}$ iteration of CADMM applied to the problem \eqref{Problem 1 distributed formulation} are as follows.
\paragraph{Local variables update (Block - I)} The update step for the local variables is given by
%
% The AL of the problem is minimized with respect to the local variables as follows -
\begin{align}
    & \{ \Tilde{\bmu}_{\bar{u}}^{i, l+1}  , \; \Tilde{\bmu}_{d}^{i,l+1}  \}_{i \in \calV}  \nonumber \\
    &~~~
    = \argmin \; \calL_{\rho_{\bar{u}}, \rho_d} ( \{  \Tilde{\bmu}_{\bar{u}}^i, \Tilde{\bmu}_{d}^i , 
     \Tilde{\bnu}_{\bar{u}}^{i,l},
    \Tilde{\bnu}_{d}^{i,l} ,
    \blambda^{i,l}_{\bar{u}} , 
     \blambda^{i,l}_{d} \}_{i \in \calV}  ) \nonumber
\end{align}
Note that there is no coupling between the local variables of different agents. Thus, the above update step can be performed in a decentralized manner where each agent $i$ solves its sub-problem to update the local variables  $\Tilde{\bmu}^{\bar{u}, l+1}_i, \; \Tilde{\bmu}^{d, l+1}_i$ {$\Tilde{\bmu}_{\bar{u}}^{i, l+1}, \; \Tilde{\bmu}_{d}^{i,l+1} $}. 
It is also worth noting that each agent's local sub-problem involves constraints, such as \eqref{nonconvex obstacle final ubar} and \eqref{inter-agent collision avoidance ubar}, which are linearized around the previous iteration value $\Tilde{\bmu}_{\bar{u}}^{i, l}$.
\paragraph{Global variables update (Block - II)} Next, the global variables are updated through 
\begin{align}
    & \{ \Tilde{\bnu}_{\bar{u}}^{i,l+1},
    \Tilde{\bnu}_{d}^{i,l+1} \}_{i \in \calV} 
    \nonumber \\
    &~~~
    = \argmin \; \calL_{\rho_{\bar{u}}, \rho_d} ( \{  \Tilde{\bmu}_{\bar{u}}^{i,l+1} , 
     \Tilde{\bmu}_{d}^{i, l+1} , 
    \Tilde{\bnu}_{\bar{u}}^{i},
    \Tilde{\bnu}_{d}^{i} ,
    \blambda^{i,l}_{\bar{u}} , 
    \blambda^{i,l}_{d} \}_{i \in \calV} ) \nonumber
\end{align}
Let us define $\blambda^i_{\bar{u}} $, $\blambda^i_{d} $ as sequence of dual vectors given as 
\mbox{
$\blambda^i_{\bar{u}} = \begin{bmatrix}
    \bar{\blambda}^i_{\bar{u}}; &
    \{\bar{\blambda}^{i}_{j, \bar{u}} \}_{j \in \calN_i}
\end{bmatrix}
$
}
, 
\mbox{
$\blambda^i_{\bar{u}}  = \begin{bmatrix}
     \bar{\blambda}^i_{d} ; &
    \{  \bar{\blambda}^{i}_{j, d} \}_{j \in \calN_i}
\end{bmatrix}$
}
,
such that the above update can be rewritten in a closed form  for each agent $i \in \calV $ as follows (details are provided in Appendix-I)
\begin{equation}
 \begin{aligned}
        & \bnu_{\bar{u}}^{i, l+1}= 
    \frac{1}{m_i} \bigg(
    \frac{ \bar{\blambda}^{i,l}_{\bar{u}}}{\rho_{\bar{u}}} 
    +  \hat{\bmu}^{i, l+1}_{\bar{u}} 
    + \sum_{\hat{j} \in \calP_i } \big( \frac{ \bar{\blambda}^{\hat{j}, l}_{i, \bar{u}}}{\rho_{\bar{u}}} 
    +  \Bar{\bmu}^{\hat{j}, l+1}_{i, \bar{u}}  \big) \bigg) \\
    & 
     \bnu_{d}^{i, l+1} = 
    \frac{1}{m_i} \bigg(
    \frac{ \bar{\blambda}^{i,l}_{d}}{\rho_{d}}
    + \bmu_{d}^{i, l+1} 
    + \sum_{ \hat{j} \in \calP_i } \big( \frac{ \bar{\blambda}^{\hat{j},l}_{i,d}} {\rho_{d} }
    +  \bar{\bmu}_{i,d}^{\hat{j}, l+1} \big) \bigg)
    \end{aligned} 
    \label{Problem 1 global update}
\end{equation}
where $m_i = (n({\calP_i}) + 1)$. 
After the local variables update, each agent $i$ would receive the updated variables  $\Bar{\bmu}^{\hat{j}, l+1}_{i, \bar{u}}$, $\bar{\bmu}_{i,d}^{\hat{j}, l+1}$ from all the agents $\hat{j}$ to which it is a neighbor of ($ \hat{j}  \in \calP_i$). Using these received local variables, each agent $i$ updates the global variables  $\bnu_{\bar{u}}^{i, l+1}, \bnu_{d}^{i, l+1}$ as per the update step \eqref{Problem 1 global update} .
\paragraph{Dual variables update} Finally, the dual variables are updated through the following rules
\begin{equation}
    \begin{aligned}
    & 
     \blambda^{i, l+1}_{\bar{u}} = 
    \blambda^{i, l}_{\bar{u}} 
    + \rho_{\bar{u}} ( \Tilde{\bmu}_{\bar{u}}^{i, l+1}  
    -  \Tilde{\bnu}_{\bar{u}}^{i, l+1} ) \\
    & \blambda^{i, l+1}_{d} = 
    \blambda^{i, l}_{d}
    + \rho_{d} ( \Tilde{\bmu}_{d}^{i, l+1} 
    - \Tilde{\bnu}_{d}^{i, l+1})
    \end{aligned}
    \label{Problem 1 dual update}
\end{equation}
After the global variables update, each agent $i$ would receive the updated global variables $ \bnu_{\bar{u}}^{j, l+1},  \bnu_{d}^{j,l+1}$ from all its neighbor agents $j \in \calN_i$. Using these received variables, the agent $i$ updates the dual variables $ \blambda^{i,l+1}_{\bar{u}}, \blambda^{i, l+1}_{d}$ as per the update step \eqref{Problem 1 dual update}. %
It should be noted that in each update step, all the agents do the update in parallel. 
%  {Further, the agents need not exchange other system information (such as dynamics parameters or uncertainty set parameters). }

\subsection{Computational Complexity Analysis}
\label{Computational Complexity Analysis}

In this section, we theoretically verify the advantages of the proposed approach in terms of computational complexity. For the following analysis, we consider a multi-agent system of $N$ agents, $n_{\text{obs}}$ obstacles, and with each agent having $n_{\text{inter}}$ neighbors. For simplicity, we consider the case of only having deterministic uncertainty related constraints, but the provided analysis can be readily extended for the case where the constraints \eqref{chance constraints 1}, \eqref{chance constraints 2} and \eqref{covariance_constraint_convex} are also added.  All results are derived assuming the problems are solved with the interior point method which is considered the state-of-the-art for SOCPs and SDPs \cite{ben2001lectures, lobo1998applications}.  
\subsubsection{Complexity reduction due to constraint approximation}

First, we demonstrate the complexity reduction that occurs from the constraint reformulation presented in Section \ref{new reformulation} compared to the SDP reformulation in \ref{initial approach}. For that, we provide a comparison of the complexity involved in solving each local sub-problem with the proposed distributed method versus a distributed SDP framework. The latter refers to a distributed framework that involves the initial approach SDP reformulation \ref{initial approach} for collision avoidance constraints. 
% We assume that the shared variables among the agents are the control parameters since it is not straightforward to have state means as the shared variables in the distributed SDP framework. 
%
\begin{proposition}
The computational complexity for solving each local sub-problem with the distributed SDP framework is 
\begin{equation}
O \Big( L_{\text{IP}} T^5 (n_{\text{inter}} + n_{\text{obs}})^3 n_{u_i}^2 n_{x_i}^2 n_{d_i}^2 \gamma_h^2 \Big)
\end{equation}
where $L_{\text{IP}}$ is the number of iterations of the interior point method.
With the proposed reformulation, this complexity is reduced to 
\begin{equation}
O \Big( L_{\text{IP}} T^4 (n_{\text{inter}} + n_{\text{obs}})^3 n_{u_i}^2 n_{x_i}^2 n_{d_i} \gamma_h^2 \Big).
\end{equation}
\label{complexity bounds}
\end{proposition}
\begin{proof}
    Provided in Appendix-J.
\end{proof}
\subsubsection{Complexity reduction due to decentralization}
Subsequently, we illustrate the computational benefits of the proposed distributed algorithm compared to an equivalent centralized approach. As the local subproblems are the most computationally expensive updates in the distributed ADMM algorithm, its complexity can be expressed as follows.
\begin{corollary}
The computational complexity of the proposed distributed algorithm is given by
\begin{equation}
O \Big( L_{\text{ADMM}} L_{\text{IP}} T^4 (n_{\text{inter}} + n_{\text{obs}})^3 n_{u_i}^2 n_{x_i}^2 n_{d_i} \gamma_h^2 \Big),
\end{equation}
where $L_{\text{ADMM}}$ is the number of ADMM iterations.
\end{corollary}
For comparison purposes, we also provide the complexity of an equivalent centralized approach.
\begin{proposition}
% Before applying the constraint reformulation, a centralized approach for solving the multi-agent problem () would yield a computational complexity 
% %
% {\color{red}needs to be changed}
% \begin{equation}
% []
% \end{equation}
% %
% when solved with the interior point method.
% %
% With the proposed reformulation, this complexity would be reduced to 
% %
% \begin{equation}
% O \Big( L_{\text{IP}} N^3 T^4 (n_{\text{inter}} + n_{\text{obs}})^3 n_{u_i}^2 n_{x_i}^2 n_{d_i} \gamma_h^2 \Big).
% \end{equation}
A centralized approach for solving the multi-agent problem \ref{Transformed Multi-agent robust optimization problem} with the proposed reformulated constraints would yield a computational complexity of 
\begin{equation}
O \Big( L_{\text{Lin}} L_{\text{IP}} N^3 T^4 (n_{\text{inter}} + n_{\text{obs}})^3 n_{u_i}^2 n_{x_i}^2 n_{d_i} \gamma_h^2 \Big).
\label{centralized complexity bound}
\end{equation}
where $L_{\text{Lin}}$ is the number of outer loops where the non-convex constraints are linearized based on the previous solution.
\end{proposition}
\begin{proof}
Omitted due to similarity with Proposition 4.
\end{proof}
\section{Simulation Results}
\label{sec: simulation results}
In this section, we provide simulation results of the proposed method applied to a multi-agent system with underlying stochastic and deterministic uncertainties. In Section \ref{sec: sims part 1}, the effectiveness of the proposed robust constraints is analyzed under different scenarios. In Section \ref{sec: sims part 2}, the scalability of the proposed distributed framework is illustrated. The agents have 2D double integrator dynamics with the exact dynamic matrices, and the uncertainty information is provided in Appendix-K. In all experiments, we set the time horizon to $T = 20$, except if stated otherwise. For every agent, the neighbor agents are considered to be a set of specified number of agents closest to the agent at the initial state. All experiments are also illustrated through the supplementary video \footnote{\url{https://youtu.be/Q5xghaqt4SQ}}.
\subsection{Effectiveness of Proposed Robust Constraints}
\label{sec: sims part 1}
We start by validating the effectiveness of the proposed robust constraints by considering different cases of underlying uncertainties. First, we analyze the constraints that are only on the expectation of the state. Subsequently, we analyze the robust chance constraints and covariance constraints, which are dependent on both the deterministic and stochastic disturbances.
\subsubsection{With deterministic disturbances}
\begin{figure}[t] 
    \centering
  \subfloat[Robust case \label{1a}]{
       \includegraphics[width= 0.49\linewidth, trim={0.5cm 1.5cm 0.5cm 2.5cm},clip]{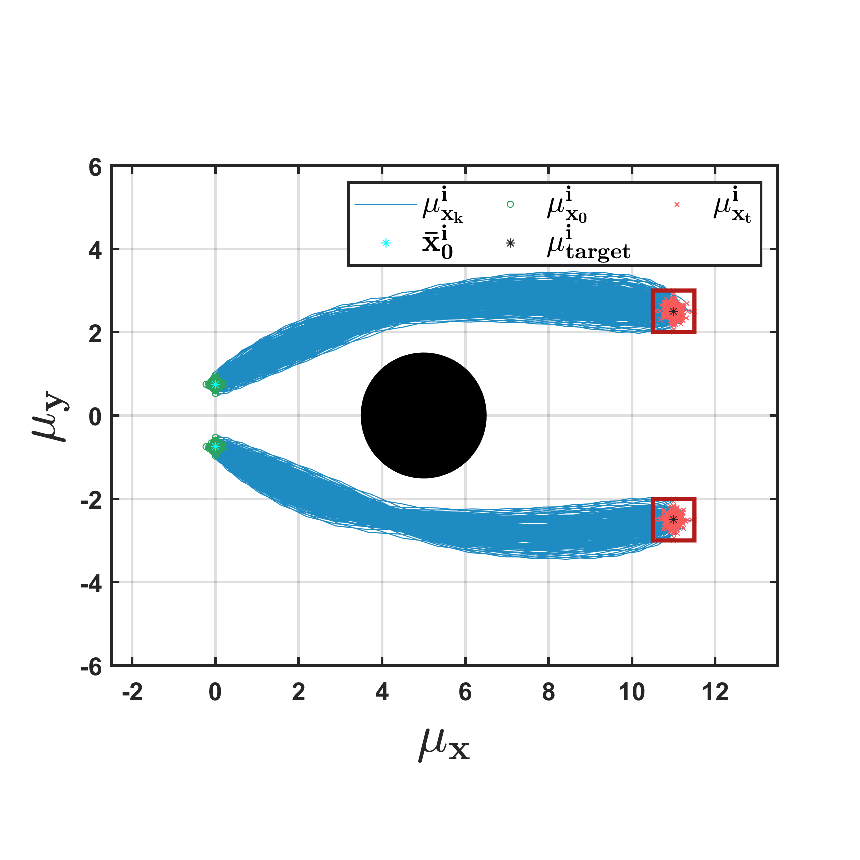}} 
  \subfloat[Non-robust case \label{1b}]{
        \includegraphics[width=0.49\linewidth, trim={0.5cm 1.5cm 0.5cm 2.5cm},clip]{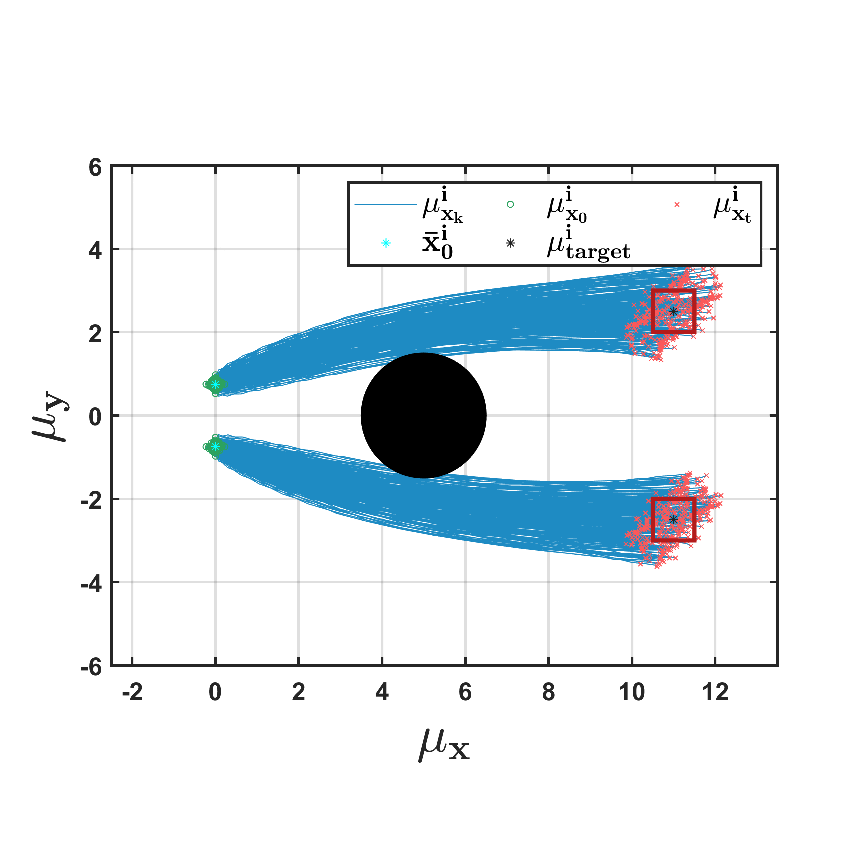}}
    \\
  \caption{\textbf{Two agents scenario with obstacle:} Performance comparison between robust and non-robust case with terminal mean and obstacle constraints.}
  \label{fig1} 
\end{figure}
\begin{table}[t!]
\centering
\begin{tabular}{|c|ccc|}
\hline
\multirow{2}{*}{} & \multicolumn{3}{c|}{No. of Obstacles}                          \\ \cline{2-4} 
                  & \multicolumn{1}{c|}{1}     & \multicolumn{1}{c|}{2}     & 3    \\ \hline
Initial Reformulation \eqref{obstacle_semidef_old} & \multicolumn{1}{c|}{11.58} & \multicolumn{1}{c|}{22.01} & -    \\ \hline
Proposed Reformulation & \multicolumn{1}{c|}{0.73}  & \multicolumn{1}{c|}{1.37}  & 1.84 \\ \hline
\end{tabular}
\caption{Computational time comparison - Initial Reformulation vs proposed reformulation}
    \label{old vs new table}
\end{table}
% \begin{table}[t!]
%     \begin{center}
%     \begin{tabular}{|c|c|c|c|}
%     \multirow
%     \hline
%     & 1
%     & 2 
%     & 3
%     \\[2.5pt]
%     \hline
%     Initial Approach
%     & 11.58
%     & 22.01
%     & -
%     \\[2.5pt]
%     \hline
%     Proposed Reformulation
%     & 0.73
%     & 1.37
%     & 1.84
%     \\[2.5pt]
%     \hline
%     \end{tabular}
%     \end{center}
%     \caption{Computational time comparison - Initial approach vs proposed reformulation}
%     \label{old vs new table}
% \end{table}
%
Here, we demonstrate the effectiveness of the proposed framework by providing a comparison with the non-robust case, and with the initial reformulation case.
\begin{figure*}[t!] 
    \centering
  \subfloat[ Robust mean trajectories \label{2a}]{
       \includegraphics[width= 0.32\linewidth, trim={0.75cm 0cm 1.25cm 1cm},clip]{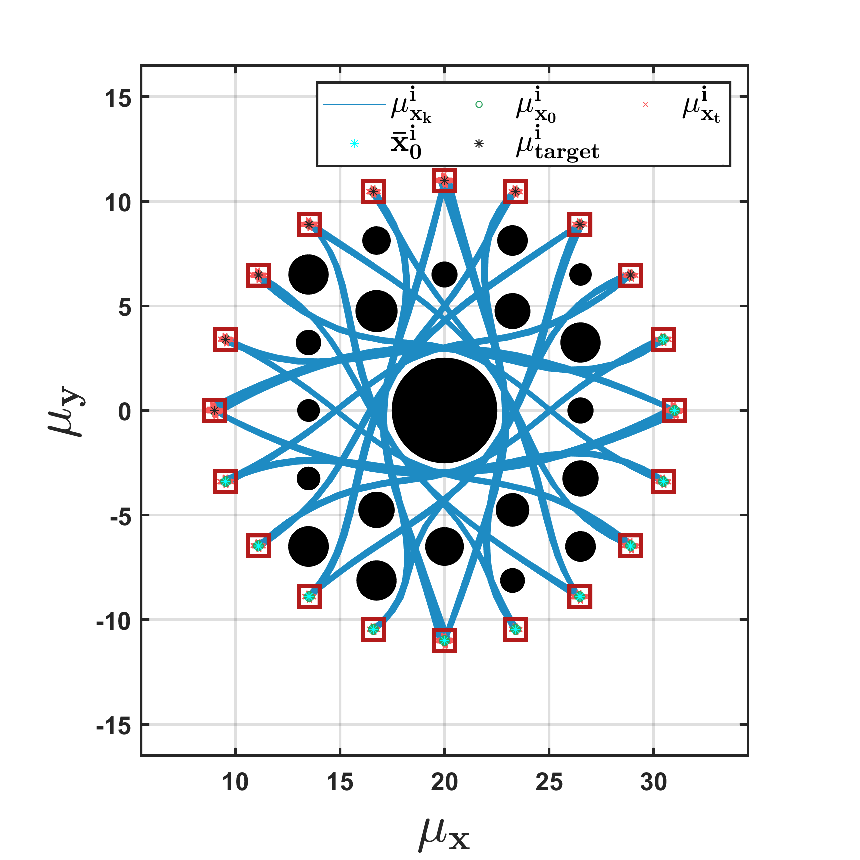}}
\subfloat[ $k = 5$ \label{2b}]{
        \includegraphics[width=0.32\linewidth, trim={0.75cm 0cm 1.25cm 1cm},clip]{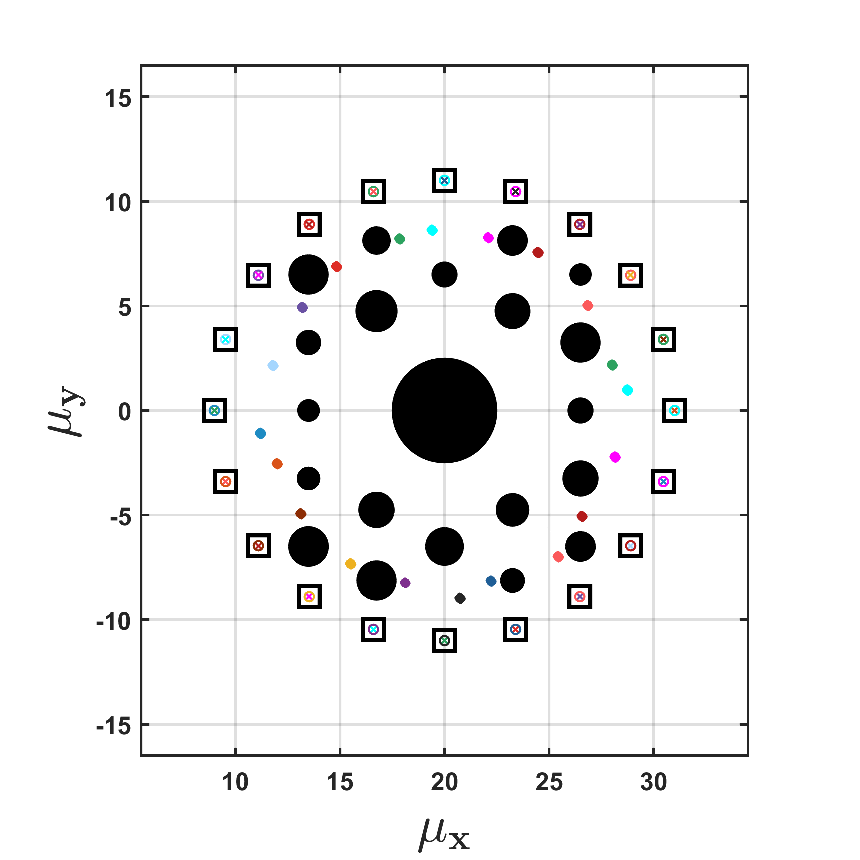}}
\subfloat[ $k = 10$ \label{2c}]{
        \includegraphics[width=0.32\linewidth, trim={0.75cm 0cm 1.25cm 1cm},clip]{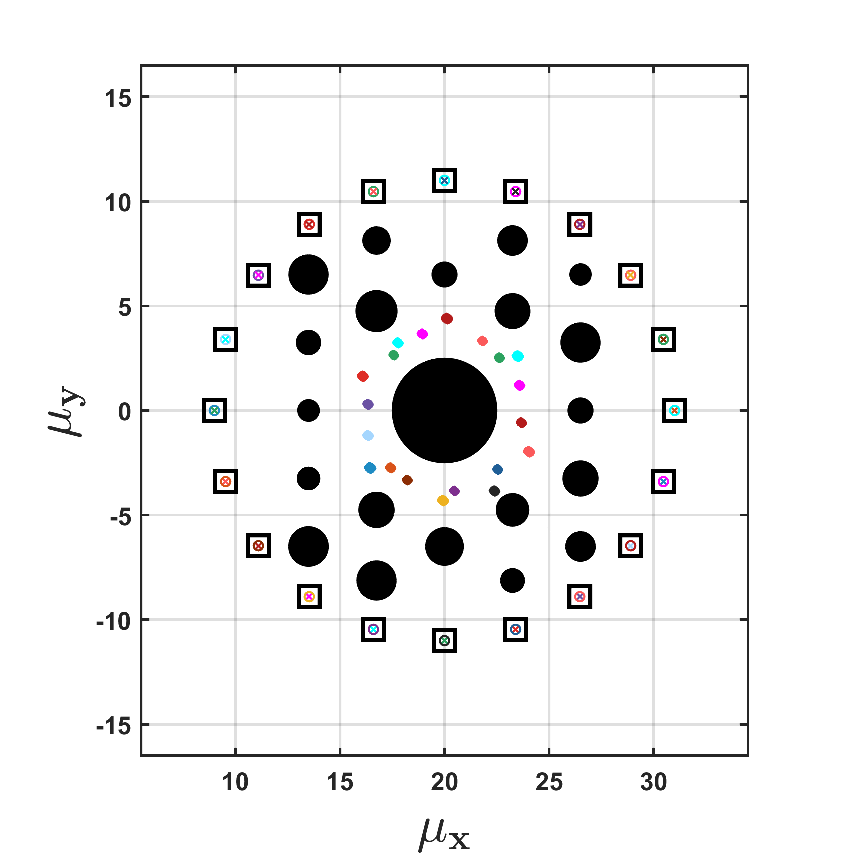}}
\hfill
  \subfloat[ $k = 15$ \label{2d}]{
        \includegraphics[width=0.32\linewidth, trim={0.75cm 0cm 1.25cm 1cm},clip]{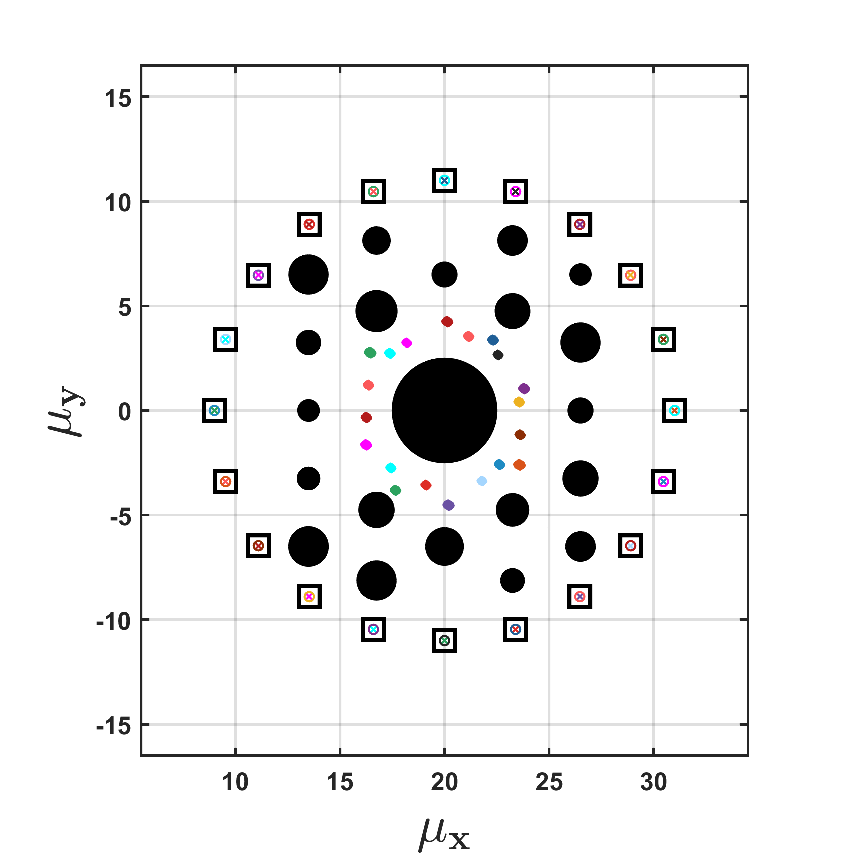}}
 \subfloat[$k = 20$ \label{2e}]{
       \includegraphics[width= 0.32\linewidth, trim={0.75cm 0cm 1.25cm 1cm},clip]{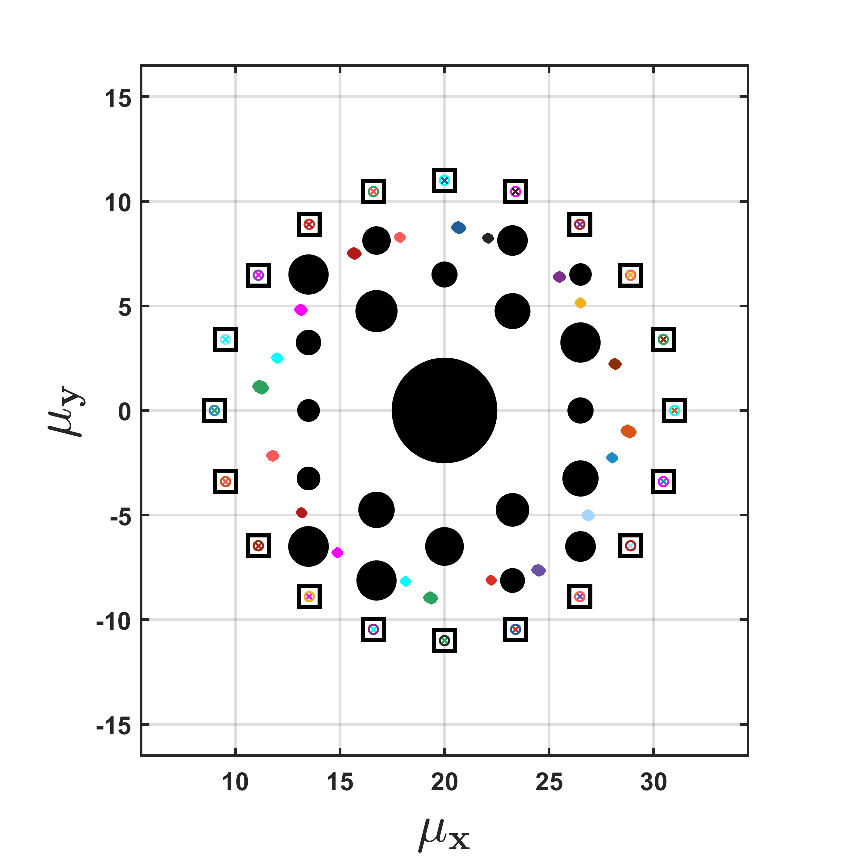}}
\subfloat[ Distance plot \label{2f}]{
\includegraphics[width=0.31\linewidth, trim={0cm -0.75cm 0cm 0cm},clip]{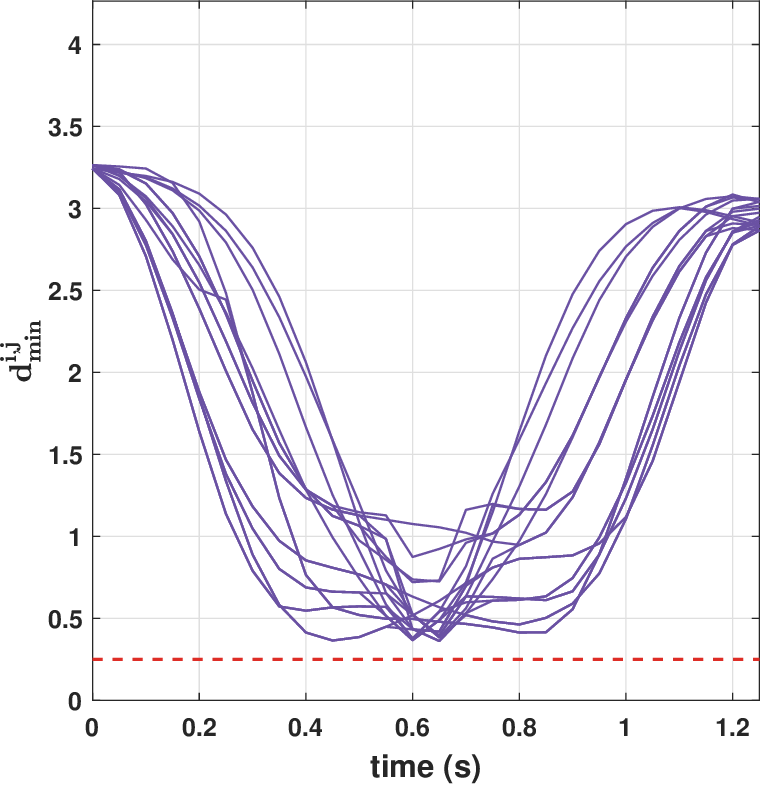}}
    \\
  \caption{ \textbf{Twenty agents scenario with circle formation task and 21 obstacles:} {(a) Robust mean trajectories, (b)-(e) Snapshots of the robust mean trajectories of agents, (f) Distance plot showing the minimum distance of each agent from its nearest neighbor at each time step.}}
  \label{fig2} 
\end{figure*}

First, we compare the results of applying the proposed distributed framework for two cases: robust and non-robust. With non-robust, we refer to the case of finding the optimal control trajectory by only considering standard optimization constraints instead of explicitly accounting for the deterministic uncertainties through robust optimization constraints (robust case).
Fig. \ref{fig1} discloses a two-agent scenario with terminal mean and obstacle avoidance constraints. The plots demonstrate $100$ trajectories of the two agents for different realizations of the disturbances. As shown in Fig. \ref{1a}, in the robust case, each agent $i$ successfully avoids the obstacle, and its disturbance-free state mean reaches the target terminal mean $\bmu_{target}^i$ with uncertainty in its terminal state mean lying inside the set target bounds (represented by red boxes). On the contrary, for the non-robust case in Fig. \ref{1b}, the agents might collide with the obstacle in the center, and the uncertainty in the terminal state mean of each agent might lie outside the target bounds. 
\begin{figure}[t!] 
    \centering
\subfloat[ Robust Mean Trajectories \label{3a}]{
       \includegraphics[width= 0.49\linewidth, trim={0.6cm 2cm 0.75cm 3cm},clip]{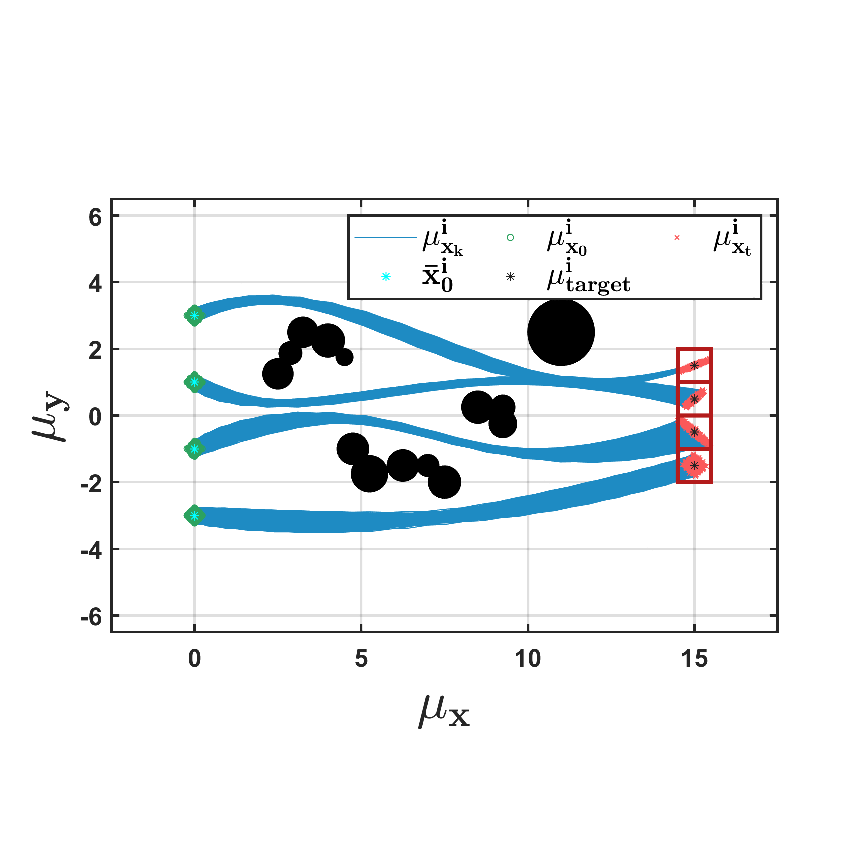}}
\subfloat[ $k = 10 $ \label{3b}]{
        \includegraphics[width=0.49\linewidth, trim={0.6cm 2cm 0.75cm 3cm},clip]{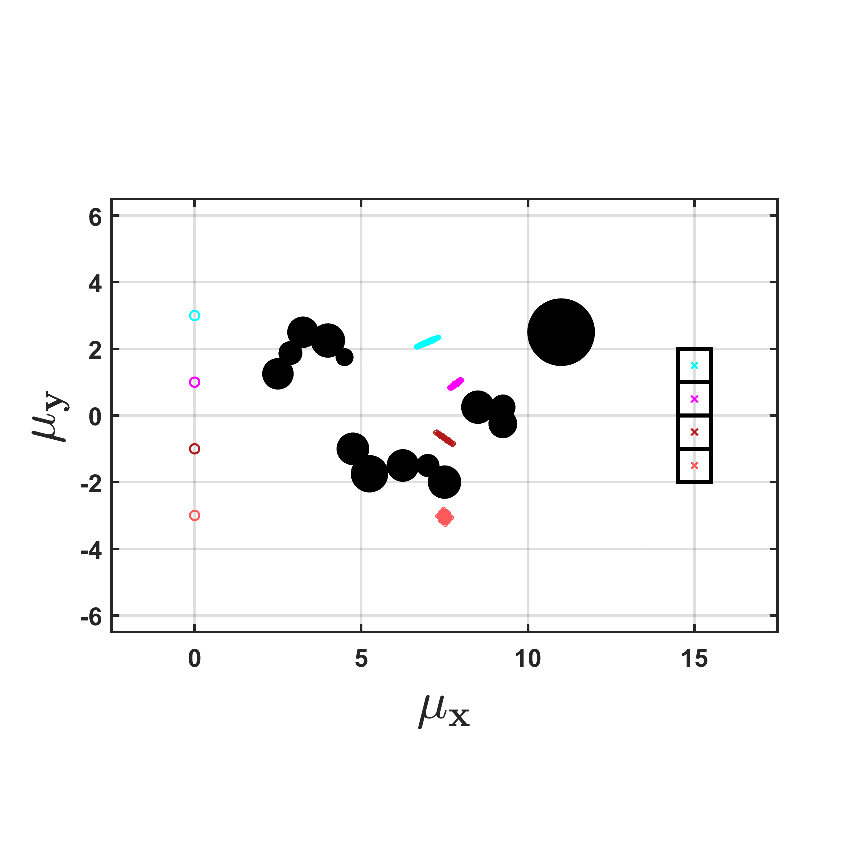}}
        \hfill
\subfloat[ $k = 15$ \label{3c}]{
        \includegraphics[width=0.49\linewidth, trim={0.6cm 2cm 0.75cm 3cm},clip]{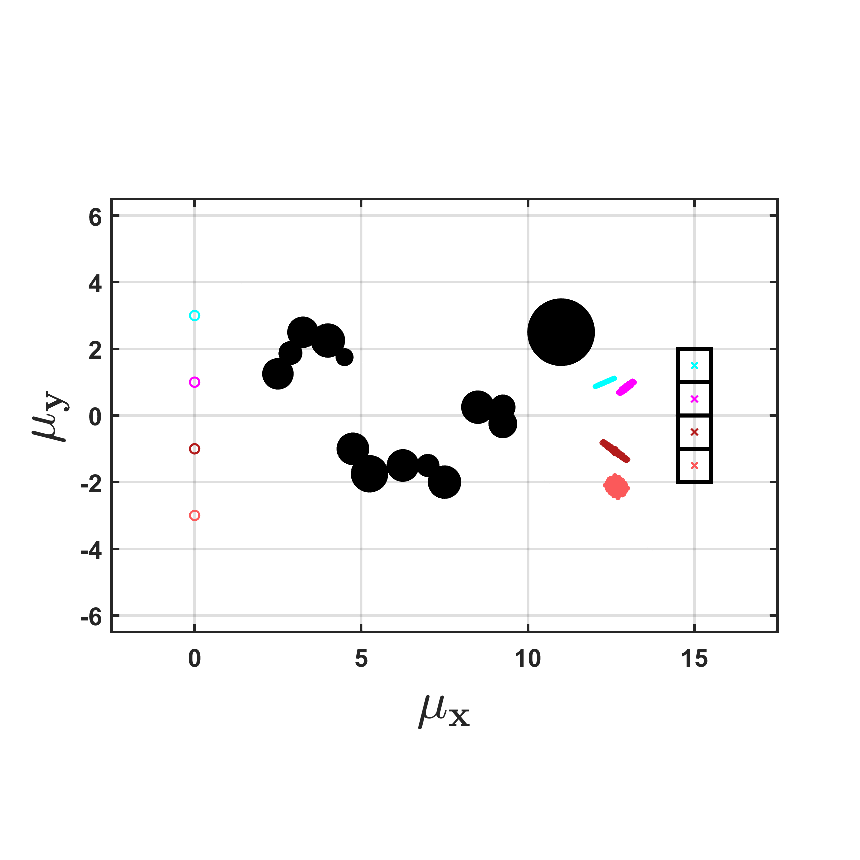}}
\subfloat[ $k = 20$ \label{3d}]{
        \includegraphics[width=0.49\linewidth,trim={0.6cm 2cm 0.75cm 3cm},clip]{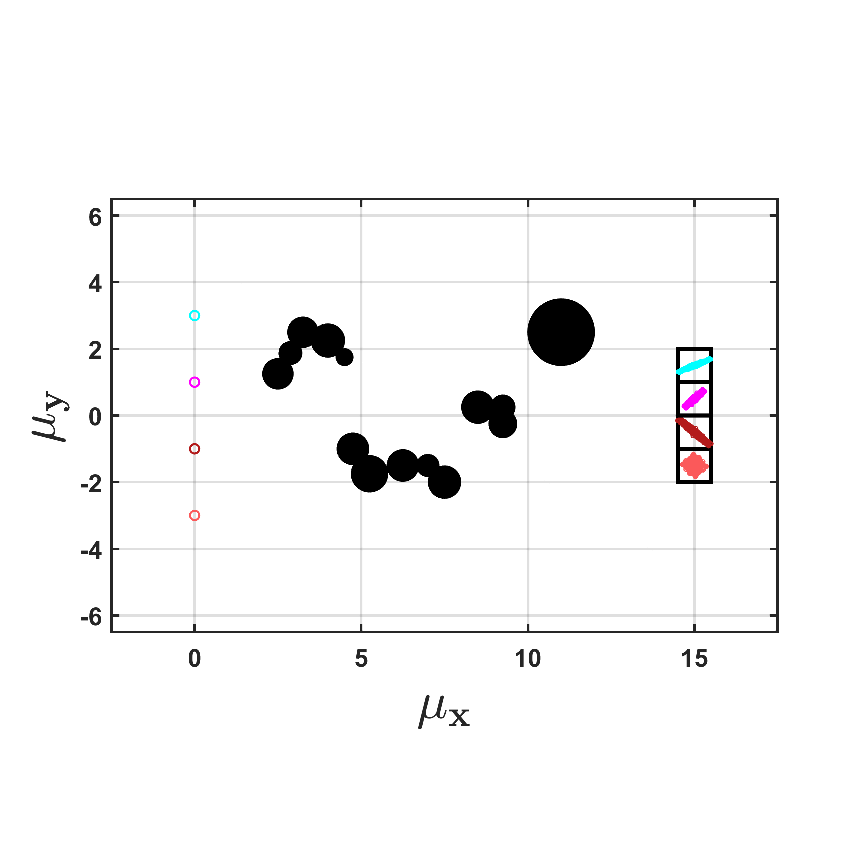}}
     \\
  \caption{ \textbf{Four agents scenario-1 with non-convex obstacles:} (a) Robust mean trajectories, (b)-(d) Snapshots of the robust mean trajectories of agents.}
  \label{fig3} 
\end{figure}
\begin{figure}[t!] 
    \centering
  \subfloat[ Robust Mean Trajectories \label{4a}]{
       \includegraphics[width= 0.49\linewidth, trim={0.6cm 1.5cm 0.75cm 3cm},clip]{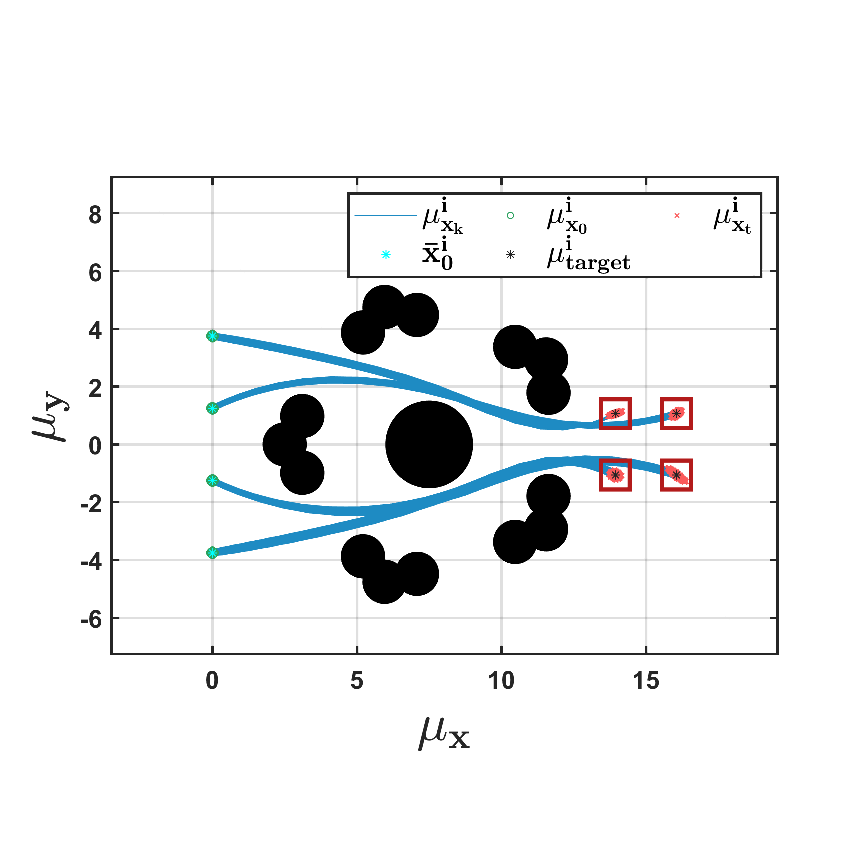}}
  \subfloat[ $k = 10 $ \label{4b}]{
        \includegraphics[width=0.49\linewidth, trim={0.6cm 1.5cm 0.75cm 3cm},clip]{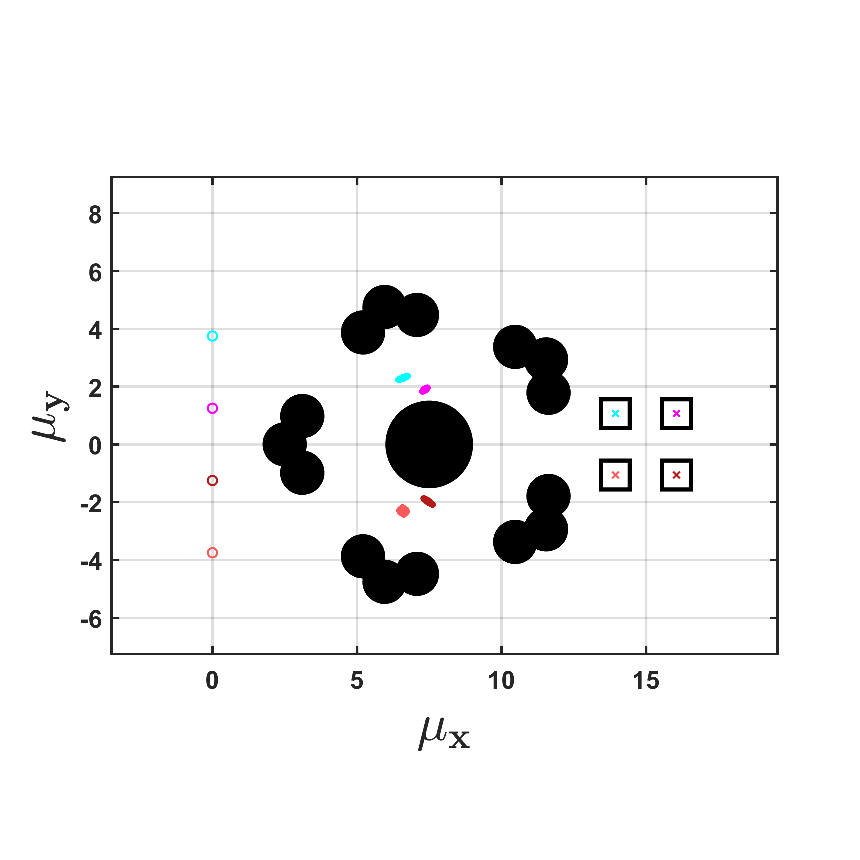}}
        \hfill
\subfloat[ $k = 15$ \label{4c}]{
        \includegraphics[width=0.49\linewidth, trim={0.6cm 1.5cm 0.75cm 3cm},clip]{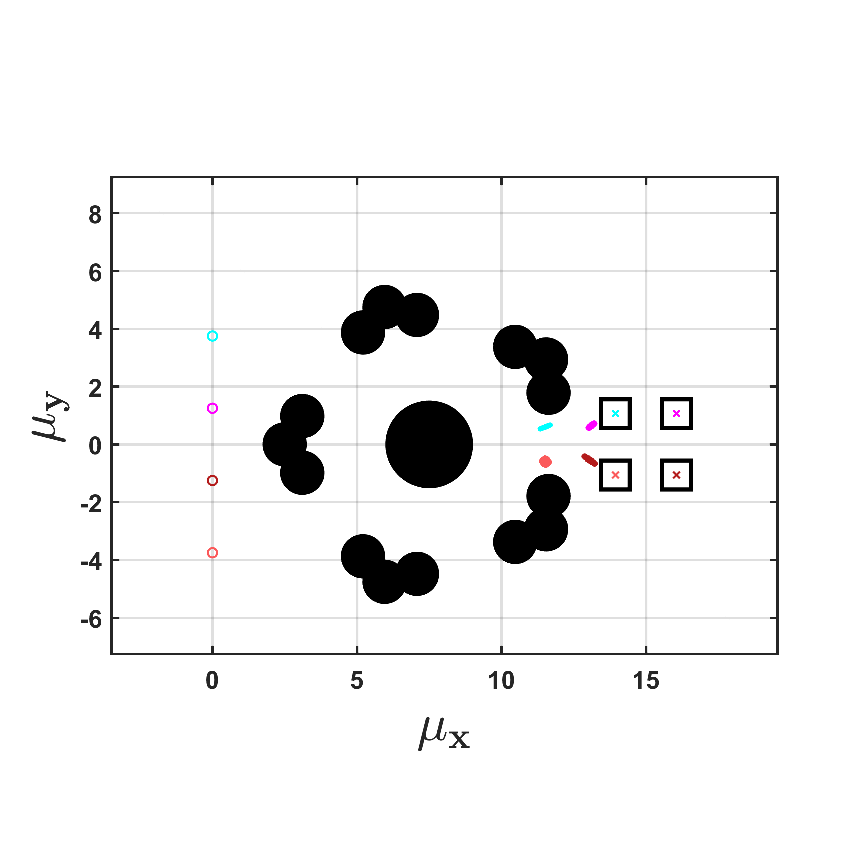}}
\subfloat[ $k = 20$ \label{4d}]{
        \includegraphics[width=0.49\linewidth, trim={0.6cm 1.5cm 0.75cm 3cm},clip]{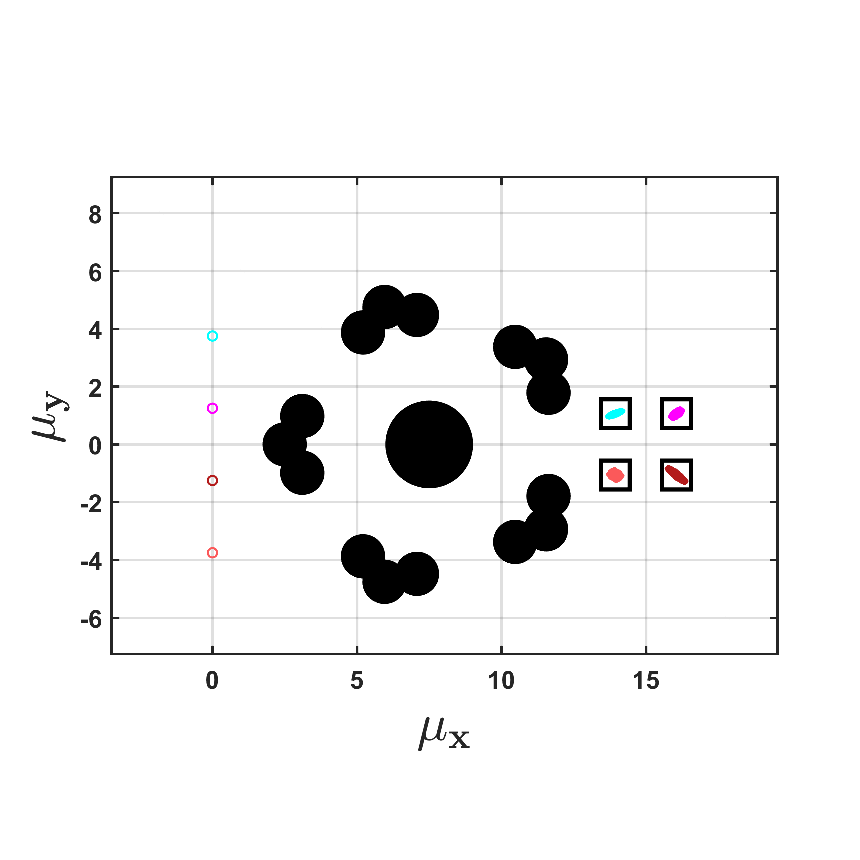}}
     \\
  \caption{ \textbf{Four agents scenario-2 with non-convex obstacles:} (a) Robust Mean Trajectories, (b)-(d) Snapshots of the robust mean trajectories of agents.}
  \label{fig4} 
\end{figure}

Next, we analyze the computational efficiency of the proposed robust reformulated constraints over the initial reformulation \eqref{obstacle_semidef_old} by comparing the computational time of solving a single-agent problem using both approaches. The problem involves solving for the agent to satisfy the terminal mean constraints and the obstacle constraints. Table \ref{old vs new table} discloses the optimizer time (in seconds) to solve this problem using each reformulation for an increasing number of obstacles. The initial reformulation fails to provide a solution when there are three or more obstacles, and the time taken with the initial reformulation is significantly longer than the proposed formulation. Moreover, it can be observed that, compared with the initial reformulation, the time taken with the proposed reformulation does not increase significantly as the number of obstacles increases. 

We now demonstrate the effectiveness of the proposed framework in complex environments. In each case, a robust mean trajectories plot is included to show that the agents reach the target terminal mean bounds without collision with the obstacles.
Fig. \ref{fig2} shows a scenario with twenty agents and 21 obstacles, where each agent in a circular formation is tasked to reach the opposite side without colliding with obstacles and other agents. We consider a time horizon T = 25. In Fig. \ref{2c} - \ref{2e}, we show snapshots of the robust trajectories of the agents for different time steps, demonstrating that the agents do not violate any of the constraints during the tasks. Additionally, we provide a distance plot in Fig. \ref{2f} showing the minimum distance of each agent from its nearest neighbor ($d_{min}^{i,j}$) at each time step, and the red dashed line representing the collision distance threshold (at 0.25). The nearest neighbor of an agent at any time step is estimated based on the distance between the realizations of the agent and that of its neighbors. It can be observed that the minimum distance decreases first as the agents reach the center and then increases as the agents move away from the center. All the minimum distances lie above the collision threshold, proving no inter-agent collisions occur. Fig. \ref{fig3} and Fig. \ref{fig4} demonstrate the performance of the proposed framework in the presence of non-convex obstacles. In Fig. \ref{fig3} and \ref{fig4}, the non-convex structures are formed using 14 and 16 circular obstacles, respectively. From the snapshots in Fig. \ref{3b} - \ref{3d} and \ref{4b} - \ref{4d}, it can be observed that the agents were able to steer through the obstacles without any collision. 
Further, it is worth noting that in the robust case, the agents not only avoid a collision but also the uncertainty in the state mean of an agent is reduced when it approaches an obstacle or another agent.    
\subsubsection{With both deterministic and stochastic disturbances}
\begin{figure}[t!] 
    \centering
  \subfloat[\label{5a}]{
       \includegraphics[width= 0.49\linewidth, trim={0.5cm 2cm 0.5cm 3.25cm},clip]{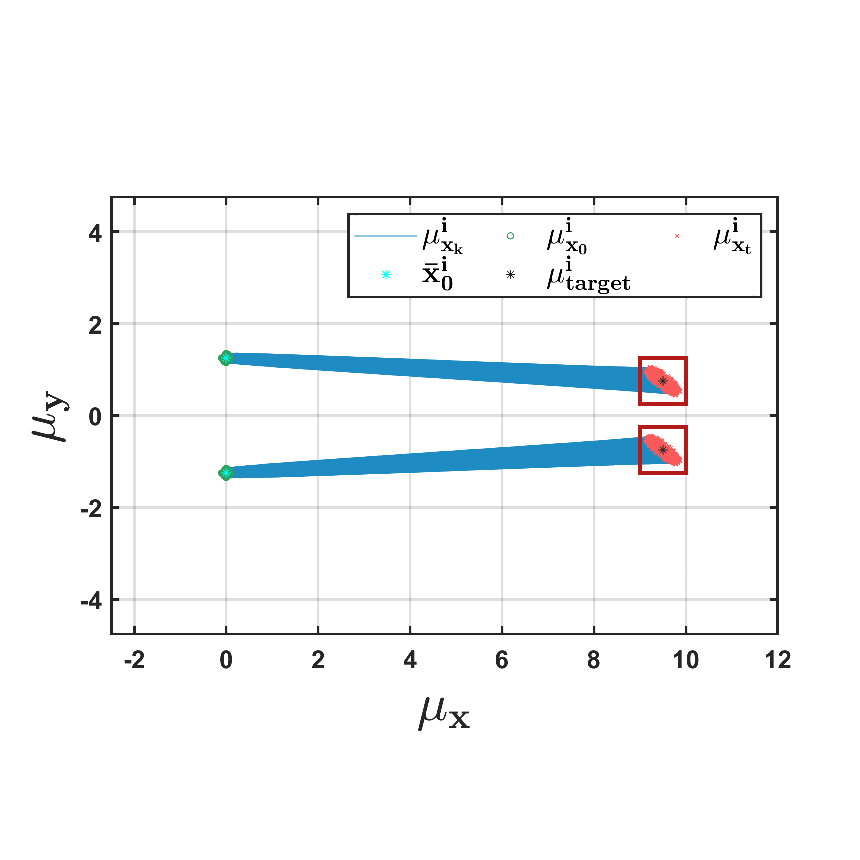}}
  \subfloat[\label{5b}]{
    \includegraphics[width=0.49\linewidth, 
    trim={0.5cm 2cm 0.5cm 3.25cm},clip]{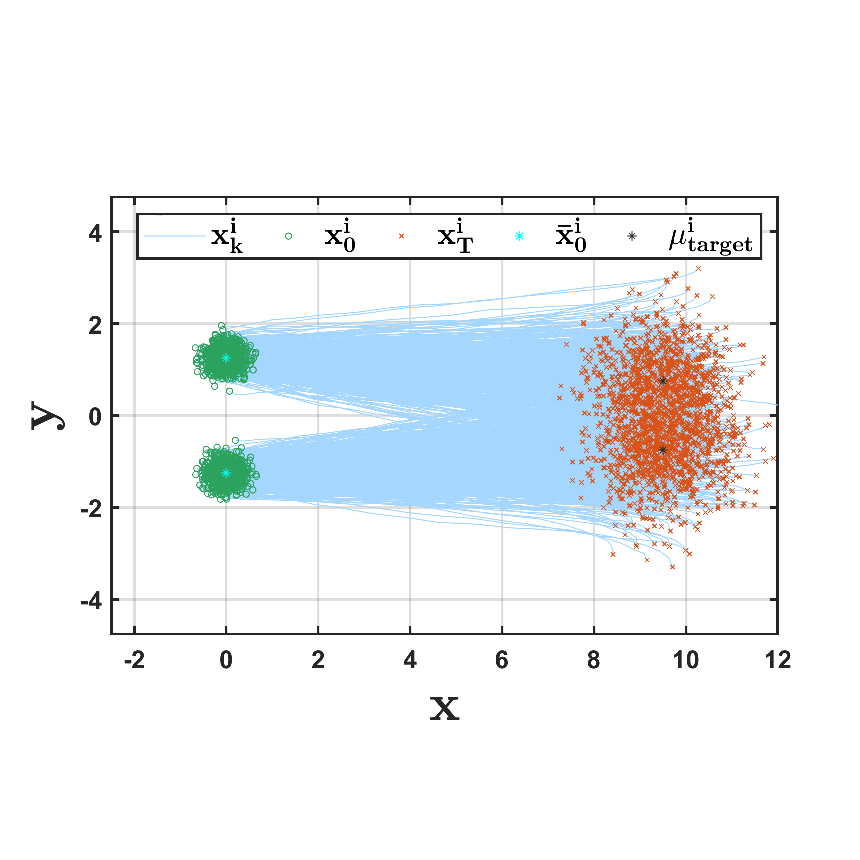}}
    \hfill
  \subfloat[\label{5c}]{
       \includegraphics[width= 0.49\linewidth, trim={0.5cm 2cm 0.5cm 3.25cm},clip]{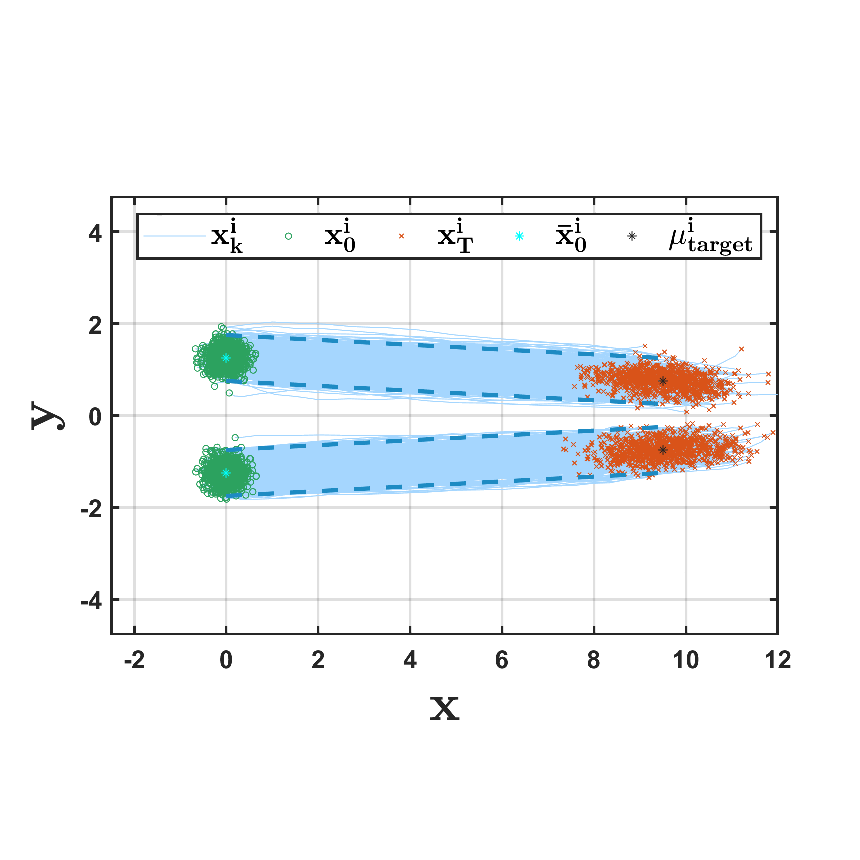}}
    \subfloat[\label{5d}]{
       \includegraphics[width= 0.49\linewidth, trim={0.5cm 2cm 0.5cm 3.25cm},clip]{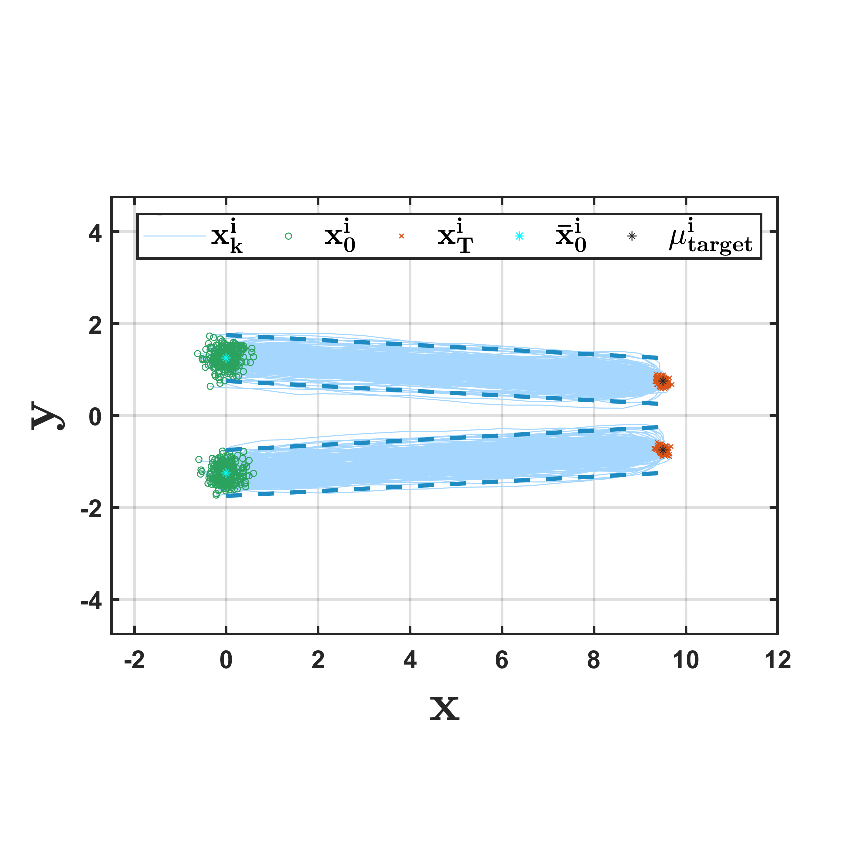}}    
    \\
  \caption{\textbf{Performance analysis of robust chance constraints and Covariance Constraints - Scenario 1:} (a) Mean trajectories of agents in deterministic case, 
  (b) Trajectories of agents in deterministic case without the robust chance constraints, 
  (c) Trajectories of agents in mixed case with robust chance constraints, 
  (d) Trajectories of agents in mixed case with robust chance constraints and terminal covariance constraints.
        }
  \label{fig5} 
\end{figure}
\begin{figure*}[t!] 
    \centering
  \subfloat[\label{6a}]{
       \includegraphics[width= 0.32\linewidth, trim={0.5cm 1.5cm 0.5cm 2.5cm},clip]{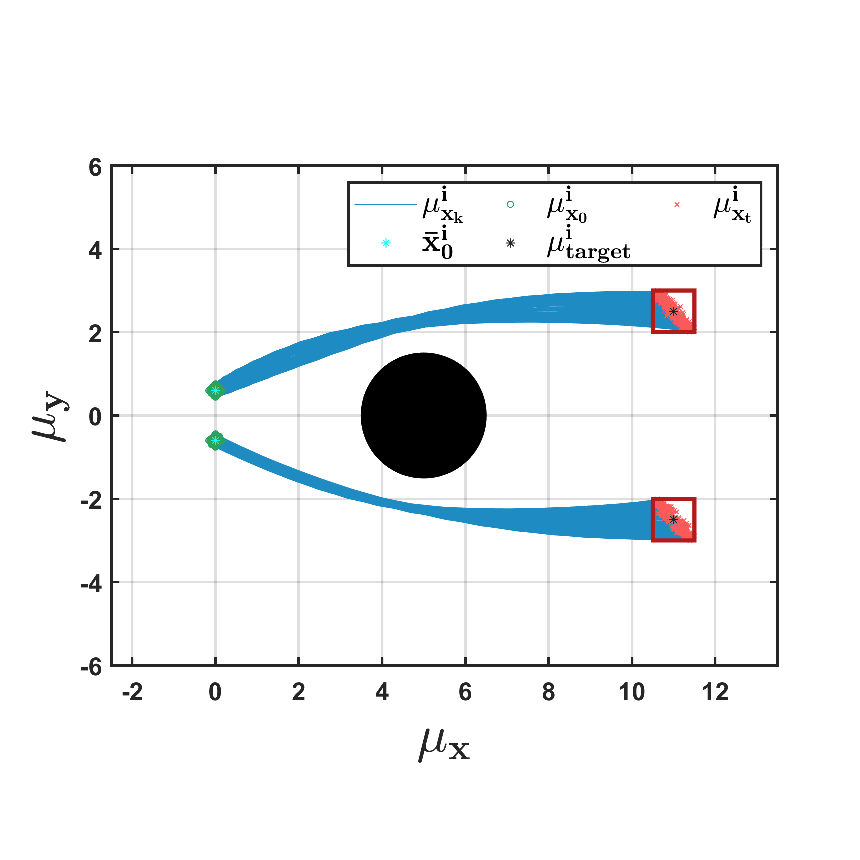}}
  \subfloat[\label{6b}]{
    \includegraphics[width=0.32\linewidth, trim={0.5cm 1.5cm 0.5cm 2.5cm},clip]{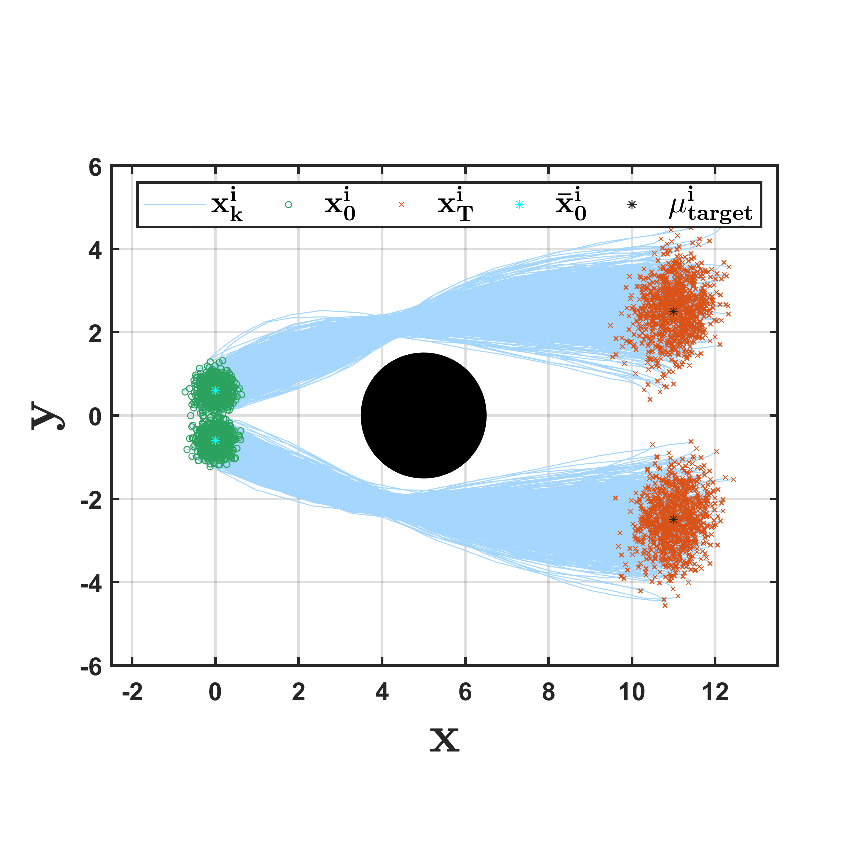}}
  \subfloat[\label{6c}]{
       \includegraphics[width= 0.32\linewidth, trim={0.5cm 1.5cm 0.5cm 2.5cm},clip]{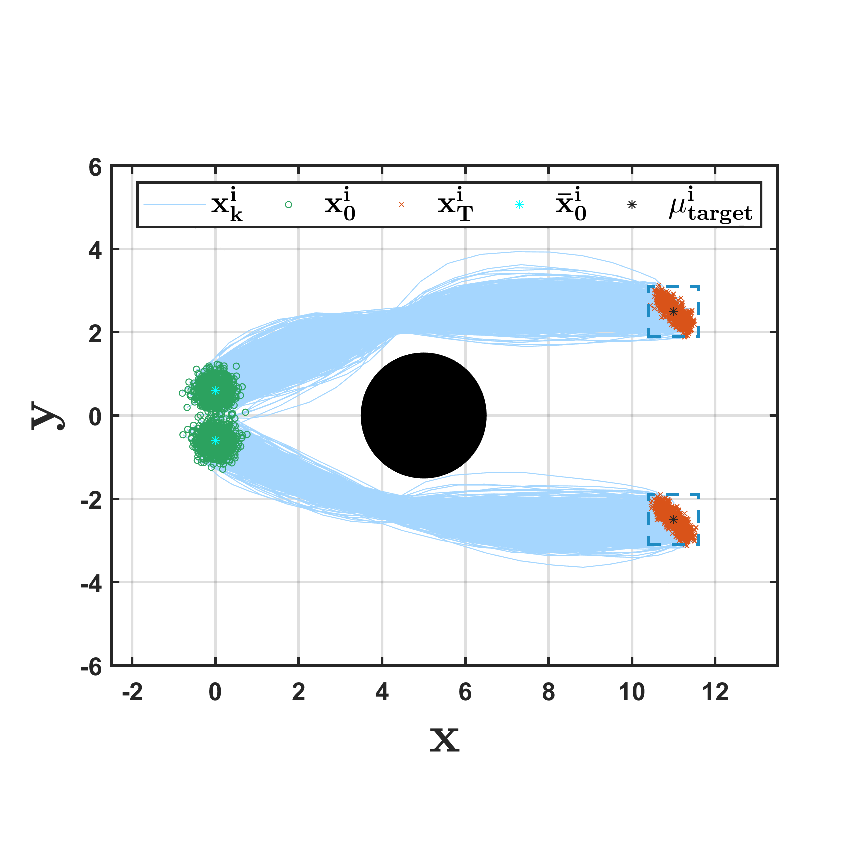}}
    \\
  \caption{\textbf{Performance analysis of robust chance constraints - Scenario 2:} (a) Mean trajectories of agents in the deterministic case, (b) Trajectories of agents in the deterministic case without the robust chance constraints, (c) Trajectories of agents in mixed case with robust chance constraints.}     
  \label{fig6} 
\end{figure*}
In this subsection, we demonstrate the effectiveness of the proposed framework with robust chance constraints and covariance constraints. For that, we compare two cases, one where 
% all the robust constraints are active except for the robust chance constraints and covariance constraints, 
the constraints that are only on the expectation of the state are active
(`deterministic case'), and another where all the robust constraints, including the robust chance and covariance constraints, are active (`mixed case'). Further, it should be noted that when both disturbances are present, the mean trajectory differs from the actual trajectory. 

We present two scenarios of a two-agent case with different levels of deterministic uncertainty controlled using $\tau_i$ but with the same stochastic uncertainty characterized by initial state covariance to be $\diag([0.2,0.2,0.5,0.5])^2$, and covariance of each stochastic noise component $w_i^k$ to be $\diag([0.02,0.02,0.2,0.2])^2$ for each agent $i \in \calV$. 
Fig. \ref{fig5} discloses the first scenario with deterministic uncertainty parameterized by $\tau = 0.01$ for both agents. It can be observed from Fig. \ref{5a} that the mean trajectories obtained by the proposed method in the deterministic case satisfy the inter-agent and the terminal mean constraints. However, Fig. \ref{5b} shows that in the presence of stochastic noise, a collision between the agents is highly probable, and the agents may not be able to reach the targets within the set bounds. Fig. \ref{5c} presents the case where we enforce robust chance constraints on the agents' positions to stay within the blue dashed lines. As shown, most of the realizations of the trajectories lie within the specified bounds of the chance constraints, which ensures the safe operation of the agents despite the stochastic uncertainty. However, the uncertainty in the terminal state of the agents is not controlled by the applied robust chance constraints. We address this using covariance constraints. Fig \ref{5d} presents the scenario with additional terminal covariance constraints with the target covariance $\bSigma^i$ set to be $\diag([0.05, 0.05, 0.5, 0.5])^2$ for each agent $i \in \calV$. It can be observed that the uncertainty in the terminal state is reduced. 

\begin{figure*}[t!] 
    \centering
  \subfloat[Robust Mean Trajectories \label{7a}]{
       \includegraphics[width= 0.32\linewidth, trim={0.2cm 1cm 0.75cm 2.2cm},clip]{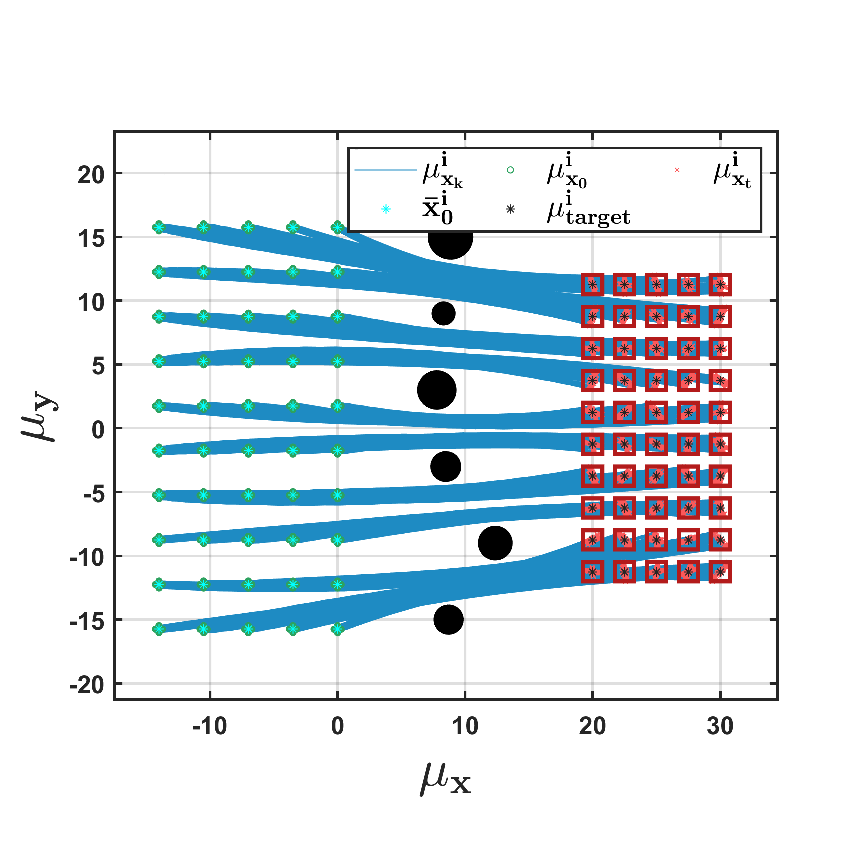}}
  \subfloat[  $ k = 10$ \label{7b}]{
    \includegraphics[width=0.32\linewidth, trim={0.2cm 1cm 0.75cm 2.2cm},clip]{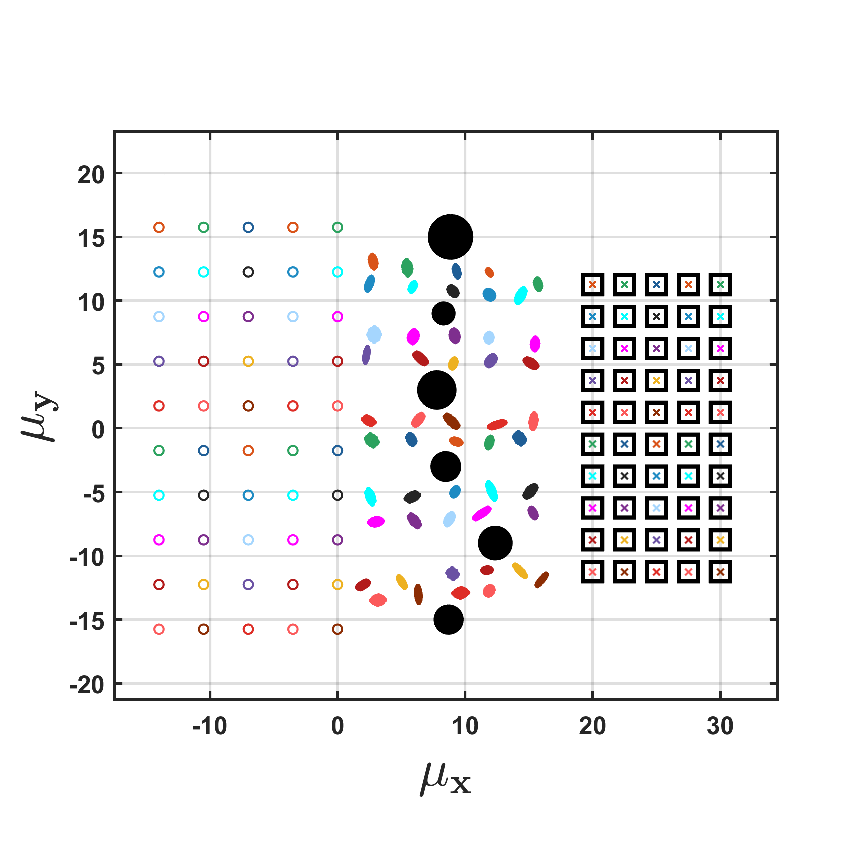}}
  \subfloat[ Distance Plot \label{7c}]{
       \includegraphics[width= 0.3\linewidth, trim={0cm -0.375cm 0cm 0cm},clip]{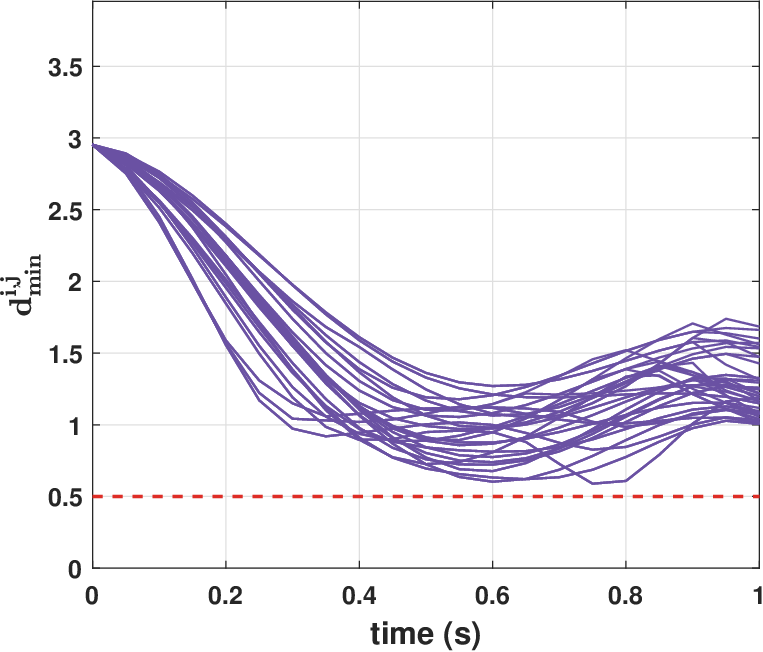}}
    \\
  \caption{\textbf{Fifty agent scenario with six obstacles:} (a) Robust mean trajectories of agents in the deterministic case, (b) Snapshot of robust mean trajectories of agents.
  (c) Distance plot showing the minimum distances of the agent from its nearest neighbor agent at each time step.} 
  \label{fig7} 
\end{figure*}
Fig. \ref{fig6} presents the second scenario of the two-agent case that involves a higher deterministic uncertainty parameterized by $\tau = 0.05$ for both agents. 
Fig. \ref{6b} shows that the uncertainty in the terminal state of the agents might lie outside the target uncertainty bounds (indicated by black boxes). To address this, we can use covariance steering as shown in the previous scenario, or robust chance constraints. We use robust chance constraints in this scenario. Fig. \ref{6c} shows results for the mixed case when robust chance constraints are applied on the terminal state to stay within the target uncertainty bounds represented by the blue dashed box. 

Further, it is interesting to note the following two observations. From Fig. \ref{5a} and \ref{5d}, it should be observed that uncertainty in the state trajectory is reduced to be smaller than the uncertainty in the mean in the deterministic case by adding covariance constraints. From Fig. \ref{6a} and \ref{6b}, it should be observed that the actual trajectories of the agents in the deterministic case avoid the obstacle without enforcing the robust chance constraints. This is because the feedback control gain $\vK$ is used to control the deviations in the state due to the deterministic and stochastic uncertainties. Thus, controlling the deviation in the state due to one type of uncertainty also controls the deviation due to another. 
 % This is achieved using the feedback control gain $\vK$, which essentially controls the deviation in the trajectory due to stochastic noise. 
 This might be useful in cases where one of the uncertainties (deterministic or stochastic) is significantly higher than the other. In such a case, we need not explicitly apply constraints to control the effect of the least pronounced uncertainty.
\begin{figure*}[t!] 
    \centering
  \subfloat[ Robust mean trajectories \label{8a}]{
       \includegraphics[width= 0.32\linewidth, trim={0cm 0cm 0cm 0cm},clip]{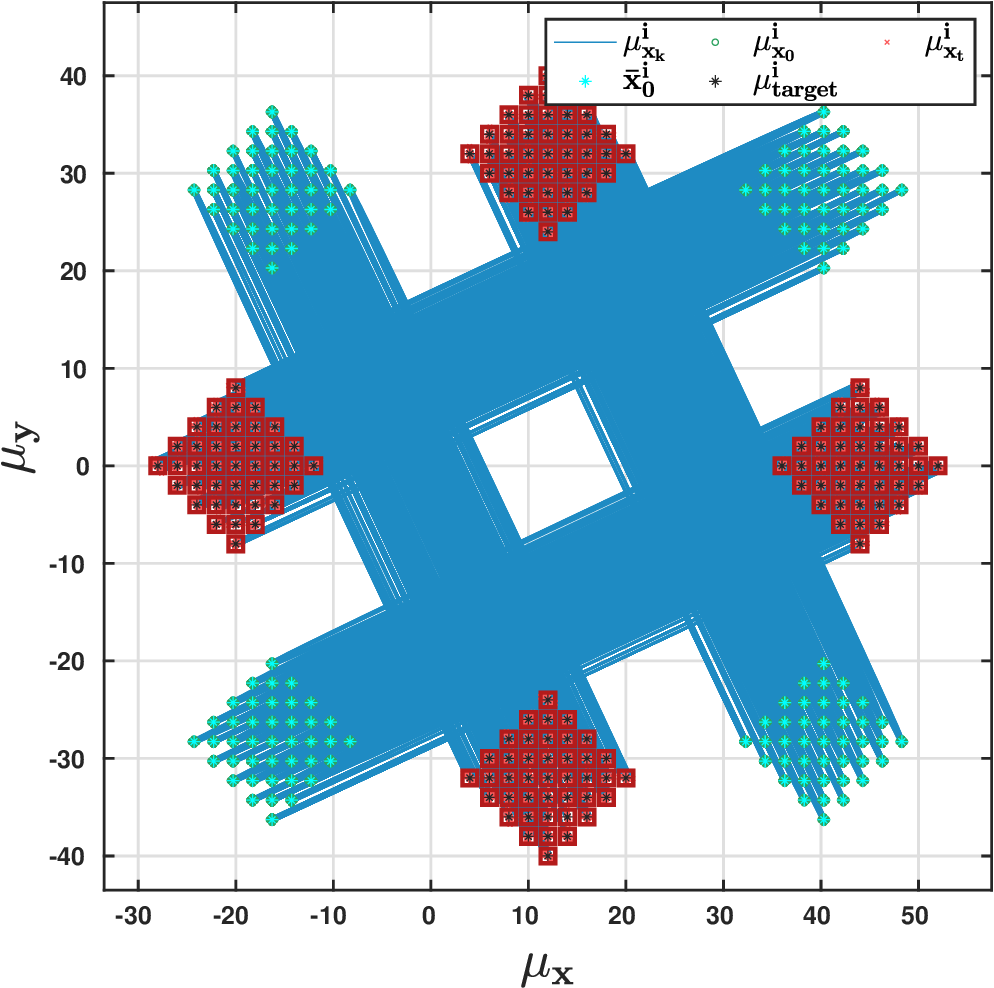}}
  \subfloat[$ k = 12$ \label{8b}]{
       \includegraphics[width= 0.32\linewidth, trim={0cm 0cm 0cm 0cm},clip]{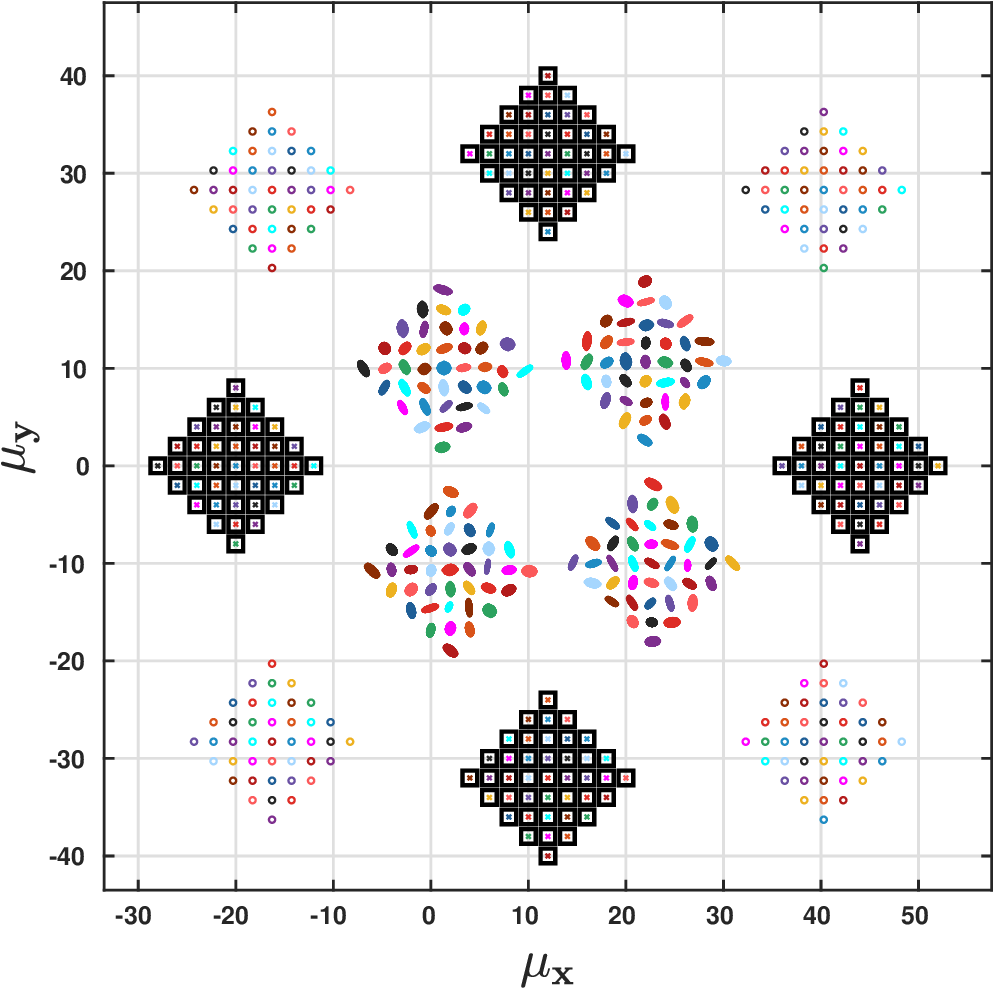}}
\subfloat[ Distance Plot \label{8c}]{
    \includegraphics[width=0.305\linewidth, trim={0cm -0.5cm 0cm 0cm},clip]{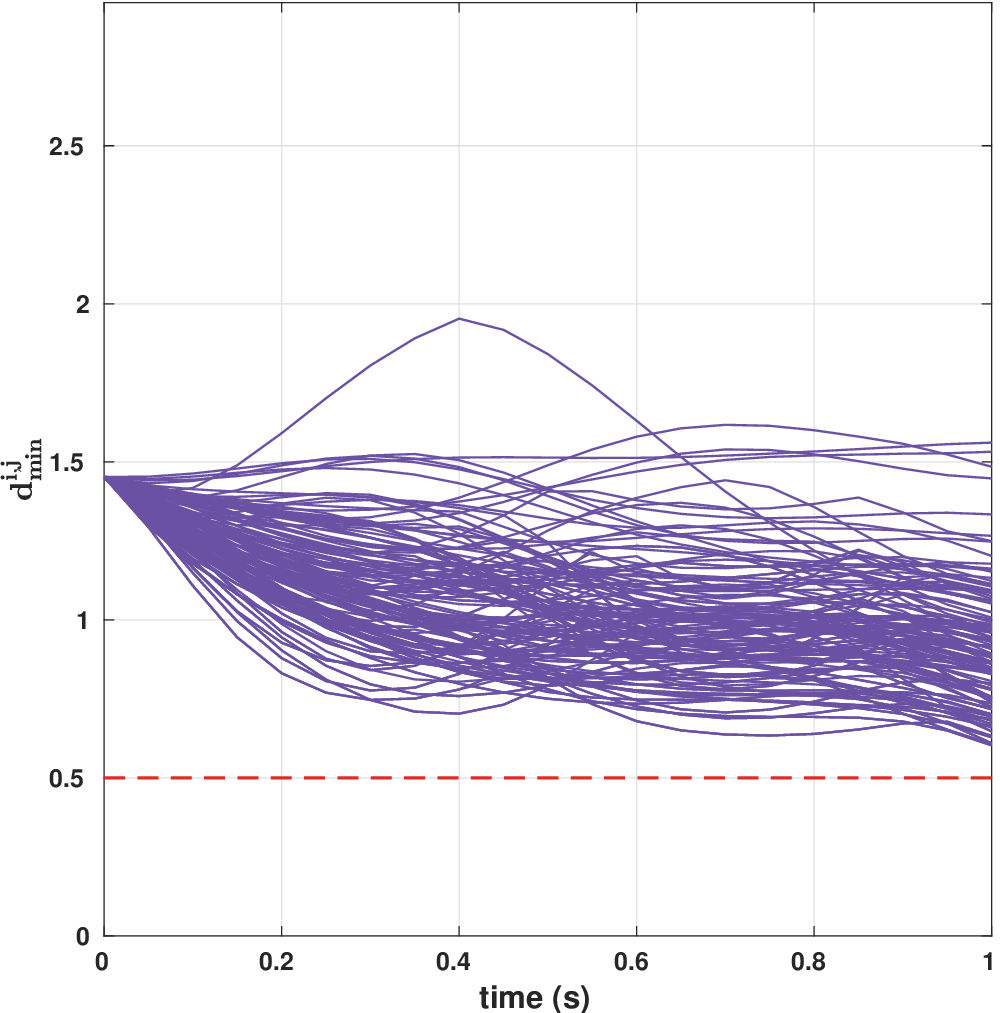}}
    \\
  \caption{\textbf{164-agent scenario:} (a) Mean trajectories of agents in the deterministic case, (b) Snapshot of mean trajectories of agents, (c) Distance plot showing the minimum distances of the agent from its nearest neighbor agent at each time step.} 
  \label{fig8} 
\end{figure*}
\subsection{Scalability of Proposed Distributed Framework}
\label{sec: sims part 2}
As mentioned, the existing robust trajectory optimization approaches are computationally expensive, even for a single-agent case. In this subsection, we show that the proposed distributed framework enables large-scale robust trajectory optimization. In the following experiments, we consider the deterministic case.  

Fig. \ref{fig7} shows a scenario with 50 agents and six obstacles. Each agent has four neighbors. Fig. \ref{7a} and \ref{7b} show that all the agents reached the terminal target mean bounds and did not collide with the obstacles. Fig. \ref{7c} discloses a distance plot of the minimum distance of each agent from its neighbors, with the red dashed line showing the collision distance threshold (at 0.5). It can be observed that the minimum distances are well above the collision distance threshold, proving that there are no inter-agent collisions. Fig. \ref{fig8} shows a scenario with 164 agents dispersed in four sets, each of a diamond formation. The task is for the agents to reach the target mean bounds by staying in the diamond formation and avoiding collisions with the neighbor agents. It can be observed that the agents reach the terminal target mean bounds. Further, Fig. \ref{8b} - \ref{8c} confirm that there are no inter-agent collisions. It is interesting to note from Fig. \ref{7c} and \ref{8c} that the minimum distances in the final time-step are not the same for all the agents, even if the final target positions are equally distanced. This is due to the dissimilarity in the shape of the state mean realizations of different agents (refer to Section III.D of the supplementary material).

\begin{figure}[t!] 
    \centering
  { \includegraphics[width= 0.9\linewidth, trim={0cm 0cm 0cm 0cm},clip]{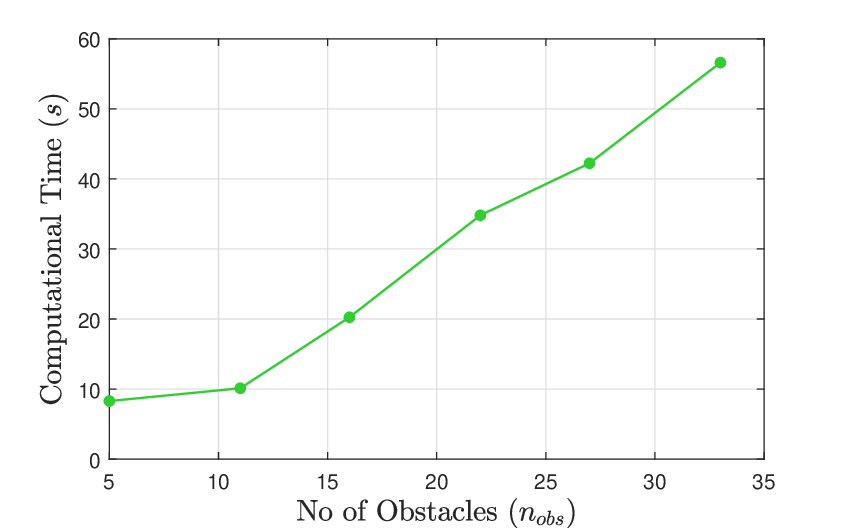}}
    \\
    \caption{\textbf{ Computational time graph w.r.t the number of obstacles.}} 
  \label{fig9} 
\end{figure}

\begin{figure}[t!] 
    \centering
  { \includegraphics[width= 0.9\linewidth, trim={0cm 0cm 0cm 0cm},clip]{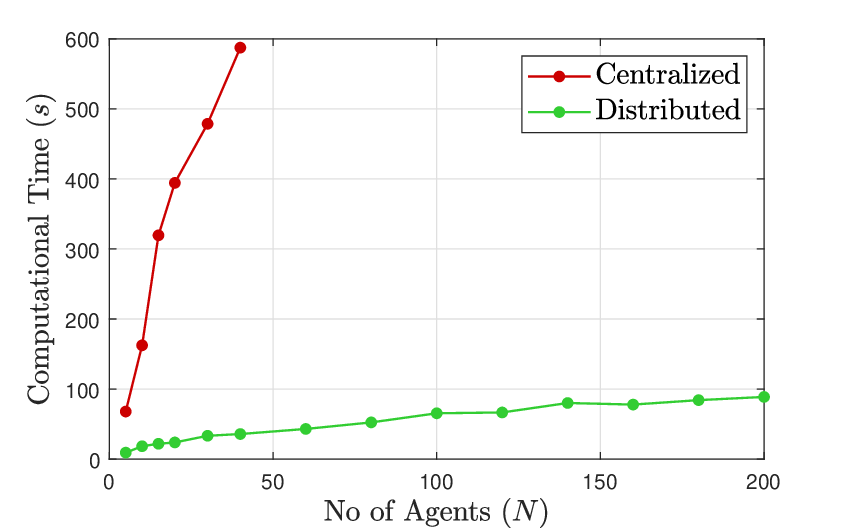}}
    \\
    \caption{\textbf{ Computational time graph w.r.t the number of agents.}}
  \label{fig10} 
\end{figure}

Now, we analyse the computational efficiency of the proposed method. First, we provide the computational time graph with respect to the number of obstacles in Fig. \ref{fig9}.  There are two significant observations. First, the complexity bounds \eqref{complexity bounds} indicate a cubic dependence of the complexity on the number of obstacles. However, from the computational plot, we can infer that, in practice, the complexity is significantly less than the worst-case bound. Second, the increase in the computational time from five obstacles case to the ten obstacles case is less. When fewer obstacles exist, the significant computational burden would be due to the SOCP constraints \eqref{mu_d expression} on $\bmu_d^i$ bounds.
 
Next, we compare the proposed distributed framework with the centralized framework using the proposed reformulated constraints. As we have derived in the earlier section, the complexity of the centralized framework \eqref{centralized complexity bound} is proportional to the cube of the number of agents,  while the complexity of the distributed framework does not directly depend on the number of agents. This is justified by the computational graph in Fig. \ref{fig10}, where we can observe that the centralized framework does not scale favorably.

 In all the above cases, the computational time for the distributed framework is estimated as follows. In each ADMM iteration, the maximum optimizer time among all the agents is considered. These per-iteration maximum time values are added over all the ADMM iterations that run until convergence. Further details about the scenarios considered for the creation of the plots can be found in Appendix-N.
 All simulations were carried out in Matlab2022b \cite{MATLAB} environment using CVX \cite{cvx} as the modeling software and MOSEK as the solver \cite{mosek} on a system with an Intel Corei9-13900K.

\section{Conclusion} 
\label{sec:conclusion}

% In this paper, we integrate principles from the areas of robust optimization, distribution steering and distributed optimization towards obtaining a scalable safe multi-agent control method under both stochastic and deterministic uncertainties. In particular, we present a distributed robust optimization method that exploits computationally tractable approximations of robust constraints and the separable nature of ADMM in order to significantly reduce computational demands. The results verify that the proposed framework is able to effectively address both types of uncertainty and scale for multi-robot systems with up to a hundred agents.
This paper presents a novel approach for multi-agent control under stochastic and deterministic uncertainty, integrating robust optimization, distribution steering, and distributed optimization principles. We developed computationally tractable approximations of robust constraints that significantly reduce the complexity compared to existing methods without compromising safety. Subsequently, we proposed a distributed framework leveraging the proposed constraint reformulations and the separable nature of ADMM, to solve the computationally challenging problem of multi-agent robust trajectory optimization. We also established rigorous computational complexity bounds for the proposed distributed framework, 
% and comparative analyses with the centralized approach 
highlighting the framework's increased computational efficiency compared to centralized approaches. The simulation results demonstrate the scalability of the proposed framework, successfully handling up to 164 agents, and empirically validate the established complexity bounds. 
% A computational time graph comparing the proposed distributed framework with the centralized approach corroborates the established complexity bounds.
Additionally, we present extensive simulations across diverse scenarios, including non-convex obstacle environments, validating the robustness and computational efficiency of the proposed constraint reformulations. 

In future work, we wish to explore new distributed robust optimization algorithms that further enhance the safety capabilities, computational efficiency and applicability of such appraoches in multi-agent robotics problems. In particular, we plan to extend the current method for addressing nonlinear dynamics through considering sequential linearization schemes \cite{ridderhof2019nonlinear, saravanos2024distributed_iros} or combinations with shooting techniques \cite{jacobson1970differential, von1992direct}. Additionally, we intend to explore distributed learning-to-optimize approaches for accelerating convergence \cite{saravanos2024deep}, as well as data-driven techniques for learning the uncertainty sets of the agents \cite{wang2023learning}. Finally, incorporating the current framework into hierarchical distributed schemes \cite{Saravanos-RSS-23} would be a promising direction for further increasing scalability.

% \section*{Acknowledgments}
\section*{Acknowledgments}

This work was supported by the ARO Award $\#$W911NF2010151. Augustinos Saravanos acknowledges financial support by the A. Onassis Foundation Scholarship.

%% Use plainnat to work nicely with natbib. 

\bibliographystyle{plainnat}
\bibliography{references}

\appendix
The following serves as supplementary material for the main paper. 
\subsection{System Dynamics Compact Form}
The system dynamics (1) can be written in a more compact form as follows
\begin{equation}
\bx^i = \vG_0^i \bar{x}_0^i + \vG_u^i \bu^i + \vG_{\bw}^i \boldsymbol{\bw}^i + \vG_{\zeta}^i \boldsymbol{\zeta}^i,
\label{dynamics_SM_compact}
\end{equation} 
where 
\begin{subequations}
\allowdisplaybreaks
\begin{align}
&
\vG_0^i = 
\begin{bmatrix}
\vI; & 
\vPhi^i(1,0) ; & 
\vPhi^i(2,0) ; &
\dots ; &
\vPhi^i(T,0)
\end{bmatrix}, 
%\in \Rb^{(T+1) n_x \times n_x }, 
\\[0.2cm]
&
\vG_u^i =  \begin{bmatrix}
\vzero & \vzero & \dots & \vzero \\ 
B_0^i & \vzero & \dots & \vzero \\ 
\vPhi^i(2,1) B_0^i & B_1^i & \dots & \vzero \\ 
\vdots & \vdots & \vdots & \vdots\\ 
\vPhi^i(T,1) B_0^i & \vPhi^i(T,2) B_1^i & \dots & B_{T-1}^i
\end{bmatrix}, 
\\[0.2cm]
&
\vG_{\bw}^i = 
\begin{bmatrix}
\vI & \vzero & \dots & \vzero \\ 
\vPhi^i(1,0) & D_0^i & \dots & \vzero \\ 
\vPhi^i(2,0) & \vPhi^i(2,1) D_0^i & \dots & \vzero \\ 
\vdots & \vdots & \vdots &  \vdots\\ 
\vPhi^i(T,0) & \vPhi^i(T,1) D_0^i & \dots & D_{T-1}^i
\end{bmatrix}, 
\\[0.2cm]
&
\vG_{\zeta}^i = 
\begin{bmatrix}
\vI & \vzero & \dots & \vzero \\ 
\vPhi^i(1,0) & C_0^i & \dots & \vzero \\ 
\vPhi^i(2,0) & \vPhi^i(2,1) C_0^i & \dots & \vzero \\ 
\vdots & \vdots & \vdots &  \vdots\\ 
\vPhi^i(T,0) & \vPhi^i(T,1) C_0^i & \dots & C_{T-1}^i
\end{bmatrix}, 
\end{align}
\end{subequations}
with 
%
% \begin{subequations}
\begin{align}
&
\vPhi^i(k_1,k_2) = A_{k_1 -1}^i A_{k_1-2}^i \dots A_{k_2}^i \;
\text{for} \; k_1 > k_2. 
% & \vPhi^i(k_1,k_1) = \vI_{n_{x_i} \times n_{x_i}}.
\end{align}
\section{Robust Linear Constraints}
\subsection{{Expectation of State}}
Let us rewrite the state equation for an agent $i$ as follows
\begin{align}
    {\bx^i = \vG_0^i \bar{x}_0^i + \vG_u^i \bu^i + \vG_{\bw}^i \boldsymbol{\bw}^i + \vG_{\zeta}^i \boldsymbol{\zeta}^i \nonumber
}\end{align}
Substituting the control $\bu^i$ as $\bar{\bu}^i + \vK^i \bdelta^i$ (from (10)) in the above state equation, we get the following.
%DIF > 
\begin{align}
    {\bx^i = \vG_0^i \bar{x}_0^i + \vG_u^i \big(\bar{\bu}^i + \vK^i \bdelta^i \big) + \vG_{\bw}^i \boldsymbol{\bw}^i + \vG_{\zeta}^i \boldsymbol{\zeta}^i
}\end{align}
%DIF > 
Further, using (8), we can rewrite the above equation as follows
\begin{equation}
\begin{aligned}
    \bx^i & = \vG_0^i \bar{x}_0^i 
    + \vG_u^i \big(\bar{\bu}^i 
    + \vK^i (\vG_{\bw}^i \boldsymbol{\bw}^i 
    + \vG_{\zeta}^i \boldsymbol{\zeta}^i )   \big) \\
    &~~
    + \vG_{\bw}^i \boldsymbol{\bw}^i + \vG_{\zeta}^i \boldsymbol{\zeta}^i
\end{aligned}    
\end{equation}
which can be written in compact form as follows -
\begin{equation}
{\begin{aligned}
    \bx^i & = \vG_0^i \bar{x}_0^i 
    + \vG_u^i \bar{\bu}^i 
    + (\vG_u^i \vK^i + \vI) \vG_{\bw}^i \boldsymbol{\bw}^i \\
    &~~~
    + (\vG_u^i \vK^i + \vI) \vG_{\zeta}^i \boldsymbol{\zeta}^i  
\end{aligned}
\label{state equation in terms of zeta}
}\end{equation}
Now, the expectation of the state can be given as 
\begin{equation}
    {\begin{aligned}
    E[\bx^i] & = \vG_0^i \bar{x}_0^i 
    + \vG_u^i \bar{\bu}^i 
    + (\vG_u^i \vK^i + \vI) \vG_{\bw}^i E[ \boldsymbol{\bw}^i] \\
    &~~~
    + (\vG_u^i \vK^i + \vI) \vG_{\zeta}^i \boldsymbol{\zeta}^i
    \end{aligned}
}\end{equation}
Finally, using the fact that $E[\boldsymbol{\bw}^i] = 0$, we get
\begin{equation}
    E[{\bx^i}] = \vG_0^i \bar{x}_0^i 
    + \vG_u^i \bar{\bu}^i 
    + (\vG_u^i \vK^i + \vI) \vG_{\zeta}^i \boldsymbol{\zeta}^i
    \label{expectation of state SM}
\end{equation}

\subsection{Covariance of State}
The covariance of the state of an agent $i$ is given as follows
\begin{equation}
    \Cov[\bx^i] = \Eb \Big[ (\bx^i - \Eb[\bx^i]) (\bx^i - \Eb[\bx^i])\T \Big]
\end{equation}
Let us first write the expression for $\bx^i - \Eb[\bx^i]$ using \eqref{state equation in terms of zeta} and \eqref{expectation of state SM} as follows 
\begin{equation}
    \bx^i - \Eb[\bx^i] = (\vG_u^i \vK^i + \vI) \vG_{\bw}^i \boldsymbol{\bw}^i
\end{equation}
{Now, the covariance of the state can be rewritten as -
}\begin{equation}
    {\Cov[\bx^i] = \Eb \Big[ (\vG_u^i \vK^i + \vI) \vG_{\bw}^i \boldsymbol{\bw}^i ((\vG_u^i \vK^i + \vI) \vG_{\bw}^i \boldsymbol{\bw}^i)\T \Big]
}\end{equation}
%DIF > 
{which can be further rewritten as
}\begin{equation}
    {\Cov[\bx^i] = (\vG_u^i \vK^i + \vI) \vG_{\bw}^i 
    \Eb }[ {\boldsymbol{\bw}^i \boldsymbol{\bw}^i }{}{\T }]
    {\vG_{\bw}^i }{}{\T (\vG_u^i \vK^i + \vI)\T 
}\end{equation}
%DIF > 
{Finally, using the fact that $\Eb [ \boldsymbol{\bw}^i \boldsymbol{\bw}^i {}\T ] = \bSigma_{\bw^i}$, we get 
%DIF > 
}\begin{equation}
    {\Cov[\bx^i] = (\vG_u^i \vK^i + \vI) \vG_{\bw}^i \bSigma_{\bw^i}
    \vG_{\bw}^i }{}{\T (\vG_u^i \vK^i + \vI)\T 
}\end{equation}
%DIF > 
 \subsection{Proposition 1 Proof}
% \begin{equation}
%     \max_{\bzeta^i \in \calU_i} \ba_i \T \Eb[\bx^i] =
%     % (\tau^i)^{1/2} \| \vP^i \T \vG_{\zeta}^i \T (\vG_u^i \vK^i + \vI) \T \ba\|_{\vS_i^{-1}}
%     \ba_i \T \bmu_{x,\bar{u}}^i + (\tau^i)^{1/2} \| \boldsymbol{\Gamma}_i \T \vM_i \T \ba_i\|_{\vS_i^{-1}}
% \end{equation}
%
The expression $\max_{\bzeta^i \in \calU_i} \ba_i \T \Eb[\bx^i]$ can be rewritten as
\begin{subequations}
\begin{align}
\max_{\bzeta^i \in \calU_i} \ba_i \T \Eb[\bx^i] & =
\ba_i \T \bmu_{x,\bar{u}}^i 
+ \max_{\bzeta^i \in \calU_i} \ba_i \T \vM_i \bzeta^i \\
& =
\ba_i \T \bmu_{x,\bar{u}}^i 
- \min_{\langle \vS_i \bz_i, \bz_i \rangle \leq \tau^i} 
- \ba_i \T \vM_i \boldsymbol{\Gamma}_i \bz_i .
\label{prop 1 min term}
\end{align}
\end{subequations}
Let us now consider the Lagrangian for the minimization problem inside \eqref{prop 1 min term}, given by 
\begin{align}
    \calL(\bz_i,\lambda) =  - \ba_i \T \vM_i \boldsymbol{\Gamma}_i \bz_i 
    + \lambda (\langle \vS_i \bz_i, \bz_i \rangle - \tau^i),
\end{align}
where $\lambda \in \Rb$ is the Lagrange multiplier corresponding to the inequality constraint $\langle \vS_i \bz_i, \bz_i \rangle \leq \tau^i$. 
The above problem is convex, and the constraint set $\langle \vS_i \bz_i, \bz_i \rangle \leq \tau^i$ is non-empty - since $\bz_i = 0$ lies in the interior of the feasible set because $\tau^i > 0$. Thus, the problem satisfies Slater's condition, and we could find the minimizer using KKT conditions as follows
\begin{subequations}
\begin{align}
& \nabla_{\bz_i} \calL (\bz_i^*, \lambda^*) = 0, 
\\
& \lambda^* (\langle \vS_i \bz_i^*,  \bz_i^* \rangle - \tau^i) = 0,
\label{KKT condition 2}
\\
& \lambda^* \geq 0, 
\\
& \langle \vS_i \bz_i^*, \bz_i^* \rangle - \tau^i \leq 0. 
\end{align}
\end{subequations}
Let us first simplify the condition $\nabla_{\bz_i}  \calL (\bz_i^*, \lambda^*) = 0$, as 
\begin{equation}
- \boldsymbol{\Gamma}_i \T \vM_i \T \ba_i + 2 \lambda^* \vS_i \bz_i^* = 0,
\end{equation}
%
% & 
% 2 \lambda^* \vS_i \bz_i^* 
% = \boldsymbol{\Gamma}_i \T \vM_i \T \ba_i \\
%
which leads to
\begin{equation}
2 \lambda^* \bz_i^* = \vS_i^{-1} \boldsymbol{\Gamma}_i \T \vM_i \T \ba_i.
\end{equation}
It can be observed that $\boldsymbol{\Gamma}_i \T \vM_i \T \ba_i$ cannot be equal to the zero vector since it is the coefficient of $\bz_i$ in the objective function. Thus, since we also have that $\vS_i \in \Sb_{++}^{\bar{n}_i}$, then the RHS of the above equation is non-zero, which implies that both $\bz_i^*$ and $\lambda^*$ need to be non-zero. Therefore, we obtain
\begin{equation}
\bz_i^* 
= (2 \lambda^*)^{-1} \vS_i^{-1} \boldsymbol{\Gamma}_i \T \vM_i \T \ba_i.
\label{prop 1 proof z_i star}
\end{equation}
Further, from the condition (\ref{KKT condition 2}) and using $\lambda^* \neq 0 $, we get
\begin{equation}
\langle \vS_i \bz_i^*, \bz_i^* \rangle - \tau^i = 0,
\end{equation}
which through \eqref{prop 1 proof z_i star} gives
\begin{equation}
(2 \lambda^*)^{-2} 
\langle \boldsymbol{\Gamma}_i \T \vM_i \T \ba_i, 
\; \vS_i^{-1} \boldsymbol{\Gamma}_i \T \vM_i \T \ba_i \rangle 
= \tau^i,
\end{equation}
%
% \begin{align}
    % & 
    % \langle \vS_i (2 \lambda^*)^{-1} \vS_i^{-1} \boldsymbol{\Gamma}_i \T \vM_i \T \ba_i, 
    % \; (2 \lambda^*)^{-1} \vS_i^{-1} \boldsymbol{\Gamma}_i \T \vM_i \T \ba_i \rangle = \tau^i 
    % \\
 %    \implies &
 % \\
 %    \implies &
    % (\tau^i)^{-1} \; \langle \boldsymbol{\Gamma}_i \T \vM_i \T \ba_i, 
    % \; \vS_i^{-1} \boldsymbol{\Gamma}_i \T \vM_i \T \ba_i \rangle 
    % = (2 \lambda^*)^{2} 
    % \label{robust linear one eq 1} \\
    % \implies &
    % (\tau^i)^{-1} \; 
    % \| \boldsymbol{\Gamma}_i \T \vM_i \T \ba_i \|_{\vS_i^{-1}}^2
    % = (2 \lambda^*)^{2} \\
    % \implies & 
%
The above relationship leads to
\begin{equation}
(\tau^i)^{-1/2} \; \| \boldsymbol{\Gamma}_i \T \vM_i \T \ba_i \|_{\vS_i^{-1}}
= 2 \lambda^*,    
\end{equation}
%
% \end{align}
%
since $\lambda^* > 0$.
% Above, we used the fact that $\lambda^* > 0$.
%
The optimal value $\calL (\bz_i^*, \lambda^*)$ is then given by 
\begin{subequations}    
\begin{align}
    \calL (\bz_i^*, \lambda^*) & = - \ba_i \T \vM_i \boldsymbol{\Gamma}_i \bz_i^* \\
    & =
    - (2 \lambda^*)^{-1} \ba_i \T \vM_i \boldsymbol{\Gamma}_i \vS_i^{-1} \boldsymbol{\Gamma}_i \T \vM_i \T \ba_i \\
    & =
    - (2 \lambda^*)^{-1} \| \boldsymbol{\Gamma}_i \T \vM_i \T \ba_i \|_{\vS_i^{-1}}^2 \\
    & =
    - (\tau^i)^{1/2} \| \boldsymbol{\Gamma}_i \T \vM_i \T \ba_i\|_{\vS_i^{-1}}.
    \label{robust linear one eq 2}
\end{align}
\end{subequations}
Therefore, we obtain 
\begin{subequations}
    \begin{align}
    \max_{\bzeta^i \in \calU_i} \ba_i \T \Eb[\bx^i] & =
    \ba_i \T \bmu_{x,\bar{u}}^i 
    - \calL (\bz_i^*, \lambda^*) \\ 
    & =
    \ba_i \T \bmu_{x,\bar{u}}^i 
    + (\tau^i)^{1/2} \| \boldsymbol{\Gamma}_i \T \vM_i \T \ba_i\|_{\vS_i^{-1}}.
\end{align}
\end{subequations}
\section{Robust Non-convex Constraints and Distributed Architecture}
\subsection{Initial Approach for robust nonconvex constraint}

Let us rewrite the robust nonconvex norm-of-mean constraint ( {26} ) as follows
\begin{align}
    & \min_{\bzeta^i \in \calU_i} \| \vA_i \vP_k^i \bmu_{x,\bar{u}}^i 
    +  \vA_i \vP^i_k \vM_i \bzeta^i - \bm{b}_i \|_2 \geq c_i,
    % & 
    % \min_{\bzeta^i \in \calU_i} \| \vA_i \vP_k^i \bmu_{x,\bar{u}}^i 
    % - \bm{b}_i
    % +  \vA_i \vP^i_k \vM_i \bzeta^i  \|_2 \geq c_i 
    \label{obstacle constraint 1}
\end{align}
which can be written in terms of $\bzeta^i$ as
\begin{equation}
\begin{aligned}
\min_{\bzeta^i \in \calU_i}  
\bzeta^i {}\T \vF_{1,k}^i(\vK^i) \bzeta^i
+ 2 \vF_{2,k}^i (\vK^i, & \bar{\bu}^i) \bzeta^i
\\[-0.1cm]
& + \vF_{3,k}^i (\bar{\bu}^i) \geq c_i^2
\end{aligned}
    \label{nonconvex constraints 1}
\end{equation}
where 
\begin{subequations}
\begin{align}
\vF_{1,k}^i(\vK^i) & = (\vA_i \vP^i_k \vM_i) \T \vA_i \vP^i_k \vM_i, 
\\
\vF_{2,k}^i (\vK^i, \bar{\bu}^i) & =  (\vA_i \vP_k^i \bmu_{x,\bar{u}}^i - \bm{b}_i) \T (\vA_i \vP^i_k \vM_i),
\\
\vF_{3,k}^i (\bar{\bu}^i) & = \| \vA_i \vP_k^i \bmu_{x,\bar{u}}^i - \bm{b}_i \|_2^2.
\end{align}
\end{subequations}
The constraint (\ref{nonconvex constraints 1}) could be reformulated through the following equivalent set of constraints (see \cite{kotsalis2020convex}),
\begin{subequations}    
\begin{align}
&
  \begin{bmatrix}
    \boldsymbol{\Gamma}_i \T  \vF_{1,k}^i(\vK^i) \boldsymbol{\Gamma}_i + \alpha \vS_i 
    & \boldsymbol{\Gamma}_i \T \vF_{2,k}^i (\vK^i, \bar{\bu}^i)  \T  \\
    \vF_{2,k}^i (\vK^i, \bar{\bu}^i) \boldsymbol{\Gamma}_i & \vF_{3,k}^i ( \bar{\bu}^i) - c_i^2 - \alpha \tau^i
\end{bmatrix} 
 \succeq 0,
 \\[0.2cm]
 & ~~~~~~~~~~~~~~~~~~~~~~~~~~~ \alpha \geq 0.
\end{align}
\end{subequations}
Nevertheless, the above constraint would still be a nonconvex constraint, as $\vF_{1,k}^i(\vK^i)$, $\vF_{2,k}^i (\vK^i, \bar{\bu}^i)$, and $\vF_{3,k}^i ( \bar{\bu}^i)$ are nonlinear in $\vK^i$, $\bar{\bu}^i$. To address this issue, we first linearize the constraint (\ref{obstacle constraint 1}) with respect to $\vK^i, \bar{\bu}^i$ around the nominal $\vK^{i,l}, \bar{\bu}^{i,l}$, to obtain the following affine robust constraint
\begin{equation}    
  \begin{aligned}
 \boldsymbol{\zeta}^i {}\T \mathcal{Q} ( \vK^i, \vK^{i,l})
\boldsymbol{\zeta}^i 
+ 2 & \bar{\mathcal{Q}} ( \vK^i, \vK^{i,l}, \Bar{\bu}^i, \Bar{\bu}^{i,l})  \boldsymbol{\zeta}^i \\
& + q(\Bar{\bu}^i, \Bar{\bu}^{i,l})
\geq c_i^2, \quad \forall \; \boldsymbol{\zeta}^i \in \calU_i
\end{aligned} 
\end{equation}
where
\begin{subequations}   
\begin{align}
& \mathcal{Q} = (\vA_i \vP^i_k \vM_i^l) \T \vA_i \vP^i_k \vM_i 
\nonumber
\\
& ~~~~~~~~~~~~~~~~~ + (\vA_i \vP^i_k (\vM_i - \vM_i^l)) \T \vA_i \vP^i_k \vM_i^l,
\\[0.2cm]
& \bar{\mathcal{Q}}  = (\vA_i \vP_k^i \bmu_{x,\bar{u}}^{i,l} - \bm{b}_i) \T (\vA_i \vP^i_k \vM_i) 
\nonumber
\\
& ~~~~~~~~~~~~~~~~~ + (\vA_i \vP_k^i (\bmu_{x,\bar{u}}^i - \bmu_{x,\bar{u}}^{i,l})) \T (\vA_i \vP^i_k \vM_i^l),
\\[0.2cm]
& q = (\vA_i \vP_k^i \bmu_{x,\bar{u}}^{i,l} - \bm{b}_i) \T (\vA_i \vP_k^i (2\bmu_{x,\bar{u}}^{i}-\bmu_{x,\bar{u}}^{i,l})  - \bm{b}_i ).
\end{align}
\end{subequations}
\subsection{Proposition 2 Proof}
Let us initially rewrite the constraint ( {33} ) as follows
\begin{equation}
\max_{\bzeta^i \in \calU_i} \| \vA_i \tilde{\vM}_k^i \bzeta^i \|_2
 \leq   \tilde{c}_i - c_i  .
\label{interagent constraint 1}
\end{equation}
The first step is to construct an upper bound for $\| \vA_i \tilde{\vM}_k^i \bzeta^i \|_2^2$. Using Proposition 1, we obtain the following two equalities,
\begin{align}
\max_{\bzeta^i \in \calU_i} h_{k,\bar{m}}^i {}\T \vM_i \bzeta^i 
& = (\tau^i)^{1/2} \| \boldsymbol{\Gamma}_i \T \vM_i \T h_{k,\bar{m}}^i \|_{\vS_i^{-1}}, 
\label{prop 2 proof max}
\\
\min_{\bzeta^i \in \calU_i} h_{k,\bar{m}}^i {}\T \vM_i \bzeta^i 
& = - (\tau^i)^{1/2} \| \boldsymbol{\Gamma}_i \T \vM_i \T h_{k,\bar{m}}^i \|_{\vS_i^{-1}},
\label{prop 2 proof min}
\end{align}
where $h_{k,\bar{m}}^i {}\T$ is $\bar{m}^{th}$ row of $\vA_i \vP_k^i$, for all $\bar{m} = 1,\dots, m$. 
By combining \eqref{prop 2 proof max} and \eqref{prop 2 proof min} for every row of $\vA_i \tilde{\vM}_k^i \bzeta^i$, we can arrive to the following relationship,
\begin{subequations}
\begin{align}
    \max_{\bzeta^i \in \calU_i}  \| \vA_i \tilde{\vM}_k^i \bzeta^i \|_2 
    & = 
    \max_{\bzeta^i \in \calU_i} \bigg(  \sum_{\bar{m} = 1}^{m} 
    (h_{k,\bar{m}}^i {}\T \vM_i \bzeta^i)^2 \bigg)^{1/2} \\
    & 
    \leq 
    \bigg( \sum_{\bar{m} = 1}^{m} 
    \max_{\bzeta^i \in \calU_i} (h_{k,\bar{m}}^i {}\T \vM{_i }\bzeta^i)^2 \bigg)^{1/2}\\
    & ~~ =
    \bigg( \sum_{\bar{m} = 1}^{m} \tau^i \| \boldsymbol{\Gamma}_i \T \vM_i \T h_{k,\bar{m}}^i \|_{\vS_i^{-1}}^2 
     \bigg)^{1/2}
     \label{obstacle constraint norm}
\end{align} 
\end{subequations}
Next, by introducing a variable $\bmu_{d,k}^i$ such that 
\begin{equation}
(\tau^i)^{1/2} \| \boldsymbol{\Gamma}_i \T \vM_i \T h_{k,\bar{m}}^i\|_{\vS_i^{-1}} 
\leq (\bmu_{d,k}^i)_{\bar{m}},
\end{equation}
for every $\bar{m} = 1, \dots, m$, we obtain
%
% \begin{align}
%     \forall \bzeta^i \in \calU_i, \quad & 
%     -(\bmu_{d,k}^i)_{\bar{m}}
%     \leq
%     h_{k,\bar{m}}^i {}\T \vM_i \bzeta^i 
%     \leq 
%     (\bmu_{d,k}^i)_{\bar{m}} 
%     \label{obstacle constraint bounds}
%     %
%     % - \bmu_{d,k}^i 
%     % \leq \vA_i \tilde{\vM}_k^i \bzeta^i
%     % \leq \bmu_{d,k}^i 
% \end{align}
% %
% %
% Combining (\ref{obstacle constraint bounds}) and (\ref{obstacle constraint norm}), we get
\begin{align}
    { \max_{\bzeta^i \in \calU_i}
    \| \vA_i \tilde{\vM}_k^i \bzeta^i \|_2  \leq  \|\bmu_{d,k}^i {} \|_2} .
\end{align}
Therefore, we can consider the following tighter approximation of the constraint (\ref{interagent constraint 1}), given by
\begin{align}
    \| \bmu_{d,k}^i {} \|_2  \leq \tilde{c}_i - c_i.
\end{align}
\subsection{Proposition 3 Proof}
The proof of Proposition 3 is similar to the one of Proposition 2. First, let us rewrite constraint ( {40} ) as follows
\begin{align}
    \max_{\bzeta^i \in \calU_i, ~ \bzeta^j \in \calU_j}
         \| \vA_i \tilde{\vM}_k^i \bzeta^i 
         -  \vA_j \tilde{\vM}_k^j \bzeta^j \|_2  
          \leq  \tilde{c}_{ij} - c_{ij} .
        \label{interagent constraint 2}
\end{align}
The bound on each $\bar{m}^{th}$ row of $\vA_i \tilde{\vM}_k^i \bzeta^i 
         -  \vA_j \tilde{\vM}_k^j \bzeta^j$, with $\bar{m} = 1,\dots,m$, can be given as follows
\begin{subequations}
 \begin{align}
    \max_{\bzeta^i \in \calU_i, ~ \bzeta^j \in \calU_j} (\vA_i \tilde{\vM}_k^i \bzeta^i 
         -  \vA_j \tilde{\vM}_k^j \bzeta^j)_{\bar{m}} 
    = \bmu_{max,\bar{m}}^i - \bmu_{min, \bar{m}}^j \\
    \min_{\bzeta^i \in \calU_i, ~ \bzeta^j \in \calU_j} (\vA_i \tilde{\vM}_k^i \bzeta^i 
    -  \vA_j \tilde{\vM}_k^j \bzeta^j)_{\bar{m}} 
    = \bmu_{min,\bar{m}}^i - \bmu_{max, \bar{m}}^j
    % = \max_{\bzeta^i \in \calU_i} 
    % (\vA_i \tilde{\vM}_k^i \bzeta^i)_{\bar{m}} 
    % - \min_{\bzeta^j \in \calU_j}  
    % (\vA_j \tilde{\vM}_k^j \bzeta^j)_{\bar{m}}
\end{align}   
\label{inter agent combined bound}
\end{subequations}
where
\begin{equation}
\begin{aligned}
    \bmu_{min,\bar{m}}^i & = \min_{\bzeta^i \in \calU_i} 
    (\vA_i \tilde{\vM}_k^i \bzeta^i)_{\bar{m}}, \\
    \bmu_{max,\bar{m}}^i & = \max_{\bzeta^i \in \calU_i} 
    (\vA_i \tilde{\vM}_k^i \bzeta^i)_{\bar{m}}, \\
    \bmu_{min, \bar{m}}^j & = \min_{\bzeta^j \in \calU_j}  
    (\vA_j \tilde{\vM}_k^j \bzeta^j)_{\bar{m}}, \\
    \bmu_{max, \bar{m}}^j & = \max_{\bzeta^j \in \calU_j}  
    (\vA_j \tilde{\vM}_k^j \bzeta^j)_{\bar{m}}.
\end{aligned} 
\end{equation}
Using Proposition 1, we have 
\begin{subequations}
\begin{align}
    &
    \bmu_{max,\bar{m}}^i 
    = - \bmu_{min,\bar{m}}^i 
    =  (\tau^i)^{1/2} \| \boldsymbol{\Gamma}_i \T \vM_i \T h_{k,\bar{m}}^i\|_{\vS_i^{-1}} \\
    &
    \bmu_{max,\bar{m}}^j 
    = - \bmu_{min,\bar{m}}^j
    =  (\tau^j)^{1/2} \| \boldsymbol{\Gamma}_j \T \vM_j \T h_{k,\bar{m}}^j \|_{\vS_j^{-1}}
\end{align}
\label{inter-agent bounds on each}%
\end{subequations}
where $h_{k,\bar{m}}^i {}\T, h_{k,\bar{m}}^j {}\T$ are the $\bar{m}^{th}$ rows of the matrices $\vA_i \vP_k^i, \vA_j \vP_k^j$ respectively.
% Using Proposition 1, we obtain -
% \begin{align}
% &
% \begin{aligned}
%     \max_{\bzeta^i \in \calU_i} h_{k,\bar{m}}^i {}\T \vM_i \bzeta^i 
%     & = (\tau^i)^{1/2} \| \boldsymbol{\Gamma}_i \T \vM_i \T h_{k,\bar{m}}^i \|_{\vS_i^{-1}}, \\
%     %
%     \min_{\bzeta^i \in \calU_i} h_{k,\bar{m}}^i {}\T \vM_i \bzeta^i 
%     & = - (\tau^i)^{1/2} \| \boldsymbol{\Gamma}_i \T \vM_i \T h_{k,\bar{m}}^i \|_{\vS_i^{-1}},
% \end{aligned} \\
%     &
%     \begin{aligned}
%     \max_{\bzeta^j \in \calU_j} h_{k,\bar{m}}^j {}\T \vM_i \bzeta^j 
%     & = (\tau^j)^{1/2} \| \boldsymbol{\Gamma}_j \T \vM_j \T h_{k,\bar{m}}^j \|_{\vS_j^{-1}}, \\
%     %
%     \min_{\bzeta^j \in \calU_j} h_{k,\bar{m}}^j {}\T \vM_i \bzeta^j 
%     & = - (\tau^j)^{1/2} \| \boldsymbol{\Gamma}_j \T \vM_j \T h_{k,\bar{m}}^j \|_{\vS_j^{-1}},
% \end{aligned}
% \end{align}
%
By combining (\ref{inter agent combined bound}) and (\ref{inter-agent bounds on each}), we obtain
\begin{equation}
\begin{aligned}
    & \max_{\bzeta^i \in \calU_i, ~ \bzeta^j \in \calU_j}
          \| \vA_i \tilde{\vM}_k^i \bzeta^i 
         -  \vA_j \tilde{\vM}_k^j \bzeta^j \|_2 \\
    & \qquad \leq
    \bigg( \sum_{\bar{m}= 1}^{m} \max_{\bzeta^i \in \calU_i, ~ \bzeta^j \in \calU_j} 
    (\vA_i \tilde{\vM}_k^i \bzeta^i -  \vA_j \tilde{\vM}_k^j \bzeta^j)_{\bar{m}}^2 \bigg)^{1/2} \\
    &~ \qquad 
    = 
    \bigg( \sum_{\bar{m}= 1}^{m} 
    (\bmu_{max,\bar{m}}^i + \bmu_{max,\bar{m}}^j)^2 \bigg)^{1/2}
\end{aligned} 
\end{equation}
As in the proof of Proposition 2,
let us now introduce the variables $\bmu_{d,k}^i, \bmu_{d,k}^j$ such that
\begin{align}
    & 
    (\bmu_{d,k}^i)_{\bar{m}} \geq  
    (\tau^i)^{1/2} \| \boldsymbol{\Gamma}_i \T \vM_i \T h_{k,\bar{m}}^i\|_{\vS_i^{-1}} \\
    &
    (\bmu_{d,k}^j)_{\bar{m}} \geq  
    (\tau^j)^{1/2} \| \boldsymbol{\Gamma}_j \T \vM_j \T h_{k,\bar{m}}^j\|_{\vS_j^{-1}}
\end{align}
for every $\bar{m} = 1, \dots, m$. Then, we obtain
\begin{equation}
 \begin{aligned}
    \max_{\bzeta^i \in \calU_i, ~ \bzeta^j \in \calU_j}
         \| \vA_i \tilde{\vM}_k^i \bzeta^i &
         -  \vA_j \tilde{\vM}_k^j \bzeta^j \|_2 
    \leq
    \| \bmu_{d,k}^i + \bmu_{d,k}^j \|_2
\end{aligned} 
\end{equation}
so we can consider the following tighter approximation to the constraint (\ref{interagent constraint 2}),
\begin{align}
    \| \bmu_{d,k}^i + \bmu_{d,k}^j \|_2 \leq  \tilde{c}_{ij} - c_{ij}
\end{align}
\subsection{Interpretation of tighter approximation in 2D space}
Here, we provide the intuition behind the approximation in a 2D case considering that the non-convex norm constraints correspond to the obstacle avoidance constraints. In that case, $\vA_i \bmu_{x_k,\bar{u}}^i$ represents the disturbance-free position of the agent $i$, and $\bm{b}_i$ represents the center of the obstacle. \\
We considered the following two approximations -
\subsubsection{\underline{First Approximation}}
\begin{figure}[h!] 
    \centering
    \subfloat[\label{12a}]{
       \includegraphics[width= 0.4\linewidth, trim={1cm 0cm 0cm 0cm},clip]{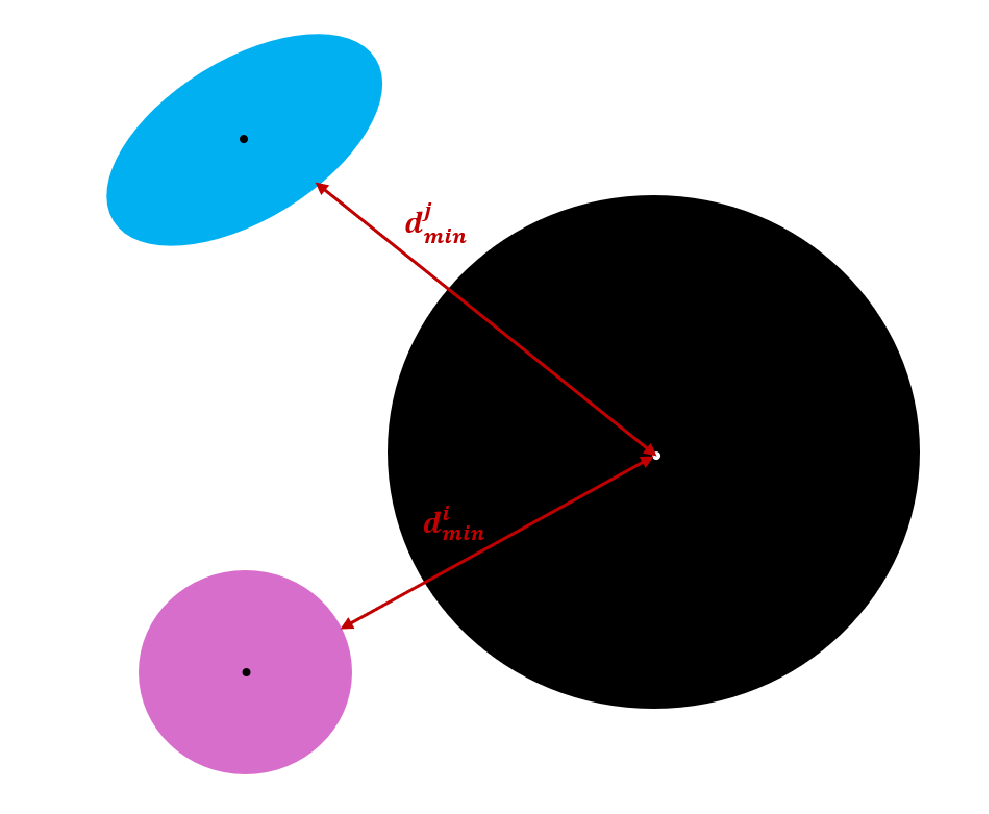}}
    \subfloat[\label{12b}]{
    \includegraphics[width=0.4\linewidth, trim={0cm 0cm 0cm 0cm},clip]{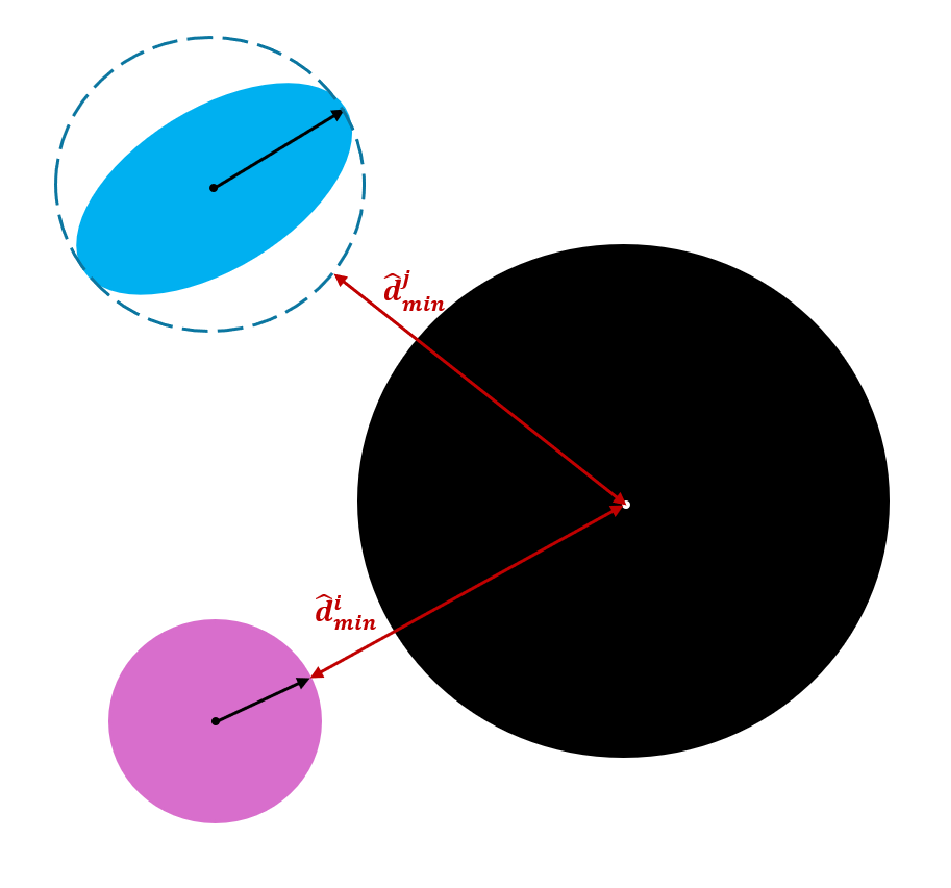}}
  \caption{\textbf{Demonstration of first approximation of obstacle avoidance constraint:}  Robust mean trajectory realizations of two agents represented as ellipses.}     
  \label{fig12} 
\end{figure}

Using the triangle inequality, we considered the following approximation.
    The obstacle avoidance constraint
    \begin{align}
    \min_{\boldsymbol{\zeta}^i \in \calU_i} \| \vA_i \bmu_{x_k,\bar{u}}^i +  \vA_i \tilde{\vM}_k^i \boldsymbol{\zeta}^i - \bm{b}_i \|_2 
    \geq c_i
    \label{obstacle avoidance -  original}
    \end{align}
    is approximated by the following constraint
    % \begin{align}
    % \| \vA_i \bmu_{x_k,\bar{u}}^i - \bm{b}_i \|_2 
    % - \| \vA_i \tilde{\vM}_k^i \boldsymbol{\zeta}^i \|_2
    % \geq c_i, \quad \forall \boldsymbol{\zeta}^i \in \calU_i,
    % \end{align}
    % which can be rewritten as 
\begin{align}
    \| \vA_i \bmu_{x_k,\bar{u}}^i - \bm{b}_i \|_2 
    - \max_{\boldsymbol{\zeta}^i \in \calU_i} 
    \| \vA_i \tilde{\vM}_k^i \boldsymbol{\zeta}^i \|_2
    \geq c_i.
    \label{obstacle avoidance - tighter approximation}
\end{align}
Let us now analyze each of the above constraints. Since, the disturbance $\boldsymbol{\zeta}^i$ lies in an ellipse, and the mean is a linear transformation of $\boldsymbol{\zeta}^i$, the possible realizations of the state mean at each time step would also form an ellipse. Fig.1 and Fig.2 illustrate a two-agent case with an obstacle (black circle) and the mean state of two agents $i$ and $j$ represented by the ellipses. In Fig. \ref{fig12}, $d_{min}^i$ and $d_{min}^j$ represent the smallest distance between the obstacle and the state mean realizations of the agents $i$ and $j$, respectively. The RHS of the original constraint \eqref{obstacle avoidance -  original}, i.e., $ \min_{\boldsymbol{\zeta}^i \in \calU_i} \| \vA_i \bmu_{x_k,\bar{u}}^i +  \vA_i \tilde{\vM}_k^i \boldsymbol{\zeta}^i - \bm{b}_i \|_2$, represents $d_{min}^i$. Thus, the original constraint is to have $d_{min}^i \geq c_i$. 

Now, the RHS of the tighter approximate constraint \eqref{obstacle avoidance - tighter approximation} involves two terms: $\| \vA_i \bmu_{x_k,\bar{u}}^i - \bm{b}_i \|_2$ and $\max_{\boldsymbol{\zeta}^i \in \calU_i} 
    \| \vA_i \tilde{\vM}_k^i \boldsymbol{\zeta}^i \|_2$. The term $\| \vA_i \bmu_{x_k,\bar{u}}^i - \bm{b}_i \|_2$ represents the distance between the disturbance-free state of the agent $i$ (represented by the corresponding center of ellipse in Fig \ref{fig12}) and the obstacle, and the term $\max_{\boldsymbol{\zeta}^i \in \calU_i} 
    \| \vA_i \tilde{\vM}_k^i \boldsymbol{\zeta}^i \|_2$ is
 the maximum possible deviation (represented by a black arrow in Fig \ref{12b}) in the state mean due to the deterministic disturbance. Thus, the tighter approximation is analogous to considering that the deviation in the state mean is the same in all directions, with magnitude equal to $\max_{\boldsymbol{\zeta}^i \in \calU_i}\| \vA_i \tilde{\vM}_k^i \boldsymbol{\zeta}^i \|_2$. 
 As shown in Fig. \ref{12b}, $\hat{d}_{min}^i$ represents the smallest distance between the boundary of the circle formed with the center as disturbance-free state's mean $\vA_i \bmu_{x_k,\bar{u}}^i$ and radius equal to the maximum deviation $\max_{\boldsymbol{\zeta}^i \in \calU_i}\| \vA_i \tilde{\vM}_k^i \boldsymbol{\zeta}^i \|_2$. Thus, the original constraint $d_{min}^i \geq c_i$ is approximated to $\hat{d}_{min}^i \geq c_i$. Further, it can be observed that the approximation only affects the agent $i$ and not the agent $j$.
\subsubsection{\underline{Second Approximation}}
\begin{figure}[h!] 
    \centering
    \subfloat[\label{13a}]{
       \includegraphics[width= 0.4\linewidth, trim={0cm 0cm 0cm 0cm},clip]{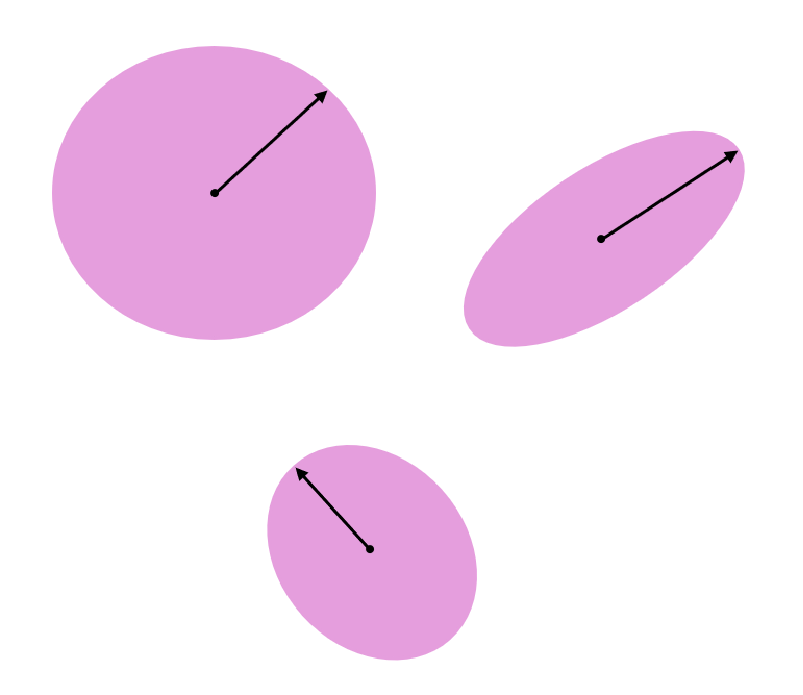}}
    \subfloat[\label{13b}]{
    \includegraphics[width=0.4\linewidth, trim={0cm 0cm 0cm 0cm},clip]{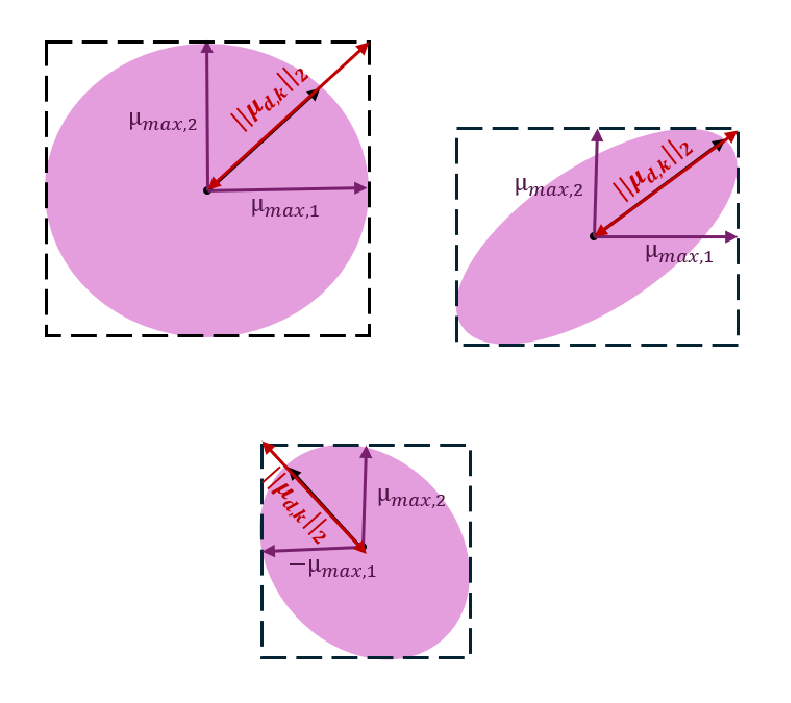}}
  \caption{\textbf{Demonstration of second type of approximation:} Effect of the approximation on various robust mean trajectory realization shapes.}  
  \label{fig13} 
\end{figure}

The second type of approximation is based on the fact that
    \begin{align}
        \max_{\by \in \mathcal{Y}}  \| \vA \by \|_2 
        & = 
        \max_{\by \in \mathcal{Y}} \bigg( \sum_{\bar{m} = 1}^{m} 
        (\ba_{\bar{m}} \T \by)^2 \bigg)^{1/2} \\
        &
        \leq 
        \bigg( \sum_{\bar{m} = 1}^{m} 
        \max_{\by \in \mathcal{Y}} (\ba_{\bar{m}} \T \by)^2 \bigg)^{1/2} 
    \end{align}
    for some matrix $\vA \in \Rb^{m \times n_y}$, and vector $\by \in \Rb^{n_y}$.
    The constraint 
    \begin{equation}
    \max_{\boldsymbol{\zeta}^i \in \calU_i} \| \vA_i \tilde{\vM}_k^i \boldsymbol{\zeta}^i \|_2
    \leq \tilde{c}_i - c_i.
    \label{second Approx - original}
    \end{equation} is approximated by 
    \begin{align}
    \| \bmu_{d,k}^i {} \|_2 \leq \tilde{c}_i - c_i.
    \end{align}
    with the introduction of the slack variable $\bmu_{d,k}^i$. It should be noted that the value of $\bmu_{d,k}^i$ should provide a good approximation to the maximum state deviation term $\max_{\boldsymbol{\zeta}^i \in \calU_i} \| \vA_i \tilde{\vM}_k^i \boldsymbol{\zeta}^i \|_2$ when the agent $i$ is closer to the obstacle. Since $\bmu_{d,k}^i$ is a slack variable, $(\tau^i)^{1/2} \| \boldsymbol{\Gamma}_i \T \vM_i \T h_{k,\bar{m}}^i\|_{\vS_i^{-1}} 
    = (\bmu_{d,k}^i)_{\bar{m}}$ during these instants. Fig. \ref{fig13} provides various examples of realizations of the state mean. The black arrows in the ellipses in Fig. \ref{13a} represent the maximum deviation in the state mean. The red arrows in Fig \ref{13b} represent the quantity $\| \bmu_{d,k}^i \|_2$ with each element $(\bmu_{d,k}^i)_{\bar{m}} = (\tau^i)^{1/2} \| \boldsymbol{\Gamma}_i \T \vM_i \T h_{k,\bar{m}}^i\|_{\vS_i^{-1}} 
    $. It can be observed that $\| \bmu_{d,k}^i \|_2$ provides a good approximation of the maximum deviation for different shapes of the state mean realizations.
\subsection{Distributed Architecture Global Updates}
The global update steps are given by
\begin{align}
    & \{  \Tilde{\bnu}_{\bar{u}}^{i,l+1},
    \Tilde{\bnu}_{d}^{i,l+1}  \}_{i \in \calV} 
    \nonumber \\
    &~~~
    = \argmin \; \calL_{\rho_{\bar{u}}, \rho_d} ( \{ 
    \Tilde{\bmu}_{\bar{u}}^{i,l+1} , 
    \Tilde{\bmu}_{d}^{i, l+1} , 
    \Tilde{\bnu}_{\bar{u}}^{i},
    { \Tilde{\bnu}_{d}^{i} ,
    \blambda^{i,l}_{\bar{u}} , 
    \blambda^{i,l}_{d}} \}_{i \in \calV} ) \nonumber \\
    &~~~
    = \argmin \;
    \sum_{i = 1}^{N} \blambda^{i,l}_{\bar{u}} {}\T
    ( \Tilde{\bmu}_{\bar{u}}^{i,l+1} 
    - \Tilde{\bnu}_{\bar{u}}^{i} )
    +  \blambda^{i,l}_{d} {}\T 
    (\Tilde{\bmu}_{d}^{i, l+1}  - 
     \Tilde{\bnu}_{d}^{i}  ) \\
    &~~~~ \qquad \qquad
    + \frac{\rho_{\bar{u}}}{2} 
    \| \Tilde{\bmu}_{\bar{u}}^{i,l+1} 
    - \Tilde{\bnu}_{\bar{u}}^{i} \|_2^2 \nonumber 
    + \frac{\rho_d}{2} 
    \|  \Tilde{\bmu}_{d}^{i, l+1}  - 
    \Tilde{\bnu}_{d}^{i} \|_2^2 \nonumber    
\end{align}
This update can be decoupled for each agent $i$ as follows
\begin{equation}
 \begin{aligned}
    & 
    \bnu_{\bar{u}}^{i, l+1}
    ,  \bnu_{d}^{i, l+1} \\
    &~~~ =
    \argmin_{\bnu_{\bar{u}}^{i}
    ,  \bnu_{d}^{i} } 
    \; \bar{\blambda}^{i,l}_{\bar{u}} {}\T 
    ( \hat{\bmu}_{\bar{u}}^{i, l+1}  
    -\bnu_{\bar{u}}^i )
    + \bar{\blambda}^{i,l}_{d}  {}\T 
    ( \bmu_{d}^{i,l+1} 
    - \bnu_{d}^i )  \\
    &~~~ \qquad \qquad
    + \frac{\rho_{\bar{u}}}{2} 
    \| \hat{\bmu}_{\bar{u}}^{i, l+1} 
    - \bnu_{\bar{u}}^{i} \|_2^2 
    + \frac{\rho_d}{2} 
    \| \bmu_{d}^{i, l+1} 
    - \bnu_{d}^i  \|_2^2  \\
    &~~~ \qquad \qquad
    + \sum_{\hat{j} \in \calP_i } 
    \big( \bar{\blambda}^{\hat{j},l}_{i, \bar{u}} {}\T 
    ( \bar{\bmu}_{i,\bar{u}}^{\hat{j}, l+1} 
    - \bnu_{\bar{u}}^i) 
    \\
    &~~~ \qquad \qquad
    + \bar{\blambda}^{\hat{j},l}_{i, d} {}\T (
     \bar{\bmu}_{i,d}^{\hat{j}, l+1}
    - \bnu_{d}^i)
    + \frac{\rho_{\bar{u}}}{2} \| \bar{\bmu}_{i,\bar{u}}^{\hat{j}, l+1} 
    -  \bnu_{\bar{u}}^i \|_2^2\\
    &~~~ \qquad \qquad
    + \frac{\rho_d}{2} \| 
     \bar{\bmu}_{i,d}^{\hat{j}, l+1}
    - \bnu_{d}^i \|_2^2
    \big) 
    \end{aligned} 
\end{equation}
To find the minimum on the RHS, let us differentiate the RHS with respect to the global variables $\bnu_{\bar{u}}^{i},  \bnu_{d}^{i} $ and equate it to zero. For $\bnu_{\bar{u}}^{i, l+1}$, we get
\begin{equation}
 \begin{aligned}
    & - \bar{\blambda}^{i,l}_{\bar{u}}
    + \rho_{\bar{u}}
    (\hat{\bmu}_{\bar{u}}^{i, l+1} 
    - \bnu_{\bar{u}}^i) \\
    & \qquad 
    + \sum_{\hat{j} \in \calP_i } 
    \big( - \bar{\blambda}^{\hat{j},l}_{i, \bar{u}}
    + \rho_{\bar{u}} 
    (\bar{\bmu}_{i,\bar{u}}^{\hat{j}, l+1} 
    - \bnu_{\bar{u}}^i ) 
    \big) 
    = 0  \\
\end{aligned} 
\end{equation}
which leads to
\begin{equation}
 \begin{aligned}
    \bnu_{\bar{u}}^{i, l+1} & = 
    \frac{1}{(n(\calP_i) + 1)} \bigg(
    \frac{\bar{\blambda}^{i,l}_{\bar{u}}}{\rho_{\bar{u}}} 
    +  \hat{\bmu}^{i, l+1}_{\bar{u}} 
    \\
    & \quad \quad \quad
    + \sum_{\hat{j} \in \calP_i } \big( \frac{\bar{\blambda}^{\hat{j}, l}_{i, \bar{u}}}{\rho_{\bar{u}}} 
    + \Bar{\bmu}^{\hat{j}, l+1}_{i, \bar{u}}  \big) \bigg)
\end{aligned} 
\end{equation}
%
%
% Similarly, for $\bnu^{d,l+1}_i$ the update is given as -
% \begin{align}
%     \boxed{ \bnu^{d, l+1}_i = 
%     \frac{1}{(n(\calP_i) + 1)} \bigg(
%     \frac{\bar{\blambda}_i^{d,l}}{\rho_{d}}
%     + \bmu^{d,l+1}_i 
%     + \sum_{j \in \calP_i } \big( \frac{\bar{\blambda}_{j,i}^{d,l}}{\rho_{d} }
%     + \bmu^{d,l+1}_{j,i} \big) \bigg) }
% \end{align}
%
The update for $\bnu_{d}^{i, l+1}$ can be derived similarly.
%% Use plainnat to work nicely with natbib. 
%
\subsection{Proposition 4 Proof}
We derive the complexity of solving each local subproblem of an agent in the distributed SDP framework versus the proposed distributed framework. The computational burden of the SDP and SOCP constraints is significant compared to the linear constraints. Therefore, we consider these constraints to determine the worst-case complexity of the algorithms. 
Further, the complexity of solving an optimization problem with SDP or SOCP constraints \cite{LOBO1998193} depends on the number of variables ($n_{\text{var}}$), the number of SDP or SOCP constraints ($n_{\text{SDP}}$ or $n_{\text{SOCP}}$), and the size of each SDP or SOCP constraint involved in the optimization problem. Let us also assume that $n_c$ represents the size of $c^{th}$ constraint. 
% Now, we derive the complexity for each framework.
%
\subsubsection{\underline{Complexity of the Distributed SDP Framework}}
For this framework. we assume that the shared variables among the agents are the control parameters since it is not straightforward to have state means as the shared variables in the distributed SDP approach. 
Now, let us write down the total number of variables involved in the sub-problem of agent $i$ as follows
\begin{equation}
\begin{aligned}
    n_{\text{var}} &=
    T \big( n_{u_i} + n_{u_i} \gamma_h n_{x_i} + n_{obs}+ n_{\text{inter}} 
    \\
    &~~~~ \quad \quad \quad
    + \sum_{j \in \calN_i} n_{u_j} + n_{u_j} \gamma_h n_{x_j} \big)
\end{aligned}  
\end{equation}
where $T(n_{u_i} + n_{u_i} \gamma_h n_{x_i})$,  $T(\sum_{j \in \calN_i} n_{u_j} + n_{u_j} \gamma_h n_{x_j})$ correspond to the control parameters of agent $i$ and its neighbor agents $j \in \calN_i$ respectively, and $T(n_{obs}+ n_{\text{inter}} )$ correspond to the dual variables (as in $\beta$) in the collision avoidance constraints in (28).
Further, this framework involves SDP constraints whose details are given as follows
\begin{itemize}
    \item 
    For obstacle avoidance: number of constraints = $Tn_{obs}$, size of each constraint = $\bar{n}_i + 1$.
    \item 
    For inter-agent collision avoidance: number of constraints = $Tn_{\text{inter}}$, size of each constraint = $\bar{n}_i + \bar{n}_j + 1$.
\end{itemize}
The complexity of each iteration of an interior-point method to solve the local sub-problem with the SDP constraints is given as $O(n_{\text{var}}^2 \sum_{c= 1}^{n_{\text{SDP}}} n_{c}^2)$ \cite{LOBO1998193}. First, let us rewrite $\sum_{c= 1}^{n_{\text{SDP}}} n_{c}^2$ as follows -
\begin{align}
    \sum_{c= 1}^{n_{\text{SDP}}} n_{c}^2 & =
    Tn_{\text{obs}} (\bar{n}_i + 1)^2 
    + T n_{\text{inter}} (\bar{n}_i + \bar{n}_j + 1)^2
\end{align}
As we are interested in finding $O(n_{\text{var}}^2 \sum_{c= 1}^{n_{\text{SDP}}} n_{c}^2)$, let us simplify the contribution of $n_{\text{var}}^2$ and $\sum_{c= 1}^{n_{\text{SDP}}} n_{c}^2$ to eliminate the lesser order terms in $n_{\text{var}}^2 \sum_{c= 1}^{n_{\text{SDP}}} n_{c}^2$. Assuming all the agents have the same control and state dimensions, and are equal to $n_{u_i}$ and $n_{x_i}$ respectively, $n_{\text{var}}^2$ can be simplified to
\begin{equation}
    T^2 \big( (n_{\text{inter}} +1)n_{u_i} \gamma_h n_{x_i} + n_{\text{obs}} \big)^2
\end{equation}
Next, we assume that $\boldsymbol{\Gamma}_i = I$, which gives us $\bar{n}_i = T n_{d_i} + n_{x_i}$. Using this, $\sum_{c= 1}^{n_{\text{\text{SDP}}}} n_{c}^2$ yields
\begin{equation}
\begin{aligned}
    % \sum_{c= 1}^{n_{SDP}} n_{c}^2 
    % \approx 
    % &
    T \big( n_{\text{obs}} T^2 n_{d_i}^2 
    % \\
    % &~~~~~ \quad
    + n_{\text{inter}}  (T^2 n_{d_i}^2 + T^2 n_{d_j}^2 + T^2 n_{d_i} n_{d_j}) \big)
\end{aligned}
\end{equation}
Assuming $n_{d_i} = n_{d_j}$, we could further simplify the above to the following.
\begin{align}
    % \sum_{c= 1}^{n_{SDP}} n_{c}^2 
    % \approx
     (n_{\text{obs}} + n_{\text{inter}}) T^3 n_{d_i}^2
\end{align}
By combining the above simplifications, we get the complexity per interior point iteration to be  
\begin{equation}
\begin{aligned}
    O \Big( T^2 \big( (n_{\text{inter}} +1)n_{u_i} & \gamma_h n_{x_i} + n_{obs} \big)^2 \\
    &~~~~ \quad
    (n_{obs} + n_{\text{inter}}) T^3 n_{d_i}^2 \Big)
\end{aligned} 
\end{equation}
%
% \begin{equation}
% \begin{aligned}
%     n_{var }^2 \sum_{c= 1}^{n_{SDP}} n_{c}^2
%     & \approx
%     T^2 \big( (n_{\text{inter}} +1)n_{u_i} \gamma_h n_{x_i} + n_{obs} \big)^2 \\
%     &~~~~ \quad \quad
%     (n_{obs} + n_{\text{inter}}) T^3 n_{d_i}^2
% \end{aligned}
% \end{equation}
%
% Therefore, the total computational complexity per interior point iteration of solving the subproblem is
which can be upper bounded by 
\begin{equation}
O \Big( T^5 (n_{\text{inter}} + n_{obs})^3 n_{u_i}^2 n_{x_i}^2 n_{d_i}^2 \gamma_h^2 \Big)
\end{equation}
Let $L_{\text{IP}}$ be the number of interior point iterations taken to reach the desired convergence. Then, the total complexity of solving the sub-problem is given as 
\begin{align}
    O \Big( L_{\text{IP}} T^5 (n_{\text{inter}} + n_{obs})^3 n_{u_i}^2 n_{x_i}^2 n_{d_i}^2 \gamma_h^2 \Big).
\end{align}
\subsubsection{\underline{Complexity of the Proposed Distributed Framework}}
Similar to the previous derivation, let us first write the expression for $n_{\text{var}}$ involved in the sub-problem of agent $i$. We have
\begin{equation}
\begin{aligned}
    n_{\text{var}} & = T (n_u + n_u \gamma_h n_x + m (1 + 2 n_{\text{inter}}) \\
    & \qquad \qquad \qquad
    + n_{\text{obs}} + n_{\text{inter}})
\end{aligned}
\end{equation}
where $T (n_u + n_u \gamma_h n_x)$ corresponds to the control parameters of agent $i$, $T m (1 + 2 n_{\text{inter}})$ corresponds to copy variables of neighbor agents at agent $i$, and $T( n_{obs} + n_{\text{inter}})$ corresponds to the slack variables $\tilde{c}_i$, $\tilde{c}_{ij}$ in the collision avoidance constraints.
This framework involves SOCP constraints whose details are given as follows
\begin{itemize}
    \item For obstacle avoidance (constraint (34)): number of constraints = $Tn_{obs}$, size of each constraint = $m +1$.
    \item For inter-agent collision avoidance (constraint (39)): number of constraints = $Tn_{\text{inter}}$, size of each constraint = $m + 1$.
    \item Bounds on $\bmu_d^i$ (constraint (35)): number of constraints = $Tm$, size of each constraint = $\bar{n}_i + 1$.
\end{itemize}
Firstly, it should be noted that the constraints of type (35) are of significantly larger size than the constraints of type (34) and (39). The constraints (35) depend neither on the number of obstacles nor on the number of neighbor agents. Second, all the above constraints are SOCP constraints and are computationally less expensive than solving the SDP constraints of the same size. The complexity of each iteration of an interior-point method to solve the local sub-problem with the SOCP constraints is given as $O(n_{\text{var} }^2 \sum_{c= 1}^{n_{\text{SOCP}}} n_{c})$ \cite{LOBO1998193}. Let us now rewrite $\sum_{c= 1}^{n_{\text{SOCP}}} n_{c}$ as follows 
\begin{equation}
\begin{aligned}
    \sum_{c= 1}^{n_{\text{SOCP}}} n_{c} &=
    Tn_{\text{obs}} (m + 1)
    + Tn_{\text{inter}} (m+1) \\
    & \qquad \qquad \qquad \qquad \qquad
    + Tm (\bar{n}_i + 1).
\end{aligned}
\end{equation}
Based on the same assumptions as in the previous derivation, we simplify the contribution of each term in $O(n_{\text{var}}^2 \sum_{c= 1}^{n_{\text{SOCP}}} n_{c})$ starting with $n_{\text{var}}^2$ simplified to the following
\begin{align}
    T^2(n_u \gamma_h n_x + m n_{\text{inter}} + n_{\text{obs}})^2
\end{align}
since $m$ is constant. We can further reduce the above to
\begin{align}
    T^2(n_u \gamma_h n_x + n_{\text{inter}} + n_{\text{obs}})^2.
\end{align}
Now, the contribution of $\sum_{c= 1}^{n_{\text{SOCP}}} n_{c}$ can be written as
\begin{equation}
 T(m n_{\text{obs}} + m n_{\text{inter}}         
    + m T n_{d_i})
\end{equation}
which can be further simplified to
\begin{align}
     T (n_{\text{obs}} + n_{\text{inter}}         
    + T n_{d_i}). 
\end{align}
Combining the above simplifications for $n_{\text{var}}^2$ and $\sum_{c= 1}^{n_{\text{SOCP}}} n_{c}$, we have the computational complexity per interior point iteration for solving the local subproblem given as 
\begin{equation}
O \Big( T^3 (n_u \gamma_h n_x + n_{\text{inter}} + n_{\text{obs}})^2 \big( n_{\text{obs}} + n_{\text{inter}}                 
    + T n_{d_i} \big) \Big)
\end{equation}
which could be upper bounded by 
\begin{equation}
O \Big( T^4 (n_{\text{inter}} + n_{obs})^3 n_{u_i}^2 n_{x_i}^2 n_{d_i} \gamma_h^2 \Big)
\end{equation}
If $L_{\text{IP}}$ is the number of interior point iterations taken to reach the optimal value, the total complexity of solving the sub-problem is given by
\begin{align}
    O \Big( L_{\text{IP}} T^4 (n_{\text{inter}} + n_{obs})^3 n_{u_i}^2 n_{x_i}^2 n_{d_i} \gamma_h^2 \Big).
\end{align}
\section{Simulation Data}
\subsection{System Parameters}
Each agent $i \in \calV$ is subject to the following 2D double integrator dynamics,
\begin{align}
 % \forall i \in & \llbracket 1, N \rrbracket, \; k \in  \llbracket 0, T-1 \rrbracket, \nonumber \\
 &
 A_k^i = \begin{bmatrix}
    1	& 0	& 0.05 & 0 \\
    0	& 1 & 0 & 0.05 \\
    0	& 0	& 1	& 0 \\
    0	& 0	& 0	& 1
 \end{bmatrix}, \quad
 B_k^i = \begin{bmatrix}
    0.0013 & 	0 \\
    0 &	0.0013 \\
    0.05 &	0 \\
    0 &	0.05
 \end{bmatrix}, 
  \nonumber \\[0.2cm]
 & 
 C_k^i = \texttt{randn}(4, 2), \quad
 D_k^i = \vI_{4 \times 4} \nonumber
\end{align}
for $k \in \llbracket 0,T-1 \rrbracket$.
Further, we enforce that $\| C_k^i \|_F = 1$. We consider $\vS_i = \vI, \vGamma_i = \vI$ for all agents. In each experiment, we use $\tau^i$ as the uncertainty level.
The cost matrices are set to be $\vR_{\bar{u}} = 0.05 \vI$, and  $\vR_{\vK} = 0.05 \vI$. 
% {\color{blue} We set the time horizon (T) to be 25 for the experiment (2) and for the cases considered for creating the computational time graphs with respect to the number of obstacles.}
%
\subsection{ADMM Parameters}
The penalty parameters are set as  $\rho_{\bar{u}} = 100$, $\rho_d = 1$ for two-agent cases and $\rho_{\bar{u}} = 100$, $\rho_d = 10$ for the others, with maximum ADMM iterations set to be 200. Further, we consider the following primal and dual residual conditions as the convergence criteria for ADMM at the $l^{th}$ iteration. The primal residual condition is given by
\begin{equation}
    \bigg(
    \frac{ \sum_{i=1}^N  \| \Tilde{\bmu}^{\bar{u}, l}_i - \Tilde{\bnu}^{\bar{u},l}_i \|^2 
    +  \| \Tilde{\bmu}^{d,l}_i - \Tilde{\bnu}^{d,l}_i \|^2 }{ N + \sum_{i = 1}^{N} n(\calN_i) } \bigg)^{1/2} \leq \epsilon_{\text{primal}}   
\end{equation}
while the dual residual one is 
\begin{equation}
    \frac{ \big( \sum_{i=1}^N  \rho_{\bar{u}}^2 \| \bnu^{\bar{u}, l}_i - \bnu^{\bar{u}, l-1}_i \|^2 
    + \rho_d^2 \| \bnu^{d, l}_i - \bnu^{d, l-1}_i \|^2 \big)^{1/2} }{ N }  \leq \epsilon_{\text{dual}} \nonumber 
\end{equation}
Finally, we set both residual thresholds $\epsilon_{\text{primal}}, \epsilon_{\text{dual}}$ to  $0.1$ for the experiment (2), and to $1$ for all the other experiments. 
\subsection{Constraint Parameters}
For both inter-agent constraints and obstacle avoidance constraints, we use $\vH_{pos} = \begin{bmatrix}
    \vI_{2 \times 2} & 0
\end{bmatrix}$. The collision distance threshold for inter-agent and obstacle avoidance constraints are set at 0.25 and 0.5, respectively, for the experiments (1, 5, 6). Both inter-agent and obstacle collision distance thresholds are set at 0.25 for experiments (2,3,4) and at 0.5 for large-scale experiments (7,8).  Further, in the experiments (3, 4) involving chance constraints, the probability threshold of chance constraints is set to be $p = 0.05$. 
\subsection{Scenarios considered for the computational plots} 
The computational graph with respect to the number of obstacles is based on a six-agent system, where each agent is required to move from a multi-agent vertical formation to a terminal vertical formation. The path between the formations contains obstacles of different radii. 
The computational graph is created based on a multi-agent scenario where each agent must move from an initial rectangular formation to a terminal rectangular formation. This is similar to the scenario shown in Fig. 7, and there are five obstacles that the agents must navigate around. In both the cases, every agent has 4 neighbor agents.

\end{document}